\documentclass[lettersize,journal]{IEEEtran}
\usepackage{amsmath,amsfonts}
\usepackage{algorithmic}
\usepackage{algorithm}
\usepackage{array}
\usepackage{textcomp}
\usepackage{stfloats}
\usepackage{verbatim}
\usepackage{graphicx}
\usepackage{cite}
\usepackage[skip=1pt,font=footnotesize]{subcaption}
\usepackage{adjustbox}
\usepackage[dvipsnames]{xcolor}

\usepackage{float}

\usepackage{todonotes}
\usepackage{xparse}
\usepackage{fancyhdr}
\usepackage{xurl}
\usepackage{enumitem}
\usepackage{environ}
\usepackage{xspace}
\usepackage{scalefnt}
\usepackage{comment}
\usepackage{multirow}

\fancypagestyle{plain}{
	\fancyhf{} 
	\fancyfoot[C]{\thepage} 
}
\newcommand{\dochanges}{}  

\ifdefined\dochanges
 \definecolor{changedcol}{rgb}{0., 0., 0.}
  \definecolor{changedcolTVCG}{RGB}{0., 0., 0.}
\else
  \definecolor{changedcol}{rgb}{0., 0., 0.}
  \definecolor{changedcolTVCG}{RGB}{0., 0., 0.}
\fi

\newenvironment{new}[1]{\color{changedcol}#1}{\color{black}}
\newenvironment{newTVCG}[1]{\color{changedcolTVCG}#1}{\color{black}}
\makeatletter
\newcommand{\cmmnt}[1]{\@bsphack\@esphack}
\makeatother
\environbodyname\envbody
\NewEnviron{old}{\cmmnt{\envbody}}

\newcommand{\Dhat}{{\hat D}}
\newcommand{\Fts}[1]{{\hat F^{#1}_{t\rightarrow s}}}
\newcommand{\Ftu}[1]{{\hat F^{#1}_{t\rightarrow u}}}
\newcommand{\Fteachts}{{\hat F^{\,teach}_{t\rightarrow s}}}
\newcommand{\Fteachtu}{{\hat F^{\,teach}_{t\rightarrow u}}}
\newcommand{\Mteach}{M^{teach}}
\newcommand{\Ft}[1]{{\hat F^{\,#1}_{t}}}
\newcommand{\Dts}[1]{{\hat D^{#1}_{t\leftarrow s}}}
\newcommand{\Dtu}[1]{{\hat D^{#1}_{t\leftarrow u}}}
\newcommand{\Dt}[1]{{\hat D^{\,#1}_{t}}}
\newcommand{\Fhat}{{\hat F}}

\newcommand{\highleftarrow}{\hbox{\raise.80ex\hbox{$\mathbin{^\leftarrow}$}}}
\newcommand{\W}{\hbox{$\mathbin{W\mskip-15.0mu\highleftarrow}$}}

\newcommand{\Loss}{\mathcal{L}}
\newcommand{\Lrec}{\mathcal{L}_{rec}}

\newcommand{\Lflow}{\mathcal{L}_{flow}}

\newcommand{\smallGT}{\mbox{{\scalefont{0.6}{\it{GT}}}}}
\newcommand{\DGT}{D^{\smallGT}_t}
\newcommand{\FGT}{F^{\smallGT}_t}

\newcommand{\cb}[1]{\mbox{Conv Block$^{#1}$}}

\newcommand{\acronym}{FLINT\xspace}

\usepackage{tikz}
\usetikzlibrary{shapes.geometric, arrows, calc}
\usetikzlibrary{positioning, shapes.multipart, arrows.meta}
\usetikzlibrary{arrows,shapes,automata,backgrounds,petri,positioning}
\usetikzlibrary{decorations.pathmorphing}
\usetikzlibrary{decorations.shapes}
\usetikzlibrary{decorations.text}
\usetikzlibrary{decorations.fractals}
\usetikzlibrary{decorations.footprints}
\usetikzlibrary{shadows}
\usetikzlibrary{calc}
\usetikzlibrary{spy}

\usepackage[hidelinks]{hyperref}

\graphicspath{{figures/}{pictures/}{images/}{./}} 

\usepackage{microtype}                 
\PassOptionsToPackage{warn}{textcomp}  
\usepackage{textcomp}                  
\usepackage{mathptmx}                  
\usepackage{times}                     
\usepackage{cite}                      
\usepackage{tabu}                      
\usepackage{booktabs}                  
\usepackage{amsmath}
\usepackage{amssymb}

\newcommand{\orangecircle}{\tikz{\fill[orange] (0,0) circle (2pt);}}
\newcommand{\bluecircle}{\tikz{\fill[blue] (0,0) circle (2pt);}}
\newcommand{\greencircle}{\tikz{\fill[green] (0,0) circle (2pt);}}

\begin{document}

\title{\begin{newTVCG}\acronym: Learning-based Flow Estimation and Temporal Interpolation for Scientific Ensemble Visualization\end{newTVCG}}

\author{Hamid Gadirov, Jos B.T.M. Roerdink, and Steffen Frey
\thanks{Hamid Gadirov, University of Groningen. E-mail: h.gadirov@rug.nl.}
\thanks{Jos Roerdink, University of Groningen. E-mail: j.b.t.m.roerdink@rug.nl.}
\thanks{Steffen Frey, University of Groningen. E-mail: s.d.frey@rug.nl.}}

\markboth{SUBMITTED TO IEEE TRANSACTIONS ON VISUALIZATION AND COMPUTER GRAPHICS}
{Shell \MakeLowercase{\textit{et al.}}: }


\maketitle

\begin{abstract}
We present \begin{newTVCG}\acronym(learning-based {FL}ow estimation and temporal {INT}erpolation),\end{newTVCG}
a novel deep learning-based approach to \begin{newTVCG}estimate\end{newTVCG} flow fields for 2D+time and 3D+time scientific ensemble data. 
  \begin{newTVCG}\acronym\end{newTVCG}can flexibly handle different types of scenarios with (1)~a flow field being partially available for some
  members (e.g., omitted due to space constraints) or (2)~no flow
  field being available at all (e.g., because it could not be acquired
  during an experiment). 
  The design of our architecture allows to flexibly cater to both cases simply by adapting our modular loss functions, effectively treating the different scenarios as 
  \begin{newTVCG}
      flow-supervised and flow-unsupervised
  \end{newTVCG}
  problems, respectively (with respect to the presence or absence of ground-truth flow).
  \begin{old}
  To the best of our knowledge, \begin{newTVCG}\acronym\end{newTVCG}is the first approach that not only generates high-quality temporal interpolants between scalar fields, but especially also estimates a corresponding flow
  field for each intermediate timestep, even when no original flow
  information is provided. 
  \end{old}
  \begin{newTVCG}
    To the best of our knowledge, \begin{newTVCG}\acronym\end{newTVCG}is the first approach to perform flow estimation from scientific ensembles, generating a corresponding flow field for each discrete timestep, even in the absence of original flow information. Additionally, \begin{newTVCG}\acronym\end{newTVCG}produces high-quality temporal interpolants between scalar fields.
  \end{newTVCG}%
  \begin{newTVCG}\acronym\end{newTVCG}
  employs several neural blocks, each featuring several convolutional and deconvolutional layers.
  We demonstrate performance and accuracy for different usage scenarios
  with scientific ensembles from both simulations and experiments.
\end{abstract}


\vspace{-3pt}
\section{Introduction}

\maketitle

 
Technological advancements allow to capture and simulate
time-dependent processes at high resolution, both spatially and
temporally.  
This facilitates the acquisition of results from for
a large number of runs, forming spatio-temporal ensembles.
Such ensembles enable scientists to search a parameter space or estimate the impact of stochastic factors.
However, the large volume of generated data often cannot entirely
be preserved due to storage or bandwidth limitations~\cite{childs2019situ}, and, e.g., timesteps and/or variables need to be omitted.
Data acquired via experiments are also typically restricted regarding
available modalities (e.g., only camera images or scans are
available).  
Reconstructing missing information post-hoc can greatly
support visual analysis in such scenarios.  
Even if all variables of
interest are captured and the full data can be preserved, assessing
how well data can be reconstructed has been demonstrated to be useful
for diverse purposes in visualization, including the choice of
techniques in flow visualization~\cite{janicke_visual_2011}, timestep
selection~\cite{frey_flow-based_2017}, and ensemble member
comparison~\cite{tkachev_local_2021}.

In this paper, we propose \begin{newTVCG}\acronym (learning-based \underline{FL}ow estimation and temporal \underline{INT}erpolation)\end{newTVCG}, a new deep learning-based approach to supplement scientific ensembles with an otherwise missing flow field---in scenarios when it had to be omitted or could not be captured, as is often the case for experiments.
\begin{newTVCG}\acronym not only estimates a corresponding flow
field for each timestep but can further reconstruct high-quality temporal interpolants between scalar fields.
\end{newTVCG}
\begin{newTVCG}

Technically speaking, we consider two types of flow:
\textit{physical flow} and \textit{optical flow}.
In the general case of spatiotemporal $nD$ scalar fields (like density of a fluid flow or luminance in 3D light microscopy image sequences) representing an underlying physical phenomenon, we define optical flow as the observed change of the $nD$ scalar field patterns. 
By contrast, physical flow directly denotes the velocity of the actual material objects (moving particles in a fluid, motion of cells in a tissue volume, etc.).
Physical flow is usually directly given as output of flow solvers,
but can also be determined experimentally where feasible (e.g., via
particle displacement velocimetry).

Optical flow can be estimated from arbitrary scalar fields, also ones that are only indirectly affected by physical flow (like a temperature field\begin{old} or luminance images from cameras\end{old}).
Of course, like in the case of optical flow in computer vision we need to assume that 1) there exist observable temporal changes of the field under investigation; and 2) these changes are correlated with physical changes (in order to be meaningful).
If no physical flow information is given at all---e.g., in the case of
an experiment without dedicated flow measurements or when such information could not be stored due to space constraints---\begin{newTVCG}\acronym\end{newTVCG}can \begin{newTVCG}estimate\end{newTVCG} high-quality optical flow from scientific data as an additional modality for analysis (e.g., using flow glyphs or integral lines allows to effectively present temporal changes).
If physical flow is at least partially given for some members of the ensemble (at the time of model training), \begin{newTVCG}\acronym\end{newTVCG}is capable of making use of this and adds estimated physical flow to members during inference.
\end{newTVCG}

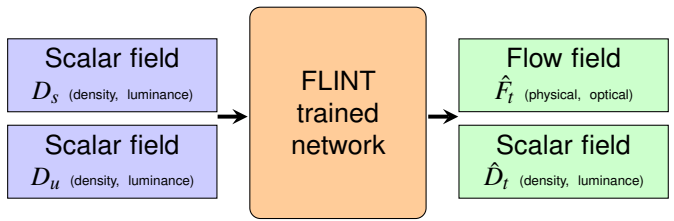
\begin{figure}[t]
\centering
\begin{tikzpicture}[>=stealth,font=\sffamily]
    \tikzstyle{input} = [rectangle, draw, fill=blue!20,     text width=7.25em, text centered, minimum height=2.5em]
    \tikzstyle{nn} = [rectangle, draw, fill=orange!40,     text width=6em, text centered, rounded corners, minimum height=8em]
    \tikzstyle{output} = [rectangle, draw, fill=green!20,     text width=7.25em, text centered, minimum height=2.5em]
    
    \node [input] (input1) {Scalar field $D_s$ {\tiny (density, luminance)}};
    \node [input, below of=input1, yshift=-0.15cm] (input2) {Scalar field  $D_u$ {\tiny (density, luminance)}};

    \node [nn, right of=input1, xshift=2.0cm, yshift=-0.5cm] (nn) {\acronym\\ trained \\ network};
      
    \node [output, right of=nn, xshift=2.0cm, yshift=0.5cm, align=center] (output1) {Flow field $\hat{F}_{t}$ {\tiny (physical, optical)
    }};
    \node [output, below of=output1, yshift=-0.15cm, align=center] (output2) {Scalar field $\hat{D}_{t}$ {\tiny (density, luminance)}};
    
    \draw [->, line width=1.5pt] (1.4, -0.55) -- (1.8, -0.55);
    \draw [->, line width=1.5pt] (4.2, -0.55) -- (4.6, -0.55);
\end{tikzpicture}
\vspace{5pt}
\caption{Overview of
      \begin{newTVCG}\acronym\end{newTVCG}pipeline during inference. The trained deep neural network performs 
      flow field \begin{newTVCG}estimation\end{newTVCG} $\hat{F}_{t}$ and temporal (scalar) field interpolation $\hat{D}_{t}$, where $s < t < u$, by utilizing the available densities $D_s$ and $D_{u}$ from the previous and following timesteps.
      \vspace{-6pt}
      }
\label{fig:nn_overview}
  \end{figure}

\begin{newTVCG}
\acronym is versatile and can operate with any scalar field data, whether 2D or 3D. 
In cases where only scalar data is available, \begin{newTVCG}\acronym\end{newTVCG}can \begin{newTVCG}estimate\end{newTVCG} optical flow. 
When \begin{newTVCG}vector data are available for some members\end{newTVCG}, it can estimate the missing physical flow. 
Even if the scalar field is correlated but not directly governed by the vector field, \begin{newTVCG}\acronym can learn the vector field\end{newTVCG}, although the optical and physical flow may differ in this case. Fig.~\ref{fig:nn_overview} shows an overview of the \begin{newTVCG}\acronym\end{newTVCG}pipeline.
\end{newTVCG}
    
\begin{newTVCG}\acronym\end{newTVCG}\begin{old}is a deep neural network that\end{old}  draws inspiration from recent
state-of-the-art advancements in computer vision
research~\cite{teed2020raft, dosovitskiy2015flownet, luo2021upflow,
  huang2022real}. 
\begin{newTVCG}\acronym\end{newTVCG}especially builds upon RIFE (Real-time
Intermediate Flow Estimation)~\cite{huang2022real}---a
method for video frame interpolation based on optical flow ---and
extends it in several ways to enable the accurate learning of flow
information\begin{old}, a process we call \emph{enrichment}\end{old},
and yield high-quality temporal interpolants for scientific data.

\begin{newTVCG}\acronym\end{newTVCG}substantially improves over RIFE for better temporal
interpolation with scientific ensembles while being significantly
faster, and enables the 
\begin{old}
\emph{enrichment} of
  spatiotemporal data with---otherwise missing---meaningful
flow fields to open up new opportunities for
analysis\footnote{``flow enrichment'' is shorthand for ``data enrichment with flow information''}.
\end{old}
\begin{newTVCG}
estimation of meaningful flow fields from spatiotemporal data, which would otherwise be missing, to open up new opportunities for analysis.
\end{newTVCG}
Its flexible design allows \begin{newTVCG}\acronym\end{newTVCG}to be applied to different scenarios
without the need for architectural modifications --- regardless of
whether the flow field is (partially) available or not --- simply by
adapting the loss functions used in training the neural network.
This makes \begin{newTVCG}\acronym\end{newTVCG}suitable for both simulation and experimental
ensembles, where the latter may typically lack available flow field
data. 
We refer to scenarios with and without ground-truth flow as ``flow-supervised'' and ``flow-unsupervised'', respectively.
\begin{newTVCG}\acronym\end{newTVCG}performs flow field estimation and temporal interpolation in an integrated way.

\begin{newTVCG}
A first application of flow estimation by \begin{newTVCG}\acronym\end{newTVCG}is its facilitation of applying flow visualization techniques to datasets initially consisting solely of scalar data, which often occurs in practice for experimental data (e.g., \cite{geppert2016classification,frey2022visual}). 
This broadens the scope of visualization possibilities and supports the presentation of dynamic processes.
Another application arises in large-scale data analysis. 
For example, simulations on supercomputers produce massive amounts of data, only a small fraction of which can practically be stored~\cite{childs2020terminology}.
This often necessitates omitting (i) certain variables (like the flow field) and/or (ii) timesteps, either uniformly or via adaptive selection~\cite{frey_flow-based_2017,yamaoka2019situ}.
With \begin{newTVCG}\acronym\end{newTVCG}we can reconstruct data reduced both by (i) and (ii), thus enabling comprehensive analysis and understanding of complex phenomena.
\end{newTVCG}

In summary, we propose \begin{newTVCG}\acronym,\end{newTVCG} a new flexible method for the \begin{newTVCG}estimation\end{newTVCG} of flow for 2D+time and 3D+time scientific ensembles.
  The \acronym code will be \begin{new}made available publicly.\end{new}
We consider our main contributions to be as follows:
\begin{enumerate}[leftmargin=*, label=\textbullet, noitemsep]
\item 
    To the best of our knowledge, \acronym is the first approach to
    \begin{new}achieve\end{new} high-quality flow estimation in scenarios (1) where flow information is partially available (\textit{flow-supervised}, e.g., simulations where flow is omitted due to storage constraints) or (2) entirely unavailable (\textit{flow-unsupervised}, e.g., experimental image sequences captured via cameras).

\item \begin{newTVCG}\acronym\end{newTVCG}produces high-quality temporal interpolants between scalar fields, such as density or luminance.

  \item \begin{newTVCG}\acronym\end{newTVCG}can handle both 2D+time and 3D+time ensembles without requiring domain-specific assumptions, complex pre-training, or intermediate fine-tuning on simplified datasets.
  
\begin{old}
\item \begin{newTVCG}\acronym\end{newTVCG}does not rely on domain-specific assumptions, does not
  require complex pre-training or intermediate fine-tuning on
  simplified datasets, and is capable of handling cases (1)~in which
  flow information is partially available (e.g., simulations where the
  information is omitted due to space constraints) or (2)~not captured
  at all (e.g., experiments captured via cameras).
\item We demonstrate high-quality flow enrichment in both
  aforementioned scenarios, considering data both from simulations and
  experiments.
\end{old}
\end{enumerate}

\vspace{-6pt}
\section{Related Work}
\label{sec:related_work}
	
\textbf{Flow field estimation in scientific visualization.}  
Kappe \emph{et al.}~\cite{kappe2015reconstruction} reconstructed and
visualized 3D local flow from 3D+time microscopy data using both
image processing and optical flow methods.  
Kumpf \emph{et al.}~\cite{kumpf2018visual} utilized ensemble sensitivity analysis (ESA) to obtain the evolution of sensitivity features in 2D or 3D
geo-spatial data.  They used forward and backward optical flow-based feature assignment for tracing the evolution of sensitivities through time.  
Manandhar \emph{et al.}~\cite{manandhar2018sparse} developed a
3D optical flow estimation method for light microscopy image volumes.
A dense 3D flow field was obtained from the 3D displacement vectors in
a pair of such volumes.  
\begin{new}
Gu \emph{et al.} introduced Scalar2Vec~\cite{gu2022scalar2vec}---a deep learning-based framework that translates scalar fields into velocity vector fields, using a k-complete bipartite translation network (kCBT-Net).
We conceptually compare our findings to Scalar2Vec in Sec.~\ref{sec:compar_eval}.
\end{new}However, these approaches involve case-specific assumptions~\cite{kappe2015reconstruction, kumpf2018visual, manandhar2018sparse} or rely on ground truth (GT)~\cite{gu2022scalar2vec} from synthetic data, limiting their generalizability to
scientific ensembles derived from physical experiments.
In contrast, \begin{newTVCG}\acronym\end{newTVCG}does neither require data-specific assumptions nor an available GT flow field.

\textbf{Optical flow learning in computer vision.}
  In computer vision, optical flow is defined as the observed change
  of the pixel brightness pattern in two (or more) images of a moving
  object. The goal is to compute an approximation to the 2D motion
  field---a projection of the 3D velocities of surface points of the
  object onto the image plane---from spatiotemporal patterns of image
  intensity.
  Ideally, the optical flow is the same as the
  motion field, but this need not always be the case. Optical flow
  methods estimate physical velocity components by relying
  on constraints such as brightness constancy and imposing local or
  global spatial smoothness of the estimated velocity field, while
  aiming to preserve motion discontinuities and detect occlusions.
Optical flow estimation has become a crucial component in computer vision applications, including tasks such as video frame interpolation.  
Recent deep learning-based methods such as RAFT and
RAFT-3D~\cite{teed2020raft, teed2021raft},
FlowNet\,(2.0)~\cite{dosovitskiy2015flownet, ilg2017flownet},
PWC-Net~\cite{sun2018pwc}, and ProFlow~\cite{maurer2018proflow}
achieved remarkable results, yet require complex multi-stage optimization (involving pre-training on simpler datasets and fine-tuning on target
datasets), and/or need GT flow vectors for training.  
Estimating optical flow in scientific ensembles, characterized by unbounded structures
and density-only representations, is difficult, and manual labeling is
impractical.  While effective on benchmark datasets, generalization to scientific ensembles poses a substantial challenge.
\begin{old} 
Our proposed method, \begin{newTVCG}\acronym\end{newTVCG}, aims to overcome these limitations by
learning, for instance, the flow field from the available scalar field
of the ensemble.  Leveraging recent deep learning advancements in
optical flow estimation, \begin{newTVCG}\acronym\end{newTVCG}can perform density \begin{newTVCG}interpolation\end{newTVCG} via
interpolation between fields of different timesteps without strictly
relying on GT flow vectors during training.  This enhances its
applicability to scientific ensembles, especially where GT flow data
from experimental ensembles is unavailable.
\end{old}

\begin{newTVCG}\acronym\end{newTVCG}is based on RIFE~(Real-Time Intermediate Flow
Estimation for Video Frame Interpolation)~\cite{huang2022real} which estimates optical flow \begin{old}for
real-time video frame interpolation\end{old} using a student-teacher
  architecture.  
  However, RIFE targets 2D real-world RGB
datasets, such as Vimeo90K~\cite{xue2019video},
Middlebury~\cite{baker2011database}, UCF101~\cite{soomro2012dataset},
and HD~\cite{bao2019memc}, while \begin{newTVCG}\acronym\end{newTVCG}can also handle 3D
  \begin{newTVCG}
      ensembles.
  \end{newTVCG}
For learning physical flow, we employ a loss function inspired by the
RAFT loss~\cite{teed2020raft}.  In scenarios where the flow field is
unavailable during model training, we add additional loss components,
such as photometric loss~\cite{yu2016back}, to facilitate flow-unsupervised learning of the model.

\textbf{Machine learning-based upscaling \& super-resolution.}  ML-based upscaling and super-resolution approaches
have gained significant attention across various domains, including
image processing, computer vision, and scientific
visualization~\cite{ledig2017photo, shi2016real}.  They can be
categorized into three main groups: spatial super-resolution (SSR),
temporal super-resolution (TSR), and spatio-temporal super-resolution
(STSR).  SRCNN~\cite{dong2015image}, SRFBN~\cite{li2019feedback}, and
SwinIR~\cite{liang2021swinir} are SSR techniques that aim to improve
the spatial resolution of images by generating realistic and
fine-grained details.  TSR techniques focus on interpolating
intermediate timesteps from subsampled time sequences while
maintaining the spatial resolution.  Methods such as phase-based
interpolation~\cite{meyer2015phase}, SepConv~\cite{niklaus2017video},
and SloMo~\cite{jiang2018super} tackle the challenge of synthesizing
high-quality intermediate frames to enhance the temporal resolution of
videos.  While previous approaches, like STNet~\cite{han2021stnet},
can upscale data with a fixed spatial or temporal scale factor, they
do not simultaneously address both spatial and temporal
super-resolution.
Han \emph{et al.} proposed a recurrent generative network,
TSR-TVD~\cite{han2019tsr}, designed to be trained on one variable and
then applied to generate a temporally higher-resolution version of
another variable from its lower-resolution counterpart.  
Recent methods like 
\begin{newTVCG}
Filling the Void~\cite{mishra2022filling}, 
SSR-TVD~\cite{han2020ssr},
\end{newTVCG}
FFEINR~\cite{jiao2023ffeinr}, HyperINR~\cite{wu2023hyperinr}, CoordNet~\cite{han2022coordnet}, and 
\begin{newTVCG}
STSR-INR~\cite{tang2024stsr}
\end{newTVCG}
have made significant strides in enabling TSR or SSR of data fields at arbitrary resolution.  
While \acronym achieves state-of-the-art accuracy in temporal interpolation between two given fields, its primary objective and notable strength lies in flow estimation. 
\begin{new}
This task involves predicting flow fields directly from given scalar fields, assuming no flow information is available during the inference stage. Methods like CoordNet or STSR-INR, which are powerful in TSR or SSR tasks, are not designed for 
predicting vector fields from scalar ones.
\end{new}Our \acronym method not only excels in reconstructing temporal data with high precision but also produces impressive results in supplementing the data with accurate flow information.

\textbf{Student-Teacher Learning.}  
  In machine learning, techniques have been developed under the name of teacher-student architecture. 
  One approach is \emph{knowledge distillation}~\cite{hinton15:_distil_knowl_neural_networ} to
  transfer the knowledge (parameters) learnt by a larger model
  (\emph{teacher model}) and transfer it to a smaller model
  (\emph{student model}). A separate strand of research uses the
  concept of \emph{privileged
    information}~\cite{vapnik15:_learn_using_privil_infor}, where the teacher provides the student during training with
  additional ``privileged'' information for knowledge transfer. Both approaches have been
  unified~\cite{lopez-paz16:_unify}.
  The learning can be \emph{offline}, where student networks learn the
  knowledge from pre-trained teacher networks, or \emph{online}, where
  student and teacher networks are simultaneously trained, so that the
  whole knowledge learning process can be end-to-end trainable~\cite{hu22:_teach_studen_archit_knowl_learn}.
  \begin{newTVCG}
      We have implemented an online student-teacher model and instead of a fully separate teacher network, we employ only one additional block that corresponds to the teacher network, akin to Huang et al.~\cite{huang2022real}.
  \end{newTVCG}


\begin{figure*}[t]
  \begin{adjustbox}{raise=1.65cm}
    \begin{minipage}[b]{0.15\textwidth}
      \centering
      \begin{tikzpicture}[>=stealth,font=\sffamily]
        \tikzstyle{input} = [rectangle, draw, fill=blue!30, text
        width=5em, text centered, minimum height=2.5em]
					
        \node (input) [text=black, above] {\small \textbf{Input}}; \node
        (input1) [input, below=0.25cm of input] {$D_s$, $D_{u}$, $t$};
        \node (input2) [input, below=0.25cm of input1, dashed,
        fill=blue!15] {$\FGT$}; \node (input3) [input, below=1.8cm
        of input2, fill=red!15] {$\DGT$};
					
        \draw [->, line width=1.2pt] (1.1, -1.25) -- (1.8, -1.25);
        \draw [->, line width=1.2pt, red] (1.1, -4.5) -- (1.8,
        -4.5);
					
      \end{tikzpicture}
    \end{minipage}
  \end{adjustbox}
	%
  \setlength{\fboxrule}{4pt}
  \begin{tikzpicture}
    \node[draw, orange!60, fill=white, rounded corners, inner sep=4pt,
    line width=3pt, text=black] at (0,0) {
      \begin{adjustbox}{raise=0.35cm}
        \begin{minipage}[b]{0.28\textwidth} %
          \tikzstyle{convblock} = [rectangle, rounded corners, minimum
          width=2.5cm, minimum height=0.9cm,text centered, draw=black,
          fill=SpringGreen] \tikzstyle{arrow} = [thick,->,>=stealth]
          \begin{tikzpicture}[node distance=0.5cm]
	\node (input) [draw, fill=gray!20, minimum width=1.0cm, minimum height=0.7cm] {$D_s$, $D_{u}$, $t$};
	\node (conv0) [convblock, below of=input, yshift=-0.7cm] {Conv Block$^0$};
        \node (sum0) [draw, circle, below of=conv0, yshift=-0.5cm, fill=green!20, inner sep=0.1cm] {{+}};
        \node [above right of=sum0, font=\scriptsize, xshift=0.55cm, yshift=-0.05cm] {$\Fts{0}, \Ftu{0}, M^{0}$};
	\node (conv1) [convblock, below of=sum0, yshift=-0.5cm] {Conv Block$^1$};
	\node (sum1) [draw, circle, below of=conv1, yshift=-0.5cm, fill=green!20, inner sep=0.1cm] {{+}};
         \node [above right of=sum1, font=\scriptsize, xshift=0.55cm, yshift=-0.05cm] {$\Fts{1}, \Ftu{1}, M^{1}$};
	\node (dots) [below=0.25cm of sum1, font=\huge] {$\dots$};
	\node (dots1) [below=0.05cm of dots, font=\huge] {$\dots$};
	\node (conv2) [convblock, below of=dots1, yshift=-0.5cm] {Conv Block$^{N-1}$};
	\node (sum2) [draw, circle, below of=conv2, yshift=-0.5cm, fill=green!20, inner sep=0.1cm] {{+}};
        \node [above right of=sum2, font=\scriptsize, xshift=0.75cm, yshift=-0.05cm] {$\Fts{N-1}, \Ftu{N-1}, M^{N-1}$};
	\node (conv3) [convblock, below of=sum2, yshift=-0.5cm, fill=red!20] {Conv Block$^{\text{teach}}$};
	\node (output) [draw, below of=conv3, fill=gray!20, minimum width=1cm, minimum height=0.7cm, yshift=-0.7cm] {\color{red}$\Fteachts, \Fteachtu, M$};
						
	\draw [arrow] (input) -- (conv0);
	\draw [arrow] (conv0) -- (sum0);
	\draw [arrow] (sum0) -- (conv1);
	\draw [arrow] (conv1) -- (sum1);
	\draw [arrow] (sum1) -- (dots);
	\draw [arrow] (dots1) -- (conv2);
	\draw [arrow] (conv2) -- (sum2);
	\draw [arrow] (sum2) -- (conv3);
	\draw [arrow] (conv3) -- (output);
	\draw [arrow] (input.west) -- ++(-1.2,0) |- (sum0.west);
	\draw [arrow] (input.west) -- ++(-1.2,0) |- (sum1.west);
	\draw [arrow] (input.west) -- ++(-1.2,0) |- (sum2.west);
	\draw [arrow] (input.west) -- ++(-1.2,0) |- (dots1.west);
	\draw [arrow, red] (conv3.west -| {-2.2cm,0}) -- node[midway, above, red] {$\DGT$}  (conv3.west);
	\draw [arrow] (input.west) -- ++(-1.2,0) |- (sum0.west);
	\draw [arrow] (conv0.east) -- ++(0.8,0) |- (sum1.east);
	\draw [arrow] (conv1.east) -- ++(1.1,0) |- (dots1.east);
	\draw [arrow] (dots.east) -- ++(1.6,0) |- (sum2.east);
	\draw [thick, dashed] (conv2.north east) -- (3.1, -0.05);
	\draw [thick, dashed] (conv2.south east) -- (3.1, -9.55);
      \end{tikzpicture}
    \end{minipage}
  \end{adjustbox}
  \begin{minipage}[b]{0.28\textwidth} 
    \color{black} \centering
    \tikzstyle{convlayer} = [rectangle, rounded corners, minimum width=3.5cm, minimum height=0.6cm,text centered, draw=black, fill=Emerald!50]
    \tikzstyle{warp} = [rectangle, rounded corners, minimum width=2.0cm, minimum height=0.6cm,text centered, draw=black]
    \tikzstyle{concat} = [ellipse, minimum width=0.5cm, minimum height=0.6cm,text centered, draw=black, fill=green!20]
    \tikzstyle{arrow} = [thick,->,>=stealth]
    \begin{tikzpicture}[node distance=0.5cm]
      \node (input) [draw, fill=gray!20, minimum width=1cm, minimum height=0.5cm, yshift=0.5cm, font=\footnotesize] {$\Fts{i-1}, \Ftu{i-1}, M^{i-1}, D_s, D_{u}, t$};
      \node (warp) [warp, below of=input, yshift=-0.4cm, fill=yellow!40, font=\footnotesize] {Warping};
        \node [below right of=warp, font=\scriptsize, xshift=0.3cm, yshift=-0.15cm] {$\Dts{i-1}, \Dtu{i-1}$};
      \node (sum0) [concat, below of=warp, yshift=-0.5cm, fill=green!20, inner sep=0.1cm, font=\footnotesize] {Concat};
      \draw [arrow] (input.east) -- ++(0.4,0) |- (sum0.east);
      \node (conv1) [convlayer, below of=sum0, yshift=-0.4cm, fill=Green!80, minimum width=3.2cm, font=\footnotesize] {2 $\times$ Conv, st.=2};
       \node (conv2) [convlayer, below of=conv1, yshift=-0.4cm, minimum width=2.6cm, font=\footnotesize] {3 $\times$ Conv, st.=1};
       \node (conv3) [convlayer, below of=conv2, yshift=-0.4cm, minimum width=2.6cm, fill=Green!80, font=\footnotesize] {1 $\times$ Conv, st.=2};
       \node (conv4) [convlayer, below of=conv3, yshift=-0.4cm, minimum width=2.0cm, font=\footnotesize] {3 $\times$ Conv, st.=1};
        \node (conv5) [convlayer, below of=conv4, yshift=-0.4cm, minimum width=2.0cm, fill=violet!20, font=\footnotesize] {1 $\times$ Deconv, st.=2};
        \node (conv6) [convlayer, below of=conv5, yshift=-0.4cm, minimum width=2.5cm, font=\footnotesize] {3 $\times$ Conv, st.=1};
	\node (sum2) [draw, circle, below of=conv6, yshift=-0.3cm, fill=green!20, inner sep=0.1cm] {+};
	\node (conv7) [convlayer, below of=sum2, yshift=-0.3cm, minimum width=2.6cm, fill=violet!20, font=\footnotesize] {2 $\times$ Deconv, st.=2};
	\node (output) [draw, below of=conv7, fill=gray!20, minimum width=1cm, minimum height=0.5cm, yshift=-0.5cm, font=\footnotesize] {$\Fts{i}, \Ftu{i}, M^{i}$};
							
							\draw [arrow] (input) -- (warp);
							\draw [arrow] (conv0) -- (sum0);
							\draw [arrow] (sum0) -- (conv1);
							\draw [arrow] (conv1) -- (conv2);
                                \draw [arrow] (conv2) -- (conv3);
                                \draw [arrow] (conv3) -- (conv4);
                                \draw [arrow] (conv4) -- (conv5);
                                \draw [arrow] (conv5) -- (conv6);
							\draw [arrow] (conv6) -- (sum2);
							\draw [arrow] (sum2) -- (conv7);
							\draw [arrow] (conv7) -- (output);
							\draw [arrow] (conv1.east) -- ++(0.4,0) |- (sum2.east);
							\draw [arrow, red] (sum0.west -| {-1.6cm,0cm}) -- node[midway, above, red, align=left] {\small $\DGT$ {\tiny $|$ only for}\\ \vspace{-0.5pt} {\tiny Conv Block$^{\text{teach}}$}}  (sum0.west);
							\draw[dashed, rounded corners=20pt] (-2.35,-9.75) rectangle (2.35,1.05);
						\end{tikzpicture}%
				\end{minipage}
				\begin{adjustbox}{raise=1.3cm}
					\begin{minipage}[b]{0.21\textwidth} 
                    \color{black}
						\centering
						\begin{tikzpicture}[>=stealth,font=\sffamily]
							\tikzstyle{loss1} = [rectangle, draw, fill=cyan!40, text width=5em, text centered, minimum height=2.5em]
                                \tikzstyle{arrow} = [thick,->,>=stealth]
							\node [text=black, above] at (0,0.75) {\small\textbf{Loss function $\Loss$}};
							\node [loss1, text width=5.75em, minimum height=3em] (loss1) {$\mathcal{L}_{rec}$};
							\node [below=0.2cm of loss1, font=\Large] {+};
							
							\node [loss1, below=1cm of loss1, dashed, fill=cyan!25] (loss2) {$\mathcal{L}_{flow}$};
							\node [below=0.2cm of loss2, font=\scriptsize] {OR};
							
							\node [loss1, below=0.85cm of loss2, fill=cyan!40] (loss3) {$\mathcal{L}_{dis}$};
							\node [loss1, below=0.6cm of loss3, fill=cyan!40] (loss4) {$\mathcal{L}_{photo}$};
							\node [loss1, below=0.6cm of loss4, fill=cyan!40] (loss5) {$\mathcal{L}_{reg}$};

                                \draw [arrow] (loss1.west -| {-2.4cm,0}) -- node[midway, above] {$\Dhat_{t}, {\color{red}\DGT}$}  (loss1.west);

                               \draw (loss1.west -| {-2.4cm,0}) -- node[midway, below] {{\color{red}$\hat{D}_{t}^{\text{teach}}$}}  (loss1.west);
                                 
                               \draw [arrow] (loss2.west -| {-2.5cm,0}) -- node[midway, above] {$\Ft{i}, {\color{red}\FGT}$}  (loss2.west);
                               \draw  (loss2.west -| {-2.4cm,0}) -- node[midway, below] {{\color{red}$\hat{F}_{t}^{\text{teach}}$}}  (loss2.west);
							

                               \draw [arrow] (loss3.west -| {-2.4cm,0}) -- node[midway, above] {$\hat{F}^{N-1}_{t \rightarrow j}$}  (loss3.west);

                               \draw  (loss3.west -| {-2.4cm,0}) -- node[midway, below] {{\color{red}$\hat{F}_{t \rightarrow j}^{\text{teach}}$}}  (loss3.west);

                               \draw [arrow] (loss4.west -| {-2.4cm,0}) -- node[midway, above] {$D_s, D_u$}  (loss4.west);

                               \draw  (loss4.west -| {-2.4cm,0}) -- node[midway, below] {$\Dhat_t$}  (loss4.west);

                               \draw [arrow] (loss5.west -| {-2.4cm,0}) -- node[midway, above] {$W_i$}  (loss5.west);

                               \draw  (loss5.west -| {-2.4cm,0}) -- node[midway, below] {{\color{red}$W_i^{\text{teach}}$}}  (loss5.west);

\draw[line width=1pt] (-1.2,-7.4) rectangle (1.2,-1.3);
							
						\end{tikzpicture}
					\end{minipage}
				\end{adjustbox}
       };
\end{tikzpicture}
		\begin{picture}(0,0)
			\put(-85, 300){\textbf{\normalsize{\begin{newTVCG}\acronym\end{newTVCG}Network}}}
		\end{picture}

		\caption{ \begin{newTVCG}\acronym\end{newTVCG}network
                  architecture
                  and pipeline during training (see corresponding
                  Fig.~\ref{fig:nn_overview} for inference): given the
                  input fields $D_s$, $D_{u}$, $\DGT$
                    (GT scalar field at time $t$) 
                  and $\FGT$ (GT flow at time
                    $t$) in scenarios in which the latter is
                  available, \begin{newTVCG}\acronym\end{newTVCG}predicts the $\hat{D}_{t}$ scalar field
                  and \begin{old}$\hat{F}_{t}$\end{old}
                  $\hat{F}^{i}_{t}$ flow
                  fields \begin{old}in the
                    output\end{old} used in the loss
                    function for optimizing network
                    parameters.  The \begin{newTVCG}\acronym\end{newTVCG}model architecture
                  and loss function are shown in the orange box.  The
                  model consists of several stacked blocks of the
                  convolutional network, which takes $D_s$, $D_{u}$,
                  and $t$ as input and \begin{old}outputs estimated
                    flows 
                    $\hat{F}_{t \rightarrow s}$,
                    $\hat{F}_{t \rightarrow u}$, and the fusion mask
                    $M$\end{old}
                  in the $i^{th}$ $\cb{}$ computes
                    estimated flows $\Fts{i}, \Ftu{i}$, and fusion
                    mask $M^i$ used for interpolation.  We
                  obtain the best results with four ($N=4$) blocks.
                  The teacher block \begin{newTVCG}Conv Block$^{teach}$\end{newTVCG}, which receives $\DGT$
                  (the GT field at time
                    $t$, \begin{newTVCG}see red arrows\end{newTVCG}) as additional input, is only used
                  during training. The zoomed-in view highlights the \begin{newTVCG}structure of a generic Conv Block\end{newTVCG}
                  consisting of backward warping,
                  concatenation, as well as convolutional and
                  deconvolutional layers with specified strides
                  ($st.$). \begin{newTVCG} The $\DGT$ input at the concatenation stage is only used for the teacher Conv Block.\end{newTVCG}   The loss function, which can be adjusted
                  depending on the scenario, is shown on the right
                  within the orange box.  $W_i$ and
                    $W_i^{teach}$ represent the $i^{th}$ weight matrix
                    of the last convolutional block in the student and
                    teacher network, respectively. \begin{newTVCG}\acronym\end{newTVCG}uses
                  the same model architecture for ensembles with and
                  without
                  available \begin{old}velocity\end{old} GT
                    flow fields. The GT flow $\FGT$ is only used in the loss function
                    $\Lflow$.
                    \vspace{-10pt}
                    }
		\label{fig:model}
\end{figure*}
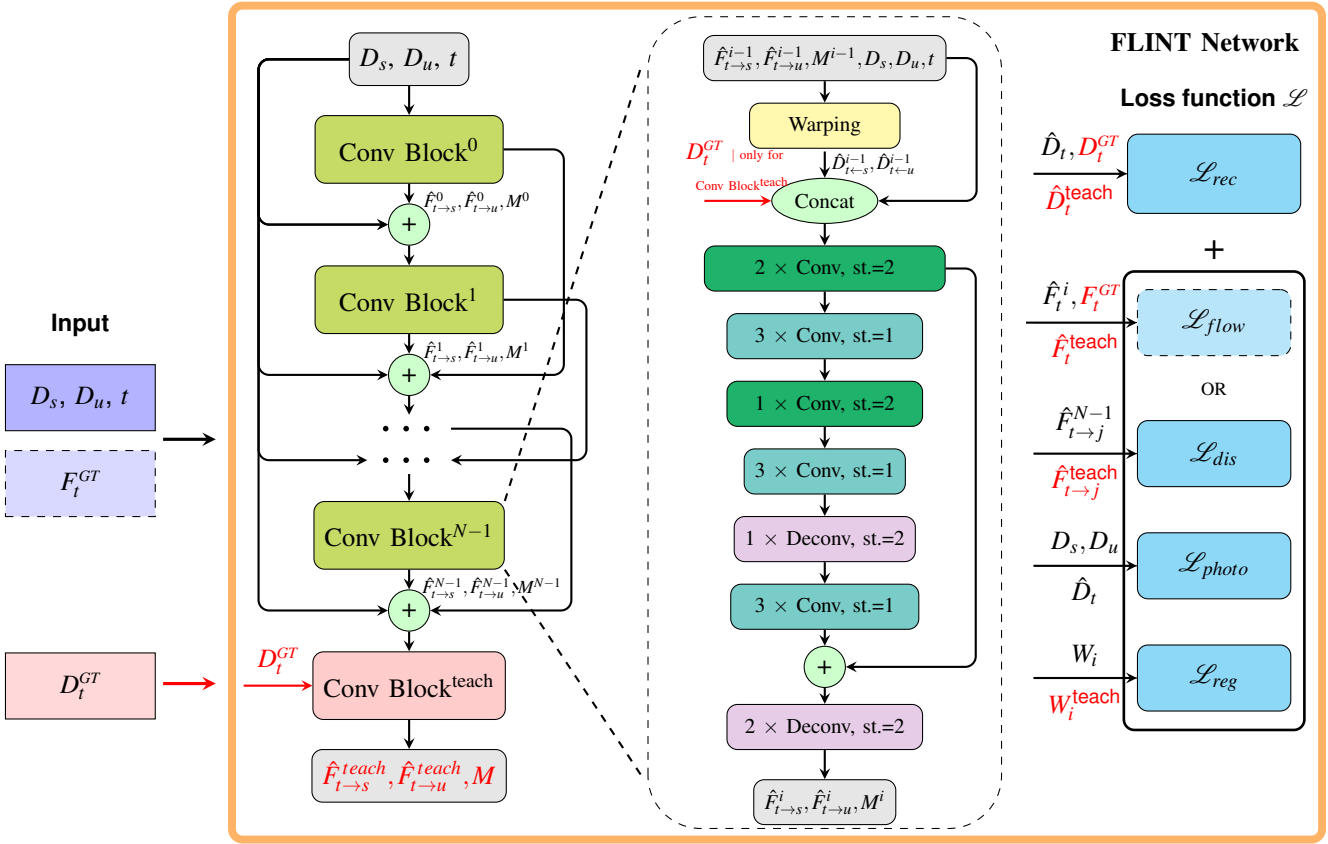

\section{Method}
\label{sec:method}
  \begin{newTVCG}
    \acronym is, to the best of our knowledge, the first method capable of performing flow estimation from available scalar fields in scientific ensembles, while simultaneously achieving temporal super-resolution.
  \end{newTVCG}
  Leveraging recent deep learning advancements in optical flow estimation, \begin{newTVCG}\acronym\end{newTVCG}can perform interpolation between scalar fields of different timesteps without strictly relying on GT flow vectors during training.  This enhances its applicability to scientific ensembles, especially where GT flow data is unavailable.

  \begin{newTVCG}\acronym\end{newTVCG}implements a student-teacher architecture
  \cite{hu22:_teach_studen_archit_knowl_learn} which has several key
  advantages.  It enables more accurate estimations as it is trained
  on GT temporal scalar fields (as demonstrated in \autoref{sec:results}).
  Guidance through the teacher model further enhances the student
  model's learning process, resulting in stable training, faster
  convergence, and improved model robustness.  \begin{newTVCG}\acronym\end{newTVCG}does not require
  pre-training and intermediate fine-tuning on simplified datasets;
  it already converges on the target datasets, in contrast
  to previous comparable works in computer vision~\cite{teed2020raft,
    dosovitskiy2015flownet, luo2021upflow}. 
    \begin{newTVCG}\acronym\end{newTVCG}further yields significantly faster convergence and achieves an accuracy of more than 90\% within the initial 30\% of the total training time.
Below, we describe the neural network architecture employed in \begin{newTVCG}\acronym\end{newTVCG}
(Sec.~\ref{subsec:mehtod_learn_rec_flow}), and present the
  temporal interpolation and flow \begin{newTVCG}estimation\end{newTVCG} pipeline, both for
  the training and inference phases (\autoref{sec:method:interp}).
  Our loss function for training can flexibly
  adapt to different scenarios as described in \autoref{subsec:loss}.
  \begin{newTVCG}\acronym\end{newTVCG}builds upon RIFE~\cite{huang2022real} and modifies as well as extends it in several ways; a
  detailed comparison is presented in Sec.~\ref{subsect:methodcompare}.

\vspace{-5pt}

  \subsection{\begin{newTVCG}\acronym\end{newTVCG}Network Architecture}
\label{subsec:mehtod_learn_rec_flow}

The network architecture is a feed-forward CNN, see
Fig.~\ref{fig:model}. The student network consists of $N$ stacked
convolutional blocks (\textit{Conv Block}), each incorporating
convolutional (\textit{Conv}) and deconvolutional (\textit{Deconv})
layers (the middle column in the orange box
of~\autoref{fig:model} shows an expanded view of a convolution
block).
There are 256 feature channels in all convolutional layers of
the first block, 192 of the second and third, and 128 of the last
block.  The number of channels in the layers of the teacher block \textit{\begin{newTVCG}Conv Block$^{teach}$\end{newTVCG}} is
set to 128, similarly to the last block of the student model.  We
utilize a PReLU activation function~\cite{he2015delving} in all layers
except for the last one.  The teacher model crucially features a
dedicated Conv Block$^{teach}$ which enables it to directly
consider a GT scalar field by receiving $\DGT$ through an additional
channel.  Note that traditionally such information is not
available directly to the network architecture but is only considered
in the loss function. 
The loss components that drive the training process are shown on the right
side in Fig.~\ref{fig:model}. 
Student and teacher models are \emph{jointly}
optimized during training (i.e., following an online
scheme~\cite{hu22:_teach_studen_archit_knowl_learn}), with the teacher
model refining the student model's results.
\begin{new}
According to our experiments, this streamlined one-stage online training approach significantly reduces training time compared to two-stage optimization, where the teacher model is trained first, and a student network is subsequently trained to align with the teacher's outputs. 
As demonstrated in~\autoref{subsec:param_study}, our one-stage approach does not yield any degradation in performance while significantly reducing training time.
\end{new}
The proposed \begin{newTVCG}\acronym\end{newTVCG}model architecture serves as the foundation for
tasks with and without available GT flow fields,
differing only in the applied loss functions (\autoref{subsec:loss}).

\begin{old}
During inference time, the final estimation is
  $\Fhat^{N-1}_{t \rightarrow s}$, $\Fhat^{N-1}_{t \rightarrow u}$, and $M^{N-1}$,
  where $N=4$ (see Fig.~\ref{fig:model}).

We employ the student-teacher model approach as it presents several
key advantages over relying solely on the reconstruction loss
calculated from the available $\DGT$.  It enables more accurate
estimations as it is trained on
ground-truth \begin{old}timestep\end{old} temporal
fields (as demonstrated in \autoref{sec:results}).  Guidance through
the teacher model further enhances the student model's learning
process, resulting in stable training, faster convergence, and
improved model robustness.  \begin{newTVCG}\acronym\end{newTVCG}does not require pre-training and
intermediate fine-tuning on simplified datasets, as it achieves
convergence already on the target datasets, in contrast to previous
comparable works in computer vision~\cite{teed2020raft,
  dosovitskiy2015flownet, luo2021upflow}.  Notably, compared to these
methods, \begin{newTVCG}\acronym\end{newTVCG}demonstrates a significantly shorter convergence time,
also achieving an accuracy of more than 90\% within the initial 30\%
of the total training time.
        
We vary the number and configuration of blocks, comparing to RIFE, and
identify the optimal setup through hyperparameter search, in
Sec.~\ref{subsec:param_study}.  In addition, we replace the use of
different scales for different blocks with deconvolutional layers
instead of bilinear resize.  We also reduce computational time by
eliminating the need for an additional encoder-decoder CNN --
RefineNet~\cite{jiang2018super, niklaus2020softmax}, which would have
otherwise doubled \begin{newTVCG}\acronym\end{newTVCG}'s training time.  According to our
experiments, the addition of RefineNet does not yield any improvements
in our targeted use cases.  Furthermore, we optimize the
hyperparameters of the neural network, such as the learning rate, the
number of channels and the kernel size in the convolutional layers, as
well as the data preprocessing steps.  Crucially, our proposed model
introduces new kinds of loss functions targeted at different
scenarios, as discussed below.
\end{old}

\subsection{\begin{newTVCG}Flow Estimation and Scalar Field Interpolation\end{newTVCG}}
\label{sec:method:interp}

  As input, \begin{newTVCG}\acronym\end{newTVCG}receives two scalar fields $D_s$ and $D_{u}$ of the same ensemble member at timesteps $s<u$ and an intermediate timestep $t$, where $s < t < u$.  
  As output, \begin{newTVCG}\acronym\end{newTVCG}(1)~provides
  interpolants $\hat{D}_{t}$ at time $t\in [s,u]$ and (2)~predicts the
  corresponding optical or physical flow field $\hat{F}_{t}$.
  First, intermediate flow fields $\Fts{}$ and $\Ftu{}$ are computed.
  The \emph{time-backward} flow $\Fts{}$ refers to the intermediate
  flow field vectors from a frame at time $t$ to an earlier frame at
  time $s$, while the \emph{time-forward} flow $\Ftu{}$ is from the
  frame at time $t$ to a later frame at time $u$.
  In the process, intermediate warped scalar fields are computed and
  fused by a fusion mask $M$, as follows.

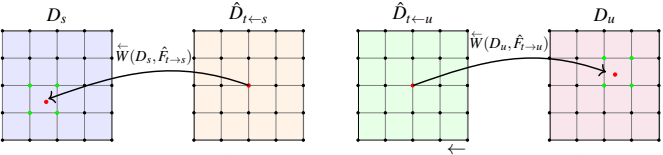
\begin{figure}[h]
  \centering
  \resizebox{\linewidth}{!}{%
  \begin{tikzpicture}

    \node[draw, rectangle, minimum width=2cm, minimum height=2cm,
    fill=blue!10, label=above:$D_s$] (Ds) at (0,0) {};

    \node[draw, rectangle, minimum width=2cm, minimum height=2cm,
    fill=orange!10, right=1.49cm of Ds, label=above:$\Dts{}$] (Dts)
    {};
    
    \node[draw, rectangle, minimum width=2cm, minimum height=2cm,
    fill=green!10, right=0.99cm of Dts, label=above:$\Dtu{}$] (Dtu) {};

    \node[draw, rectangle, minimum width=2cm, minimum height=2cm,
    fill=purple!10, right=1.49cm of Dtu, label=above:$D_u$] (Du) {};

      \draw[step=0.5cm, gray, very thin] (-1,-1) grid (1,1);
      \foreach \x in {-1,-0.5,...,1}
        \foreach \y in {-1,-0.5,...,1}
          \filldraw (\x,\y) circle (0.6pt);

        \filldraw[red] (-0.2,-0.3) circle (0.85pt);

      \draw[step=0.5cm, gray, very thin] (2.5,-1) grid (4.5,1);
      \foreach \x in {2.5,3,...,4.5}
        \foreach \y in {-1,-0.5,...,1}
          \filldraw (\x,\y) circle (0.6pt);

    \filldraw[green] (-0.5,0) circle (0.75pt);
    \filldraw[green] (-0.5,-0.5) circle (0.75pt);
    \filldraw[green] (0,0) circle (0.75pt);
    \filldraw[green] (0,-0.5) circle (0.75pt);
    \filldraw[red] (3.5,0) circle (0.85pt);

      \draw[step=0.5cm, gray, very thin] (5.5,-1) grid (7.5,1);
      \foreach \x in {5.5,6,...,7.5}
        \foreach \y in {-1,-0.5,...,1}
          \filldraw (\x,\y) circle (0.6pt);

      \draw[step=0.5cm, gray, very thin] (9,-1) grid (11,1);
      \foreach \x in {9,9.5,...,11}
        \foreach \y in {-1,-0.5,...,1}
          \filldraw (\x,\y) circle (0.6pt);

    \filldraw[green] (10,0) circle (0.75pt);
    \filldraw[green] (10,0.5) circle (0.75pt);
    \filldraw[green] (10.5,0) circle (0.75pt);
    \filldraw[green] (10.5,0.5) circle (0.75pt);
    \filldraw[red] (10.2,0.2) circle (0.85pt);

    \filldraw[red] (6.5,0) circle (0.75pt);

    \draw[->, line width=0.75pt] (Dts.center) to[out=160,in=20] node[above=0.01cm, font=\footnotesize, xshift=3pt] {$\W ( D_s, \hat{F}_{t \rightarrow s})$} ([xshift=24pt, yshift=-7pt] Ds.west);
    \draw[->, line width=0.75pt] (Dtu.center) to[out=20,in=160] node[above=0.01cm, font=\footnotesize] {$\W ( D_{u}, \hat{F}_{t \rightarrow u})$} 
    ([xshift=-28pt, yshift=7pt] Du.east);

  \end{tikzpicture}
  }
  \caption{ Illustration of backward warping $\W$:
    (scalar) fields $D_s$ and $D_u$ are reversely mapped according to
    the flow fields $\hat{F}_{t \rightarrow s}$ and $\hat{F}_{t \rightarrow u}$. 
    The fields $\Dts{}$ and $\Dtu{}$ are then reconstructed using bilinear interpolation
    considering the values at the coordinates shown in green. 
    }
  \label{fig:backward_warp}
\end{figure}

\textbf{Warping}.
In computer graphics warping means changing a source image
into a target image. In \emph{forward} warping a mapping is used to
specify where each pixel from the source image ends up in the target image; however, holes may occur in the target image. 
This can be resolved by using \emph{backward} warping.
\begin{newTVCG}
 We utilize this technique in \begin{newTVCG}\acronym\end{newTVCG}through a reverse mapping that finds, for each pixel $p_t$ in the target image, the point $p_s$ in the source image where it originated.
\end{newTVCG}
Then resampling around this point $p_s$ is applied by bilinear interpolation of source pixel values to
determine the value of the target pixel $p_t$. 
The warping operator
$\W$ denotes the combined effect of reverse mapping and bilinear
interpolation. In our case the intermediate flow fields define the
mappings, see Fig.~\ref{fig:backward_warp}. Here $\Dts{}$ and $\Dtu{}$
are target images at time $t$ with source images $D_s$ and $D_u$,
respectively. This results in \begin{newTVCG}two\end{newTVCG} warped scalar fields $\Dts{}$ and $ \Dtu{}$:
\begin{equation}
    \Dts{} = \W (D_s, \Fts{}), \quad
    \Dtu{} = \W (D_{u},\Ftu{}).
  \label{eq:warp}
\end{equation}

\vspace{-6pt}

\textbf{Fusion mask}. 
\begin{newTVCG}
The fusion mask $M$, where $M(i, j) \in [0, 1],\forall i, j$, combines two intermediate warped scalar fields $\Dts{}$ and $\Dtu{}$ at successive timesteps into an interpolated scalar field $\hat{D}_t$.
The values in $M$ are learned by \begin{newTVCG}\acronym\end{newTVCG}to ensure a smooth transition between $\Dts{}$ and $\Dtu{}$, minimizing artifacts and preserving the structural integrity of the original fields across space.
\end{newTVCG}


\textbf{Refining the intermediate flows}.
The intermediate flow fields are computed via $N$
convolutional blocks by iterative refinement, see
Fig.~\ref{fig:model}.
\begin{new}
This coarse-to-fine process iteratively and jointly refines both the flow fields and the fusion mask at each Conv Block, ensuring progressively consistent and high-quality updates throughout the network.
\end{new}
Superscripts $i$ in \begin{new}Fig.~\ref{fig:model} \end{new}denote the various
quantities at iteration $i$. Given two
input fields $D_s$ and $D_{u}$ and a timepoint $t$, with $s<t<u$,
\cb{0} computes a rough estimation of intermediate flow fields
$\Fts{0}, \Ftu{0}$ and fusion mask $M^0$ to capture large motions.
Then in $\cb{i}$, $i>0$, $D_s$ and $D_{u}$
are first backward-warped using Eq.~\eqref{eq:warp}, based on the
intermediate flows $\Fts{i-1}, \Ftu{i-1}$ and mask $M^{i-1}$ of
the previous iteration (see the yellow Warping block in
Fig.~\ref{fig:model}, middle column).
Next, $D_s$ and $D_{u}$, warped frames $\Dts{i-1}$
and $\Dtu{i-1}$, intermediate flows $\Fts{i-1}$ and $\Ftu{i-1}$, mask
$M^{i-1}$, and timestep $t$ are concatenated and processed by the
stack of convolution and deconvolution layers in \cb{i}. This
results in updated $\Fts{i}, \Ftu{i}$, and $M^{i}$, which then
enter the next Conv Block. The process continues until the last
\cb{N-1} has finished computation, producing final
estimates $\Fts{N-1}, \Ftu{N-1}$, and $M^{N-1}$.

\textbf{Interpolation and \begin{newTVCG}flow estimation\end{newTVCG}}.
Interpolated scalar field $\hat{D}_{t}$, intermediate flow fields $\Ft{i}$, and \begin{newTVCG}estimated\end{newTVCG} flow field $\Fhat_t$ are obtained via:
\vspace*{-0.3cm}
  \begin{subequations}
  \begin{align}
    \hat{D}_t &= \Dts{N-1} \odot M + \Dtu{N-1} \odot (\mathbf{I} - M)\label{eq:interpolD}\\
   \Ft{i} &= \Ftu{i}, \quad\hat{F}_t =  \Ft{N-1} \quad(N=4)\label{eq:interpolF}
\end{align}
\end{subequations}
where $\odot$ denotes element-wise multiplication, and $\mathbf{I}$ is
the identity matrix.  We determined $N=4$ as the optimal value, see
Sec.~\ref{subsec:param_study}.

\textbf{Refinement by teacher module}.
During training, a dedicated \textit{Conv $Block^{teach}$}
implements a ``privileged distillation
scheme''~\cite{lopez-paz16:_unify} in the form of a teacher module
that has access to a GT scalar field by receiving $D_{GT}$ through an
additional channel during concatenation; see Fig.~\ref{fig:model}.
This produces flow fields $\Fteachts, \Fteachtu$ and mask
$\Mteach$. By inserting these in Eq.~\eqref{eq:interpolD} and
Eq.~\eqref{eq:interpolF}, an interpolated field $\Dt{teach}$ and
\begin{newTVCG}estimated\end{newTVCG} flow field $\Ft{teach}$ are obtained for the teacher.
The outputs for student ($\Dhat_{t}, \Fhat_t$) and teacher
($\Dt{teach}, \Ft{teach}$) are used in the loss functions to compute
the prediction error which will then be backpropagated to update the
network parameters. The process is repeated for many inputs during
training (for more details, see Sec.~\ref{subsec:training}).

\textbf{Inference}.
In the inference phase, the scalar fields $D_s$ and $D_{u}$ are
processed by the \emph{trained} network, meaning
that all GT information, the teacher block, and the loss functions are
absent. The interpolated scalar field $\Dhat_{t}$ and reconstructed
flow field $\Fhat_t$ as computed by Eqs.~\eqref{eq:interpolD},
\eqref{eq:interpolF} constitute the final output, see
Fig.~\ref{fig:nn_overview}.

\subsection{Loss Function}
\label{subsec:loss}
\begin{newTVCG}\acronym\end{newTVCG}\begin{old}consists of\end{old} uses several
loss components that can be
combined depending on the scenario: (1)~with available GT flow field
and (2)~without (the components are listed in \autoref{fig:model} and described further below).
    
\textbf{(1)} The total loss for ensembles with available GT  flow field is
a linear combination of reconstruction loss $\mathcal{L}_{rec}$ and
flow loss $\mathcal{L}_{flow}$:
\begin{equation}\label{eq:loss-with-flow}
  \begin{aligned}
    \mathcal{L} = \mathcal{L}_{rec} + \lambda_{flow} \,\mathcal{L}_{flow},
  \end{aligned}
\end{equation}
where $\lambda_{flow} = 0.2$ for balancing total loss scale w.r.t.\
the reconstruction component (determined via hyperparameter search, see \autoref{subsec:param_study}).

\textbf{(2)} The total loss of \begin{newTVCG}\acronym\end{newTVCG}for ensembles without available
flow fields is a linear combination of the reconstruction
$\mathcal{L}_{rec}$, distillation $\mathcal{L}_{dis}$, photometric
$\mathcal{L}_{photo}$, and regularization $\mathcal{L}_{reg}$ losses:
\begin{equation}\label{eq:loss-without-flow}
  \begin{aligned}
    \mathcal{L} = \mathcal{L}_{rec} + \lambda_{dis}\, \mathcal{L}_{dis} 
    + \lambda_{photo}\, \mathcal{L}_{photo} + \lambda_{reg}\, \mathcal{L}_{reg} ,
  \end{aligned}
\end{equation}
where
$\lambda_{dis} = 10^{-4}, \lambda_{photo} = 10^{-6}, \lambda_{reg} =
10^{-8}$ for balancing total loss scale w.r.t.\ other loss components
(see Sec.~\ref{subsec:param_study}).

\textbf{\begin{newTVCG}Scalar field interpolation\end{newTVCG} (flow-unsupervised)}.  To temporally interpolate
between fields, we utilize student and
teacher blocks
(Sec.~\ref{subsec:mehtod_learn_rec_flow}), incorporating loss
components aimed at improving the accuracy of the interpolated density
from~Eq.~\eqref{eq:interpolD}.	
The reconstruction loss $\Lrec$ measures
the $L_1$ distance between the GT $\DGT$ and the reconstructed field
representation from both the student and teacher:
\begin{equation}
  \label{eq:Lrec}
  \begin{aligned}
    \mathcal{L}_{rec} = \lVert \DGT - \hat{D}_{t} \rVert_{1}  + \lVert \DGT - \hat{D}_{t}^{teach}\rVert_{1}.
  \end{aligned}
\end{equation}

  The first term in Eq.~\eqref{eq:Lrec} is evaluated using
  $\DGT$ but only for the loss calculation; $\DGT$ is never used as input to the student network, only to the teacher module.

\textbf{Physical flow \begin{newTVCG}estimation\end{newTVCG} (flow-supervised)}.
When we have access to GT flow information during
training, we incorporate a flow loss component to enhance the quality
of the learned flow field (physical flow).  This loss function
comprises reconstruction and teacher components, as well as the flow
loss component.  The supervised flow loss measures the $L_1$ distance
between the estimated flow from each block of the neural network and
the GT flow.  In our experiments, we found that accumulating this
measure based on all blocks rather than just the last one yields
better results.  We also adopt the concept of exponentially increasing
weights from RAFT~\cite{teed2020raft}.  This loss equation for
physical flow \begin{newTVCG}estimation\end{newTVCG} can be expressed as follows:
\begin{equation}
  \mathcal{L}_{\text{flow}} = 
  \sum_{i=1}^{N} \gamma^{N-i} \lVert \FGT - \hat{F}^{i}_{t} \rVert_{1} + \lVert \FGT - \hat{F}_{t}^{teach}\rVert_{1},
\end{equation}
where $\FGT$ is the GT flow at time $t$,
$\hat{F}^{i}_{t}$ is the flow output from the
corresponding $i^{th}$ block of the student
network (Eq.~\eqref{eq:interpolF}), $N=4$ is the
number of blocks in the model, and $\hat{F}_{t}^{teach}$ is the flow
output from the teacher block.  We experimentally established the value of $\gamma$
as 0.8, aligning with the RAFT loss and validating
this choice through our hyperparameter search.

  Again, $\FGT$ is only used for calculation of the error;
  it is never used as input to the student-teacher network.

\textbf{Optical flow \begin{newTVCG}estimation\end{newTVCG} (flow-unsupervised)}.
When the flow field is not available for model training, we add
distillation~\cite{huang2022real}, photometric~\cite{yu2016back}, and
regularization loss components as conceptual replacements of the
supervised physical flow loss.  These components help the model to
reconstruct the desired timesteps based on the optical flow field,
which in this case is learned in a flow-unsupervised mode.
	
The distillation loss is based on the fact that the model outputs more
accurate flow when it receives the GT timesteps of the different field
which is available (e.g., density).  It is computed as $L_2$ distance
between the intermediate flows
\begin{old}$\hat{F}_{t \rightarrow s}, \hat{F}_{t \rightarrow u}$\end{old} of the teacher
and student networks:
\begin{equation}
  \begin{aligned}
    \mathcal{L}_{dis} =
    \sum_{j \in\{s, u\}}  \lVert \hat{F}^{N-1}_{t \rightarrow j}  - \hat{F}_{t \rightarrow j}^{teach}\rVert_{2} ,
  \end{aligned}
\end{equation}
	
\begin{old}
  The photometric loss in our experiments is the error between a field
  of the first timestep and the backward warping of a field of the
  second timestep, which was interpolated based on the estimated flow.
\end{old}
We also conducted experiments incorporating the smoothness loss
component, commonly used in optical flow \begin{newTVCG}estimation\end{newTVCG}
tasks~\cite{yu2016back}. However, in our case, where there are
generally no clearly distinguishable objects within our ensembles, the
smoothness loss did not improve the results. The metric results
obtained by incorporating the smoothness loss component are reported
in Table~\ref{hyperparameter_search} under the ``\begin{newTVCG}\acronym\end{newTVCG}\textit{smooth}'' row.
The Charbonnier penalty function is used in both photometric and
smoothness loss components to provide robustness against
outliers~\cite{sun2014quantitative}.
The photometric loss is
computed as the difference between the fields $D_s, D_{u}$, and the
\begin{old}backward/inverse warped field of the \end{old}reconstructed
field $\hat{D}_{t}$:
\begin{equation}
  \mathcal{L}_{photo}(\vec{v}; D_j, \hat{D}_{t}) = \
  \frac{1}{2} \sum_{j \in\{s, u\}} \sum_{\vec{p} \in P} \rho(D_j(\vec{p}) - \hat{D}_{t}(\vec{p}+\vec{v}(\vec{p})),
\end{equation}
where $\vec{v}$ are the components of the estimated optical flow field
and $\rho(x) = \sqrt{x^2 + \epsilon^2}$ is the Charbonnier penalty
function with $\epsilon=10^{-9}$.  The summation
$\sum_{\vec{p} \in P}$ indicates that the loss is computed over all
pixel coordinates $\vec{p} \in P \subset \mathbb{N}^d$ \begin{old}for\end{old}
of the domain in a $d-$dimensional space, and the
expression $\vec{p}+\vec{v}(\vec{p})$ represents the updated pixel
coordinates after applying the estimated flow vectors $\vec{v}$.
\begin{old}These flow vectors are used to warp the coordinates and
  sample the corresponding values from $\hat{D}_{t}$.  \end{old}We
apply the photometric loss component to the last block of \begin{newTVCG}\acronym,\end{newTVCG} i.e.,
$Conv Block^{N-1}$ in Fig.~\ref{fig:model}.

Additionally, we apply $L_1$ regularization~\cite{bishop2006pattern}
to the weight matrix of the last convolutional block of the student
\begin{old}network as well as to the convolutional block of the
  teacher network\end{old} and teacher networks to prevent
overfitting of the flow field learning:
\begin{equation}
  \begin{aligned}
    \mathcal{L}_{reg} = \sum_{i=1}^{L} \left( \lVert W_i \rVert_{1} + \lVert W_i^{teach} \rVert_{1} \right),
  \end{aligned}
\end{equation}
\begin{old}where $L$ represents the number of weights in the weight
  matrix of the last convolutional block of the student network as
  well as in the convolutional block of the teacher network, $W_i$ and
  $W_i^{teach}$ represent the $i^{th}$ weight in the student model's
  weight matrix and the teacher model's weight matrix, respectively.
\end{old} 
where $W_i$ and $W_i^{teach}$ represent the $i^{th}$ weight
  matrix of the last convolutional block in the student and teacher
  network, respectively, with $L$ denoting the total number of weights.

	

 \vspace{-10pt}
  \subsection{Comparison between \begin{newTVCG}\acronym\end{newTVCG}and RIFE methods}
  \label{subsect:methodcompare}
  We now compare the architectures of \begin{newTVCG}\acronym\end{newTVCG}and RIFE in detail (their respective performances are discussed in \autoref{sec:results} and \autoref{sec:compar_eval}).

  \smallskip\textbf{Similarities.} 
  RIFE and \acronym both receive two scalar input fields and employ several convolution blocks to produce approximate intermediate flows and a fusion mask.
  Warped fields are computed from the input frames with the help of the intermediate flows.
  Merging the two warped frames with the help of the fusion mask via Eq.~\eqref{eq:interpolD} yields the interpolated field. 
  RIFE and \begin{newTVCG}\acronym\end{newTVCG}feature an online end-to-end trainable student-teacher architecture using a special teacher convolution block. To calculate the model residual error, both use an optimization that minimizes a loss function, consisting of various terms.
  
  \smallskip\textbf{Improvements and Extensions.}
  In RIFE the number of student convolution blocks is fixed to three, while in \begin{newTVCG}\acronym\end{newTVCG}we can flexibly adapt the number and configuration of blocks \begin{newTVCG}for the most optimal performance.\end{newTVCG}
  Furthermore, RIFE's use of different scales for different blocks with bilinear resize is replaced by convolution and deconvolutional layers, resulting in more learnable parameters. 
  Additionally, \begin{newTVCG}\acronym\end{newTVCG}comprises convolutional and deconvolutional layers with varying strides in each block, unlike RIFE's eight convolutional layers with a stride of one per block.
  \begin{newTVCG}
      This autoencoder-like design effectively captures key input features~\cite{bengio2013representation}, as demonstrated by the performance gains shown in Table~\ref{hyperparameter_search}.
  \end{newTVCG}

  Crucially, \begin{newTVCG}\acronym\end{newTVCG}features new kinds of loss functions for handling different scenarios. When no flow fields are available (case \textbf{(2)} in Sec.~\ref{subsec:loss}),
  \begin{newTVCG}\acronym\end{newTVCG}uses additional loss functions, such as photometric
  loss~\cite{yu2016back} (RIFE has only the first two terms
  in Eq.~\eqref{eq:loss-without-flow}). In contrast to RIFE, \begin{newTVCG}\acronym\end{newTVCG}can
  utilize available GT flow fields during training (case
  \textbf{(1)}) to improve
  \begin{newTVCG}estimated\end{newTVCG} flow fields quality during inference.
  
  In RIFE, the intermediate flow fields are only used to obtain the
  interpolated video frames which appear blurred and discontinuous
  (see \cite[Fig.~5]{huang2022real}), especially when the teacher
  module is omitted.
  In \begin{newTVCG}\acronym\end{newTVCG}we achieve flow fields of high quality so they can serve as meaningful
  \begin{newTVCG}
  supplements for spatiotemporal data analysis.
  \end{newTVCG}
  \begin{old}
  enrichments (see Sec.~\ref{subsect:ablation}).
  \end{old}
  \begin{newTVCG}\acronym\end{newTVCG}further exhibits significantly lower computational cost: eliminating the need for
  RIFE's encoder-decoder CNN (RefineNet~\cite{jiang2018super, niklaus2020softmax}) allowed \begin{newTVCG}\acronym\end{newTVCG}to halve its training time without impacting result quality according to our experiments.
%
  Last but not least, while RIFE is limited to 2D image sequences,
  \begin{newTVCG}\acronym\end{newTVCG}can handle arbitrary 2D scalar fields as well as 3D+time data to open up new opportunities for analysis in scientific visualization.
  \begin{newTVCG}
  To enable this, we implemented a new 3D warping technique, integrated 3D convolutional and deconvolutional layers, and utilized 3D loss functions.
  \end{newTVCG}

 \vspace{-6pt}
\section{Study Setup}
\label{sec:study_setup}
	
In this section, we describe the training setup, provide an overview of the datasets used in our experiments and discuss the evaluation methods employed to assess the \acronym results.

 \vspace{-10pt}
\subsection{Training}
\label{subsec:training}
We apply the standard prepossessing step of normalization to [0, 1]
when necessary before the start of the training.  \begin{newTVCG}\acronym\end{newTVCG}is optimized
using AdamW~\cite{loshchilov2018fixing} --- an adaptive
  gradient descent method with weight decay used in back-propagation
  algorithms for training feed-forward neural networks that
combines the benefits from both the Adam
optimizer~\cite{kingma2014adam} and $L_2$-regularization.  We employ
early stopping with a patience parameter of 30, which is equivalent to regularization~\cite{goodfellow2016deep} and helps to
prevent overfitting on the training data.  We use an
experimentally determined learning rate of $6\times10^{-4}$  for the 2D case and $1\times10^{-4}$ for the 3D case respectively 
with a cosine annealing scheduler that gradually decreases the learning rate
to $6\times10^{-6}$  and $1\times10^{-6}$ respectively by the end of the training.
We train \begin{newTVCG}\acronym\end{newTVCG}with mini-batches of size 32 for both \begin{old}the LBS and
Droplets ensembles\end{old}
  2D ensemble datasets and mini-batches of size 2 for 3D datasets
  used in our study (see Sec.~\ref{subsec:data}).
We split the set of all available data into training,
validation (for monitoring training progress), and
test subsets.

  Note that training and test data are obtained by subsampling the
  original dataset, which is assumed to provide the GT
  scalar fields. Then \begin{newTVCG}\acronym\end{newTVCG}performs interpolation at missing timesteps.

To support arbitrary interpolation, we chose $t\in [s,u]$ randomly at the training stage.  
Throughout this work, we use a maximum time
window of size 12 to sample the triplets $D_s, D_t, D_u$
for the training set, which was determined via
hyperparameter search, to support arbitrary interpolation and
\begin{newTVCG}flow estimation\end{newTVCG} during the training stage.
This window determines the maximum time gap between the timesteps that are used for interpolation (i.e., timesteps $s$ and $u$).  
Our proposed \begin{newTVCG}\acronym\end{newTVCG}model is trained for 120 epochs, \begin{newTVCG}resulting in a trained model size of 79MB.\end{newTVCG} 
\begin{newTVCG}
  The training process takes no more than 12 hours on a
single Nvidia Titan V GPU with 12GB of VRAM to converge for all datasets.
\end{newTVCG}
        
\vspace{-10pt}
\subsection{Datasets}
\label{subsec:data}
We consider \begin{newTVCG}four\end{newTVCG} scientific datasets in our study.

\begin{newTVCG}%
\textbf{LBS (flow-supervised).}
\end{newTVCG}%
The first is generated by a Lattice Boltzmann Simulation~\cite{mocz2020lbs}.  This ensemble is similar to a classic
K\'arm\'an vortex street simulation and yields density as well as flow
fields with a spatial resolution of $100\times 400$.
We consider an ensemble comprising 21 members, each consisting of 3K
timesteps, with varying cylinder size, position, radius, collision
timescale, and the dynamic viscosity of the fluid within the
simulation.
We utilize different members of the ensemble for validation and
testing of \begin{newTVCG}\acronym.\end{newTVCG}
We randomly sample a training subset of 40K timesteps from a 12-sized window in the case of the LBS ensemble as discussed above.

\begin{newTVCG}%
\textbf{Droplets (flow-unsupervised).}
\end{newTVCG}%
The second is a Drop Dynamics (Droplets) ensemble derived from a physical experiment investigating the impact of a falling droplet on a film~\cite{geppert2016classification}.
The experiment employs shadowgraphy imaging to study the splash crown and secondary droplets formed after the primary droplet's impact.
Shadowgraphs reflect changes in the second derivative in density, which lead to variations in the refractive index of the medium and can therefore be detected optically.
The experimental dataset consists of monochrome videos with a spatial resolution of $160\times 224$, totaling 135K timesteps from 1K ensemble members.
This dataset was collected to analyze various droplet impact regimes in relation to parameters like fluid viscosity, droplet velocity, film thickness, and Weber number.
We interpret changes in luminosity in the video images as variations in the density field.
Since no flow information is available, this constitutes a flow-unsupervised setup. 
We use different sets of 3K timesteps for validation and testing of \begin{newTVCG}\acronym\end{newTVCG}and a randomly sampled subset of 40K timesteps for training.

\begin{newTVCG}%
\textbf{5Jets (flow-supervised).}
\end{newTVCG}%
The third one is a 3D+time 5Jets dataset generated by a Navier Stoke flow solver originally obtained from \href{https://vis.ucdavis.edu/}{UC Davis}.
This simulation models five jets entering a cubical region and consists of 2000 timesteps with a spatial resolution of $128\times 128\times 128$.
It contains density and the $x$, $y$, $z$ components of velocity.
We utilize every $10^{th}$ timestep of the simulation and randomly sample a subset of 500 timesteps for training, validation, and testing.

\begin{newTVCG}%
\textbf{Nyx (flow-supervised).}
\end{newTVCG}%
\begin{newTVCG}
The fourth dataset is a 3D+time ensemble based on the compressible cosmological hydrodynamics simulation Nyx, developed by Lawrence Berkeley National Laboratory~\cite{sexton2021nyx}.
We consider an ensemble comprising 18 members, each consisting of a maximum of 1600 timesteps with a spatial resolution of $128\times 128\times 128$.
It contains density and the $x$, $y$, $z$ components of velocity.
Akin to InSituNet~\cite{he2019insitunet}, we vary three parameters for ensemble generation: the total matter density ($\Omega_m \in [0.12, 0.155]$), the total density of baryons ($\Omega_b \in [0.0215, 0.0235]$), and the Hubble constant ($h \in [0.55, 0.7]$).
We randomly sample a training subset of 500 timesteps \begin{new}after selecting every 5th timestep across\end{new} different ensemble members for training, validation, and testing.
\end{newTVCG}

To visualize a simulation ensemble field,
we use the \textit{Turbo}
colormap~\cite{turbo}.  
We visualize the 2D flow field via arrow glyphs and 3D flow via volume rendering with a transfer function that indicates flow direction.
Examples of the LBS and Droplets ensemble fields, 
\begin{newTVCG}
    as well as 5Jets and Nyx transfer functions,
\end{newTVCG}
are in the supplementary material.

\begin{figure*}[!t]
    \centering
    \begin{tabular}{c c}
        \begin{minipage}[t]{0.08\linewidth}
            \vspace*{-196pt}
            \text{\footnotesize Dens GT} \\ [4pt]
            \text{\footnotesize Dens \begin{newTVCG}\acronym\end{newTVCG}} \\ [4pt]
            \text{\footnotesize Dens diff $\times100$} \\ [4pt]
            \text{\footnotesize Dens RIFE} \\ [4pt]
            \text{\footnotesize Dens diff $\times100$} \\ [4pt]
            \text{\footnotesize Flow GT} \\ [4pt]
            \text{\footnotesize Flow \begin{newTVCG}\acronym\end{newTVCG}} \\ [4pt]
            \text{\footnotesize Flow diff} \\ [5pt]
            \text{\footnotesize Flow RIFE} \\ [5pt]
            \text{\footnotesize Flow diff RIFE} \\ [5pt]
            \text{\footnotesize \begin{new}Pathline GT\end{new}} \\ [5pt]
            \text{\footnotesize \begin{new}Pathline \acronym\end{new}} \\
        \end{minipage} &
        \begin{minipage}[t]{0.90\linewidth}
            \begin{tikzpicture}
                \node[anchor=south west, inner sep=0] (image) at (0,0) 
                {\includegraphics[trim=365 20 335 20, 
                clip,width=\linewidth]{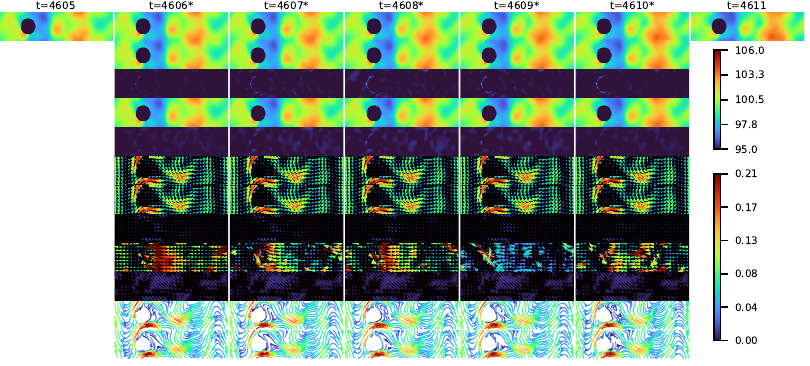}};
            \end{tikzpicture}
        \end{minipage} \\
        \multicolumn{2}{c}{\small (a) Interpolation rate of 6$\times$ (between $s=4605$ and $u=4611$).} \\
        
        \begin{minipage}[t]{0.08\linewidth}
            \vspace*{-196pt}
            \text{\footnotesize Dens GT} \\ [4pt]
            \text{\footnotesize Dens \begin{newTVCG}\acronym\end{newTVCG}} \\ [4pt]
            \text{\footnotesize Dens diff $\times100$} \\ [4pt]
            \text{\footnotesize Dens RIFE} \\ [4pt]
            \text{\footnotesize Dens diff $\times100$} \\ [4pt]
            \text{\footnotesize Flow GT} \\ [4pt]
            \text{\footnotesize Flow \begin{newTVCG}\acronym\end{newTVCG}} \\ [4pt]
            \text{\footnotesize Flow diff} \\ [5pt]
            \text{\footnotesize Flow RIFE} \\ [5pt]
            \text{\footnotesize Flow diff RIFE} \\ [5pt]
            \text{\footnotesize \begin{new}Pathline GT\end{new}} \\ [5pt]
            \text{\footnotesize \begin{new}Pathline \acronym\end{new}} \\
        \end{minipage} &
        \begin{minipage}[t]{0.90\linewidth}
            \begin{tikzpicture}
                \node[anchor=south west, inner sep=0] (image) at (0,0) 
                {\includegraphics[trim=365 20 335 20, 
                clip,width=\linewidth]{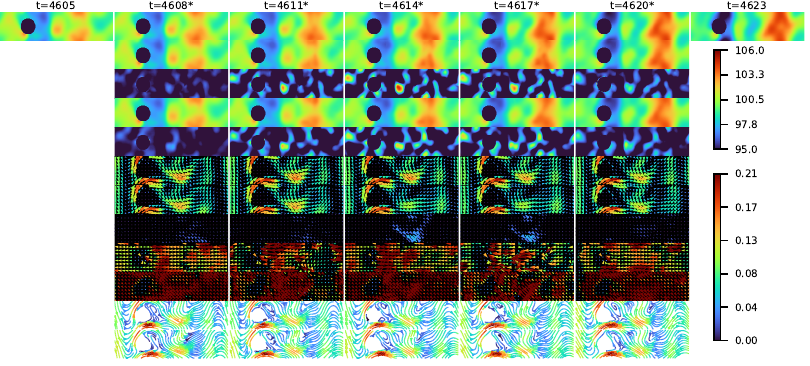}};
            \end{tikzpicture}
        \end{minipage} \\
        \multicolumn{2}{c}{\small (b) Interpolation rate of 18$\times$ (between $s=4605$ and $u=4623$).} \\
    \end{tabular}

    \caption{LBS ensemble: \begin{newTVCG}\acronym\end{newTVCG}flow field estimation and temporal density interpolation at the timesteps with an asterisk (*)---(a)~6$\times$ and (b)~18$\times$ interpolation. Each subfigure, from top to bottom shows GT density, \begin{newTVCG}\acronym\end{newTVCG}interpolated density, difference with GT density (magnified by $\times 100$), RIFE interpolated density, difference with GT density (magnified by $\times 100$); GT flow, \begin{newTVCG}\acronym\end{newTVCG}flow estimation, difference with GT flow, flow estimated by RIFE, difference with GT flow, 
    \begin{new} GT pathlines, and \acronym pathlines\end{new}. 
    The colorbar on the top right maps density, and the one on the bottom right maps flow magnitude.
    \vspace{-12pt}
    }
    \label{fig:lbs2d}
\end{figure*}

\vspace{-10pt}
\subsection{Evaluation}
We evaluate the \begin{old}quality of the reconstructed scalar
  field\end{old} performance of \begin{newTVCG}\acronym\end{newTVCG}for reconstructing
  the scalar field both qualitatively and quantitatively.
For quantitative evaluation, we utilize two different metrics:
\textit{peak signal-to-noise ratio} (PSNR) and \textit{learned
  perceptual image patch similarity}
(LPIPS)~\cite{zhang2018unreasonable}. 
LPIPS measures similarity between the activations of two images using a pre-defined network, a lower score indicates greater perceptual similarity.

\begin{old}
We assess the accuracy of the learned flow field quantitatively in the case
of available GT flow.  
In this case, we utilize the \textit{endpoint error} (EPE) which measures
the average Euclidean distance between the estimated flow vectors and
the GT flow vectors.  
A lower value indicates more accurate flow estimation.
\end{old}
We evaluate the accuracy of the learned flow field using the \textit{endpoint error} (EPE), which measures the average Euclidean distance between estimated and GT flow vectors---a lower EPE indicates higher accuracy. 
For qualitative assessment, we visualize the flow field outcomes and analyze difference plots for both simulation and experimental ensembles.
In our evaluation, we consider different interpolation (or subsampling) rates, where a rate of $x$ means that we only consider each $x^{th}$ timestep of the original data as input.

\begin{newTVCG}
For the 3D datasets, we utilize 3D PSNR and 3D EPE in the volume domain rather than the image domain.
This ensures that the evaluation metrics are appropriately adapted to the spatial characteristics of the 3D data, providing a more accurate assessment of the performance and accuracy of \begin{newTVCG}\acronym\end{newTVCG}in handling volumetric datasets.
\end{newTVCG}


%

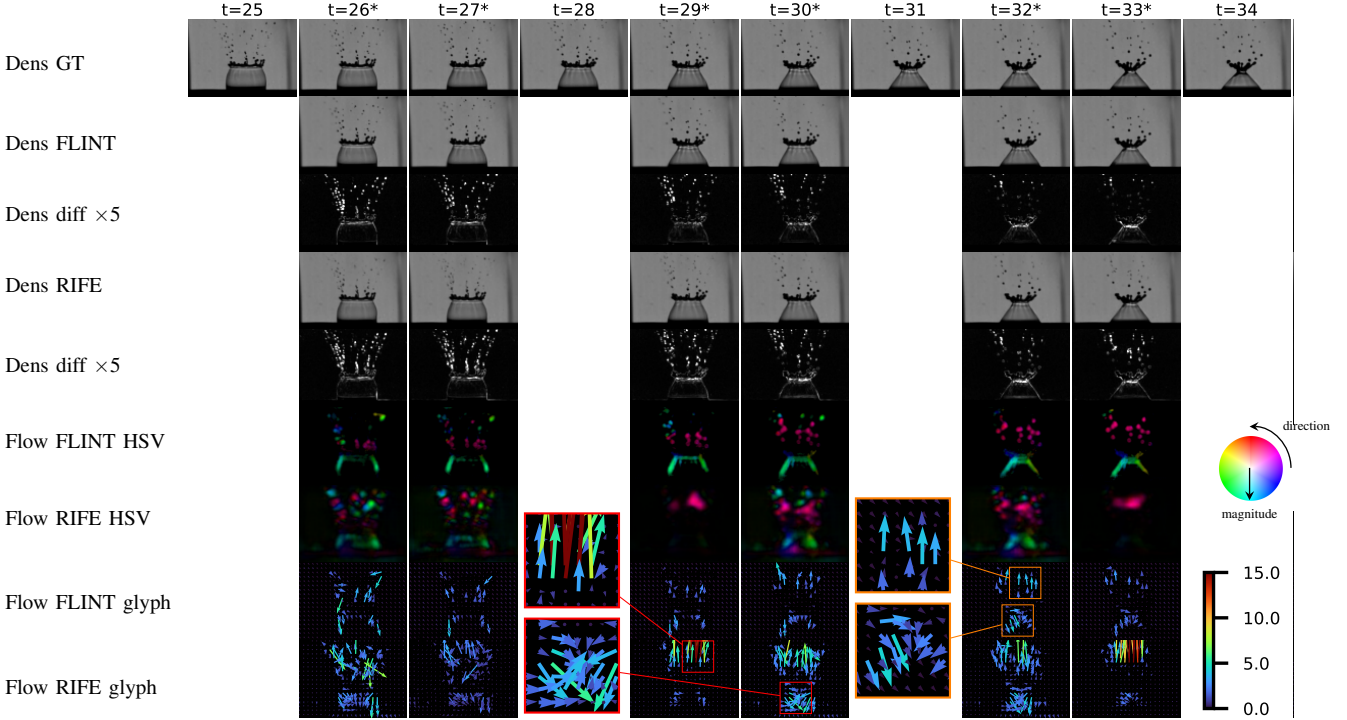
\begin{figure*}[!t]
    \centering
    \begin{minipage}[t]{0.10\linewidth}
        \vspace*{-255pt}
        \text{\footnotesize Dens GT} \\ [18pt]
        \text{\footnotesize Dens \begin{newTVCG}\acronym\end{newTVCG}} \\ [15pt]
        \text{\footnotesize Dens diff $\times5$} \\ [15pt]
        \text{\footnotesize Dens RIFE} \\ [18pt]
        \text{\footnotesize Dens diff $\times5$} \\ [17pt]
        \text{\footnotesize Flow \begin{newTVCG}\acronym\end{newTVCG}HSV} \\ [17pt]
        \text{\footnotesize Flow RIFE HSV} \\ [20pt]
        \text{\footnotesize Flow \begin{newTVCG}\acronym\end{newTVCG}glyph} \\ [20pt]
        \text{\footnotesize Flow RIFE glyph} \\
    \end{minipage}
    \hspace*{3pt}
    \begin{minipage}[t]{0.88\linewidth}
    \begin{tikzpicture}[spy using outlines={rectangle, magnification=3, size=2cm, connect spies}]
        \node (mainfig) at (0,0) {
            \includegraphics[trim=240 23 583 21, clip, width=0.93\linewidth]{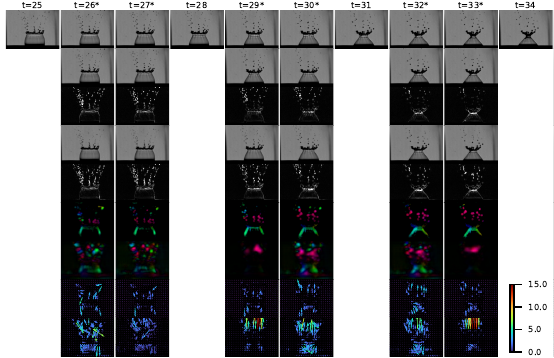}
        };

        \coordinate (rectStart) at (11.4cm, -7.9cm); 
        \coordinate (rectEnd) at (10.4cm, -7.4cm); 
        \spy [orange, draw, size=1.25cm, magnification=3, connect spies] on ($(mainfig.north west)+(rectStart)$) in node [left] at ($(mainfig.north west)+(rectEnd)$);

        \coordinate (rectStart2) at (11.3cm, -8.4cm); 
        \coordinate (rectEnd2) at (10.4cm, -8.8cm); 
        \spy [orange, draw, size=1.25cm, magnification=3, connect spies] on ($(mainfig.north west)+(rectStart2)$) in node [left] at ($(mainfig.north west)+(rectEnd2)$);

        \coordinate (rectStart3) at (7.05cm, -8.88cm); 
        \coordinate (rectEnd3) at (6.0cm, -7.6cm); 
        \spy [red, draw, size=1.25cm, magnification=3, connect spies] on ($(mainfig.north west)+(rectStart3)$) in node [left] at ($(mainfig.north west)+(rectEnd3)$);

        \coordinate (rectStart4) at (8.35cm, -9.43cm); 
        \coordinate (rectEnd4) at (6.0cm, -9.0cm); 
        \spy [red, draw, size=1.25cm, magnification=3, connect spies] on ($(mainfig.north west)+(rectStart4)$) in node [left] at ($(mainfig.north west)+(rectEnd4)$);
        
        \node[anchor=north east, inner sep=0pt] (colorwheel) at ($(mainfig.north east)+(0.05cm,-5.8cm)$) {
            \includegraphics[width=1.5cm]{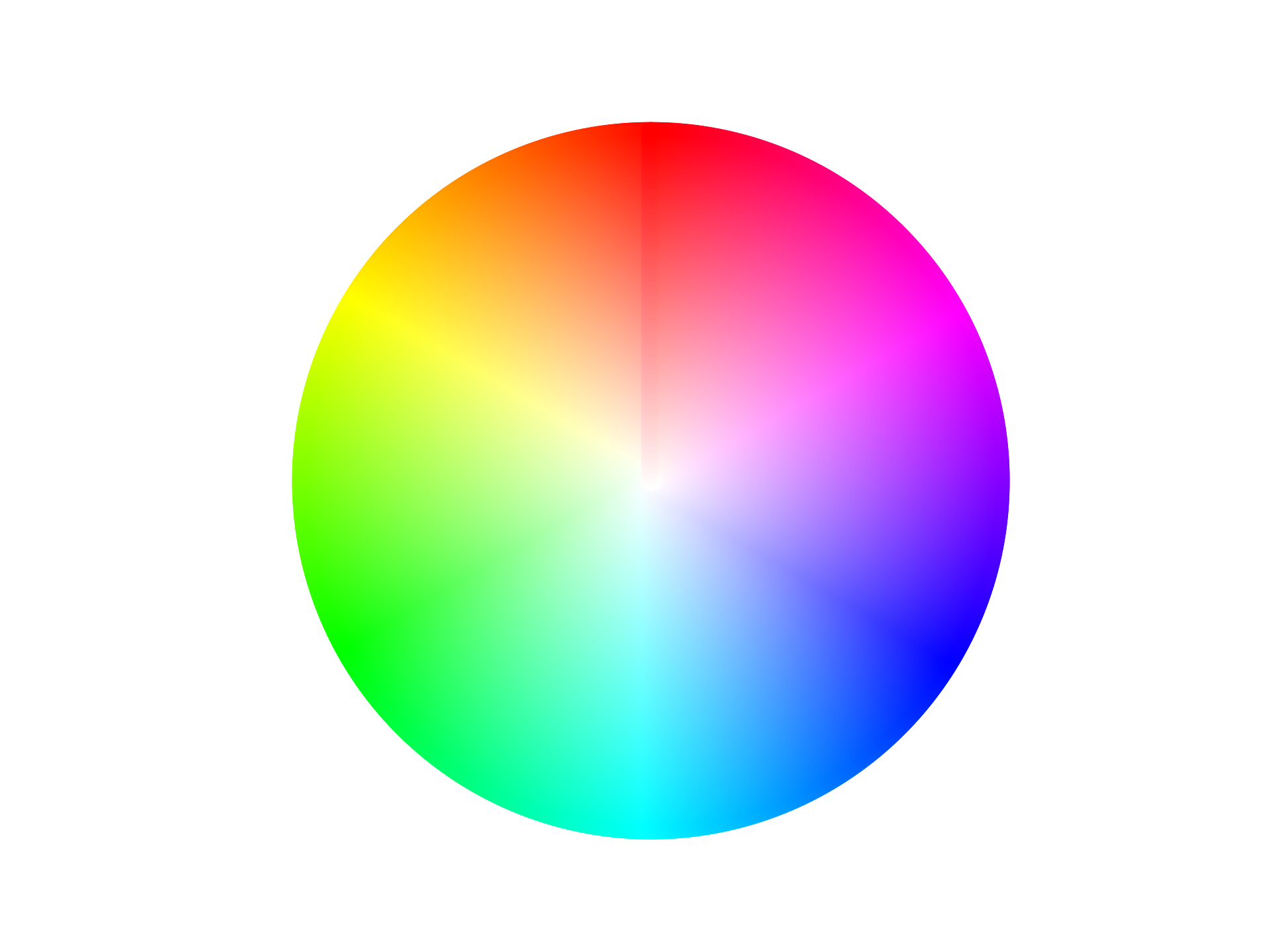}
        };

        \draw[-stealth, line width=0.5pt] ($(colorwheel)+(0.55cm,0)$) arc (0:90:0.55cm);
        \node[font=\tiny] at ($(mainfig.north east)+(0.05cm,-5.8cm)$) {direction};

        \draw[-stealth, line width=0.5pt] ($(colorwheel)$) -- ++(-90:0.43cm);
        \node[font=\tiny] at ($(colorwheel.south)+(-0.0cm,-0.05cm)$) {magnitude};
    \end{tikzpicture}
    \end{minipage}
    \caption{Droplets ensemble: \begin{newTVCG}\acronym\end{newTVCG}flow field \begin{newTVCG}estimation\end{newTVCG} and
      temporal density interpolation during inference---at the
      timesteps with an asterisk (*)---3$\times$ interpolation. 
      From top to bottom, the rows show GT density,
      \begin{newTVCG}\acronym\end{newTVCG}interpolated density, difference to
      GT density (magnified by $\times 5$),
      RIFE interpolated density, difference to
      GT density (magnified by $\times 5$),
      \begin{newTVCG}\acronym\end{newTVCG}flow estimation in HSV (see bottom right), RIFE
      flow estimation in HSV, flow glyphs for FLINT, and flow glyphs for RIFE.
      \vspace{-12pt}
      }
    \label{fig:drop2d}
\end{figure*}

\vspace{-10pt}
\section{Qualitative Results}
\label{sec:results}
\begin{newTVCG}
We evaluate \acronym via four different datasets (\autoref{subsec:data}) in 
different scenarios regarding flow \begin{newTVCG}estimation\end{newTVCG} (i.e., supplementing density with flow in both flow-supervised and flow-unsupervised cases) as well as density \begin{newTVCG}interpolation\end{newTVCG} (i.e., temporal super-resolution).
\end{newTVCG}
%
%
%

\vspace{-10pt}
\subsection{LBS: Flow Field Available for Some Members}
\label{subsec:physical_flow}
\begin{newTVCG}First,\end{newTVCG} we focus on the scenario in which we
have GT flow fields available for some ensemble members (i.e., flow-supervised learning
scenario).  
As can be seen in Fig.~\ref{fig:lbs2d} overall, our approach achieves accurate performance in terms of density field interpolation and crucially flow field \begin{newTVCG}estimation\end{newTVCG}.
In this example, results were obtained at the interpolation rates of \begin{old}$6\times$ and $16\times$\end{old}6 and 18.
\begin{old}Across both,\end{old}For both cases, the error between the reconstructed density field and its GT is \begin{old}negligible\end{old} very small (note the magnification by a factor of 100 in the \begin{newTVCG}\acronym\end{newTVCG}difference plots, third row).
Moreover, it shows that the model is able to effectively learn a flow field that closely resembles \begin{old}physical\end{old} simulated flow, resulting in a low error when compared to GT data.
Comparing our
\begin{newTVCG}\acronym\end{newTVCG}method to RIFE, we can see that \begin{newTVCG}interpolation\end{newTVCG} of the density
field worked well for both, however, the flow learned by RIFE is not
accurate and is far away structurally from the GT
flow. (See Sec.~\ref{subsect:methodcompare} for a summary
  of the differences between the \begin{newTVCG}\acronym\end{newTVCG}and RIFE methods.)
Upon closer examination of the density difference error between \begin{newTVCG}\acronym\end{newTVCG}and GT in Fig.~\ref{fig:lbs2d}b, it becomes apparent that the highest error occurs in the middle of the
frame.
This aligns with an observed error between the GT flow and \begin{newTVCG}\acronym\end{newTVCG}flow estimation in the same region.
This observation suggests that as the model begins to produce flow with a certain degree of error in this example, the density also experiences a similar error.

In general, going from an interpolation rate of
6~(Fig.~\ref{fig:lbs2d}a)
to 18~(Fig.~\ref{fig:lbs2d}b),
the density \begin{newTVCG}interpolation\end{newTVCG} remains quite accurate (differences are only visible due to significant \begin{old}scaling\end{old}contrast enhancement).  
Naturally, there are more errors in the flow
estimations of both \begin{newTVCG}\acronym\end{newTVCG}and RIFE when increasing the interpolation rate, but \begin{newTVCG}\acronym\end{newTVCG}is still able to maintain
comparably high accuracy.  
Further examples including different
interpolation rates can be found in the supplementary material.
\begin{new}
This includes evaluations at very high interpolation rates, such as $32\times$, where the quality of flow estimation begins to degrade due to the significant structural differences between widely separated timesteps. Despite this, the scalar field interpolation remains relatively robust, demonstrating \acronym's ability to handle challenging conditions effectively.
\end{new}

Additionally, we perform a 
\begin{new}pathline \end{new}analysis to further assess the accuracy of the flow estimated by \acronym. 
\begin{new}Pathlines \end{new}were generated to visualize the flow field. Results for \begin{newTVCG}\acronym\end{newTVCG}are shown in the last row and GT \begin{new}pathlines \end{new}in the 
last but one row in Fig.~\ref{fig:lbs2d}.
It can be seen that the \begin{new}pathline\end{new}
patterns produced by \acronym align closely with the GT for an interpolation rate of 6~(Fig.~\ref{fig:lbs2d}a), indicating a high level of accuracy in the flow field estimation. 
Even at an increased interpolation rate of 18~(Fig.~\ref{fig:lbs2d}b), the \begin{new}pathline \end{new}patterns closely reflect the GT, although some small deviations become noticeable~(e.g., for \begin{new}$t=4608$,\end{new} $t=4614$, and $t=4617$). 
These findings further validate the robustness of \begin{newTVCG}\acronym\end{newTVCG}in capturing complex flow dynamics across varying interpolation rates, demonstrating the utility of reconstructing this otherwise missing flow information for analyzing flow around a cylinder.

\begin{figure*}[!t]
    \centering
    \begin{minipage}[t]{0.08\linewidth}
        \vspace*{-285pt}
        \text{\footnotesize Dens GT} \\ [48pt]
        \text{\footnotesize Dens \begin{newTVCG}\acronym\end{newTVCG}} \\ [48pt]
        \text{\footnotesize Dens CoordNet} \\ [48pt]
        \text{\footnotesize Flow GT} \\ [48pt]
        \text{\footnotesize Flow \begin{newTVCG}\acronym\end{newTVCG}} \\
    \end{minipage}
    \hspace*{3pt}
    \begin{minipage}[t]{0.88\linewidth}
        \begin{tikzpicture}
            \node (mainfig) at (0,0) {
                \includegraphics[trim=428 70 400 59, clip, width=0.93\linewidth]{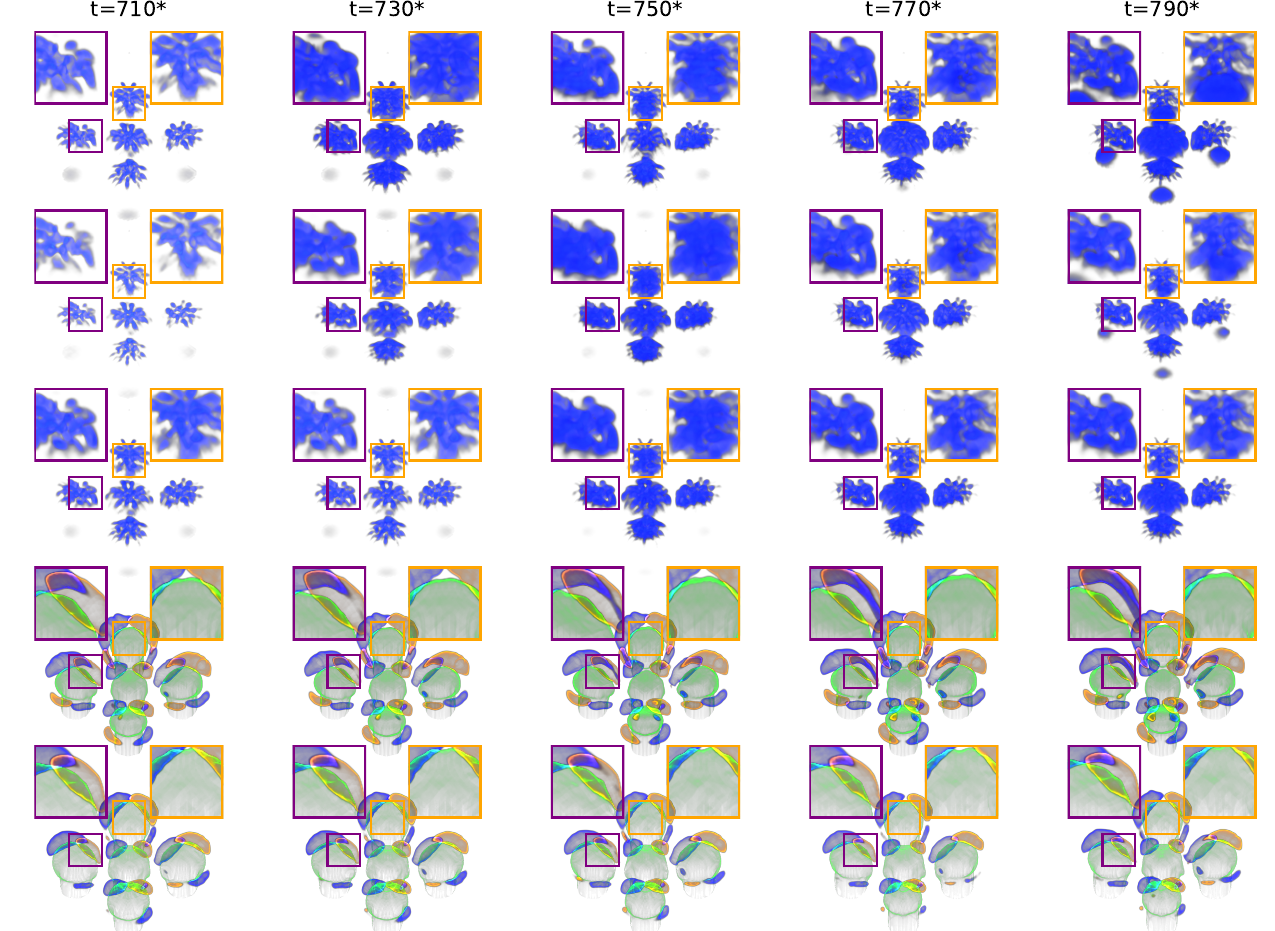}
            };
            \draw[red, line width=0.75pt] (6.75, 3.65) circle (6pt);
            \draw[red, line width=0.75pt] (6.75, 1.58) circle (6pt);
            \draw[red, line width=0.75pt] (6.75, -0.52) circle (6pt);
        \end{tikzpicture}
    \end{minipage}
    \caption{5Jets: \begin{newTVCG}\acronym\end{newTVCG}flow field \begin{newTVCG}estimation\end{newTVCG} and temporal density interpolation during inference, 20$\times$.
        From top to bottom, the rows show GT density, \begin{newTVCG}\acronym\end{newTVCG}interpolated density, CoordNet interpolated density, GT flow, and \begin{newTVCG}\acronym\end{newTVCG}estimated flow. 
        3D rendering was used for the density and flow field visualization (\protect\orangecircle{} \protect\greencircle{} \protect\bluecircle{} colors representing $x$, $y$, and $z$ flow directions respectively), transfer functions are in the supplementary material.
        \vspace{-12pt}
        }
    \label{fig:jets3d}
\end{figure*}

\vspace{-10pt}
\subsection{Droplets: No GT Flow Field Available}
\label{subsec:optical_flow}
The effectiveness of \begin{newTVCG}\acronym\end{newTVCG}in interpolating the luminance
field---captured by cameras during an experiment---based on the
estimated optical flow, even in the absence of GT flow information, is
evident from Fig.~\ref{fig:drop2d}.  The results demonstrate camera
image interpolation by \begin{newTVCG}\acronym\end{newTVCG}of high quality for the Droplets
ensemble, in the second row.
We compare \begin{newTVCG}\acronym's\end{newTVCG} temporal interpolation performance to the one
achieved with RIFE in Fig.~\ref{fig:drop2d}, fourth row.  RIFE's
\begin{newTVCG}interpolation\end{newTVCG} exhibits slightly higher variation compared to the GT,
as can be seen from the difference plots, in the third and fifth rows.
\begin{new}
  It is noteworthy that this concerns experimental data that naturally contains noise. For instance, in the top row of Fig.~\ref{fig:drop2d}, the noise is visible as small, scattered inconsistencies in the background throughout the luminance field, giving it a slightly grainy appearance. Additionally, shadows can be observed on the right side of the images, where darker regions obscure portions of the luminance field. 
  Despite this, \acronym demonstrates robust performance, effectively handling these imperfections and delivering high-quality interpolation results.
\end{new}

When examining the optical flow results generated by \begin{newTVCG}\acronym\end{newTVCG}in Fig.~\ref{fig:drop2d} (sixth row), it is clear that the model successfully captures meaningful flow patterns that correspond to the direction of fluid particle movements in the majority of cases. The visualization shows droplets moving upward with a distinct red hue, while the parts of the splashes collapsing downward are characterized by a green hue. The optical flow estimated by RIFE, in comparison, is significantly less accurate as it fails to capture the finer details of particle movements, resulting in a blurrier flow. 
This indicates that \begin{newTVCG}\acronym\end{newTVCG}is able to learn and predict flow information that aligns with the underlying dynamics of the fluid. 
\begin{newTVCG}
In Fig.~\ref{fig:drop2d} (the second-to-last row for \begin{newTVCG}\acronym\end{newTVCG}and the last row for comparison with RIFE), we further present arrow glyphs depicting flow direction and magnitude. 
With \acronym, these visualizations accurately reflect the movements of the bubble, crown, and droplets, and with this directly capture the evolution of the underlying physical phenomena in a static image (complementing the experimental images in the top row(s)).
For example, in the orange zoom-ins ($t=32$), the arrow glyphs effectively illustrate how the bottom part of the crown collapses downward, while the droplets predominantly move upward as they splash. 
This provides a clear and insightful visualization of the contrasting dynamics within the scene.
When comparing \acronym to RIFE, it becomes evident that our method offers superior accuracy in capturing the flow dynamics. 
RIFE tends to generate excessive flow in the area of the crown, particularly in timesteps such as $t=26$, 27, 30, and 32.
For example, the red bottom zoom-in ($t=30$) demonstrates such a case where in the original experimental images only the boundaries of the bubble are visibly moving (see top row, Dens GT). 
Additionally, RIFE produces less accurate flow representations for the droplet splashes in the top part of the scene, as clearly seen in timesteps $t=26$, 29 (red zoom-in), and 33.
In contrast, \begin{newTVCG}\acronym\end{newTVCG}adequately captures these flow patterns, providing an expressive representation for the analysis of fluid dynamics. 
In sum, \begin{newTVCG}\acronym\end{newTVCG}generates flow fields that align with observed movements for analysis, even without GT flow.
\end{newTVCG}

\vspace{-10pt}
\subsection{5Jets: Density \& Flow Available for Some Timesteps}
\label{subsec:physical_flow_jets}
Third, we consider the scenario in which we
have GT density and velocity fields available for some timesteps sampled from the whole 3D dataset.
In this case, the goal is to perform unsupervised density interpolation (i.e., density TSR) and flow \begin{newTVCG}estimation\end{newTVCG} via a flow-supervised learning scenario.
As \autoref{fig:jets3d} shows, our approach achieves
accurate performance in terms of density field interpolation and flow field \begin{newTVCG}estimation\end{newTVCG}.
In this example, results were obtained at an interpolation rate of $20\times$.
Visually, the error between the renderings of the reconstructed density field and its GT is very small.
Moreover, it shows that even at a comparably large interpolation rate of $20\times$, the model is able to effectively learn a flow field that structurally resembles the GT flow.
Comparing our \begin{newTVCG}\acronym\end{newTVCG}method to CoordNet, we can see that \begin{newTVCG}interpolation\end{newTVCG} of the density field worked well for both (although some differences can be observed, especially highlighted with the red circles).
\begin{newTVCG}
We are using a similar model size of both \begin{newTVCG}\acronym\end{newTVCG}and CoordNet for this comparison.
\end{newTVCG}
Crucially, however, \begin{newTVCG}\acronym\end{newTVCG}additionally supplements the density with accurate flow information (as can be seen in \autoref{fig:jets3d}), a feature that CoordNet lacks.
\begin{newTVCG}
In this dataset, the flow reveals the evolution of five distinct jets along the $x$, $y$, and $z$ directions. 
Observing the flow allows us to interpret how density evolves, particularly in the purple and orange zoom-ins from timesteps $t=710$ to $t=790$. 
Here, we can see that the density concentrates and intensifies along the paths of the jets, forming more pronounced structures as the jets push the material outward. 
This results in higher density regions aligned with the jet flows, indicating areas where the flow is driving the accumulation of matter.
This enhances our understanding of the underlying physical processes and provides a more comprehensive system analysis.
\end{newTVCG}
Examples of the metric scores for the interpolation rates of $10\times$, $15\times$, and $20\times$ can be found in \autoref{jets_comparison}.

\begin{figure*}[!t]
    \centering
    \begin{minipage}[t]{0.10\linewidth}
        \vspace*{-375pt}
        \text{\footnotesize Dens GT} \\ [68pt]
        \text{\footnotesize Dens \begin{newTVCG}\acronym\end{newTVCG}} \\ [68pt]
        \text{\footnotesize Dens STSR-INR} \\ [68pt]
        \text{\footnotesize Flow GT} \\ [68pt]
        \text{\footnotesize Flow \begin{newTVCG}\acronym\end{newTVCG}} \\
    \end{minipage}
    \hspace*{3pt}
    \begin{minipage}[t]{0.88\linewidth}
        \begin{tikzpicture}
            \node (mainfig) at (0,0) {
                \includegraphics[trim=510 70 470 55, clip, width=0.93\linewidth]{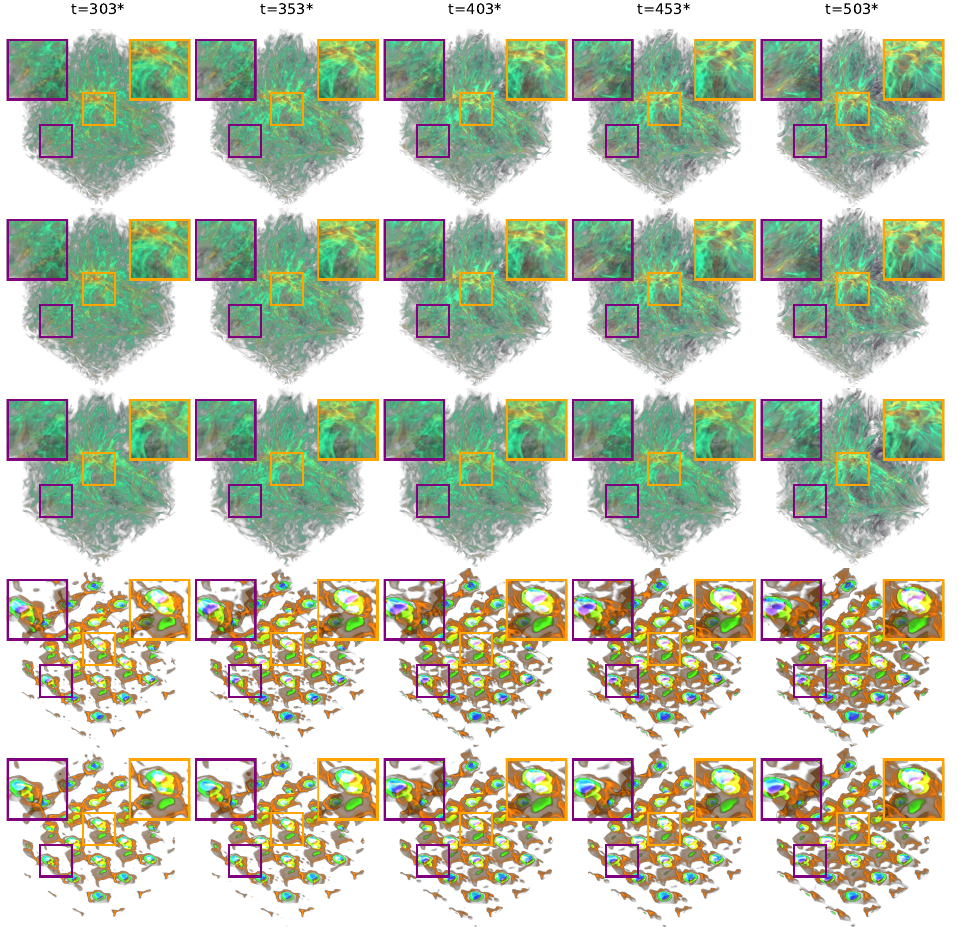}
            };
        \end{tikzpicture}
    \end{minipage}
    \caption{\begin{newTVCG} Nyx: \begin{newTVCG}\acronym\end{newTVCG}flow field \begin{newTVCG}estimation\end{newTVCG} and temporal density interpolation during inference, 5$\times$.
        From top to bottom, the rows show GT density, \begin{newTVCG}\acronym\end{newTVCG}interpolated density, STSR-INR interpolation, GT flow, and \begin{newTVCG}\acronym\end{newTVCG}flow estimation. 
        3D rendering was used for the density and flow field visualization (\protect\orangecircle{} \protect\greencircle{} \protect\bluecircle{} colors representing $x$, $y$, and $z$ flow directions respectively), transfer functions are in the supplementary material.
        \end{newTVCG}
        \vspace{-12pt}
        }
    \label{fig:nyx}
\end{figure*}

\begin{newTVCG}
\vspace{-10pt}
\subsection{Nyx: Density \& Flow Available for Some Timesteps}
\label{subsec:physical_flow_nyx}
Fourth, we consider the scenario where GT density and velocity fields are available for some members of the entire 3D ensemble.
As \autoref{fig:nyx} illustrates, \begin{newTVCG}\acronym\end{newTVCG}achieves accurate performance in terms of both density field interpolation and flow field \begin{newTVCG}estimation\end{newTVCG}.
Visually, the difference between the renderings of the reconstructed density field and its GT is minimal. 
Moreover, even at a relatively high interpolation rate of $5\times$, \begin{newTVCG}\acronym\end{newTVCG}effectively learns a flow field that structurally resembles the GT flow.
When comparing our \begin{newTVCG}\acronym\end{newTVCG}method to STSR-INR for temporal density interpolation (top three rows), it is evident that the \begin{newTVCG}interpolation\end{newTVCG} of the density field is more accurate with \acronym.
For example, at $ t=503$, STSR-INR shows a different structure and less dark matter density compared to \begin{newTVCG}\acronym,\end{newTVCG} which preserves density. 
This trend is consistently observed across all shown timesteps in \autoref{fig:nyx}.
For this comparison, we used the same model size for both \begin{newTVCG}\acronym\end{newTVCG}and STSR-INR.
\begin{new}
By default, Implicit Neural Representation (INR)-based models like CoordNet and STSR-INR have smaller model sizes compared to \acronym due to their simpler architecture, where data is represented as a continuous function parameterized by a compact set of weights. In contrast, \acronym employs a more complex design with multiple convolutional blocks, enabling it to achieve accurate flow estimation and scalar field interpolation. To ensure a fair comparison below, we configured STSR-INR to have the same model size as \acronym, allowing us to directly assess their performance under comparable conditions.
\end{new}

Crucially, \begin{newTVCG}\acronym\end{newTVCG}not only reconstructs the density field but also supplements it with accurate flow information, as shown in \autoref{fig:nyx} (last row), a feature that STSR-INR lacks.
When examining the flow, we observe circular swirling patterns, indicating the complex dynamics of the baryonic gas. 
As these flows intensify, we see evidence of dark matter moving outward---reflected in both the GT and \begin{newTVCG}\acronym\end{newTVCG}density, especially in the orange zoom-ins---consistent with an expanding universe.
This underlines the utility of FLINT in capturing not just the static density fields but also the dynamic evolution of the cosmic structures.
Examples of the metric scores for interpolation rates of $3\times$, $5\times$, and $8\times$ can be found in \autoref{nyx_comparison}.
\end{newTVCG}

\begin{new}
\vspace{-6pt}
\subsection{Nyx: Domain Expert Evaluation}
A domain expert---an assistant professor in astronomy with a research focus on cosmological simulations and observational data evaluation---provided insights into FLINT's utility for analyzing scalar fields and velocity information in the Nyx dataset. He highlighted that FLINT's ability to estimate velocity fields alongside density fields offers significant advantages. The expert remarked, ``compared to other reconstruction methods, with FLINT you get information not only on the \textit{density} in Nyx simulations but also on the \textit{velocity}". This dual capability is particularly important in cosmology, where velocity data derived from ``redshift" measurements are critical for estimating distances and analyzing spatial relationships between objects.

The expert stated that ``having a way to estimate accurately the peculiar velocities would help when comparing with observations''.
He further underlined the relevance of velocity information in creating ``lightcones", which combine simulation timesteps to mimic observations of the universe at different epochs. These lightcones are crucial for studying phenomena such as the intergalactic medium and the 3D distribution of galaxies, where ``the distance information comes really from a velocity shift". He emphasized that ``having a way to estimate accurately the velocities would help when comparing with observations". By accurately reconstructing both density and velocity fields, FLINT enhances the fidelity of simulation outputs aligned with observational data, bridging gaps between simulations and real-world measurements, and enabling a deeper understanding of cosmological dynamics.

\end{new}


\begin{figure*}[ht]
    \centering
    \begin{tabular}{ccc}
        \begin{tabular}{c}
            \includegraphics[trim=10 20 40 40, clip, width=0.28\linewidth]{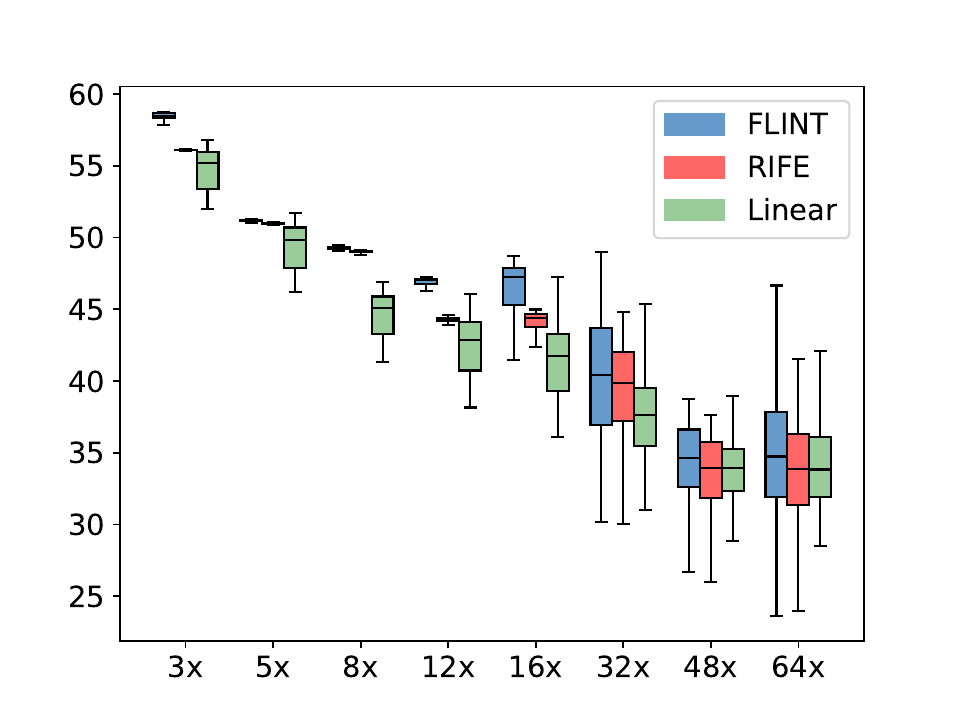} \\
            \small (a) LBS: PSNR $\uparrow$ scores
        \end{tabular} & 
        \begin{tabular}{c}
            \includegraphics[trim=10 20 40 35, clip, width=0.28\linewidth]{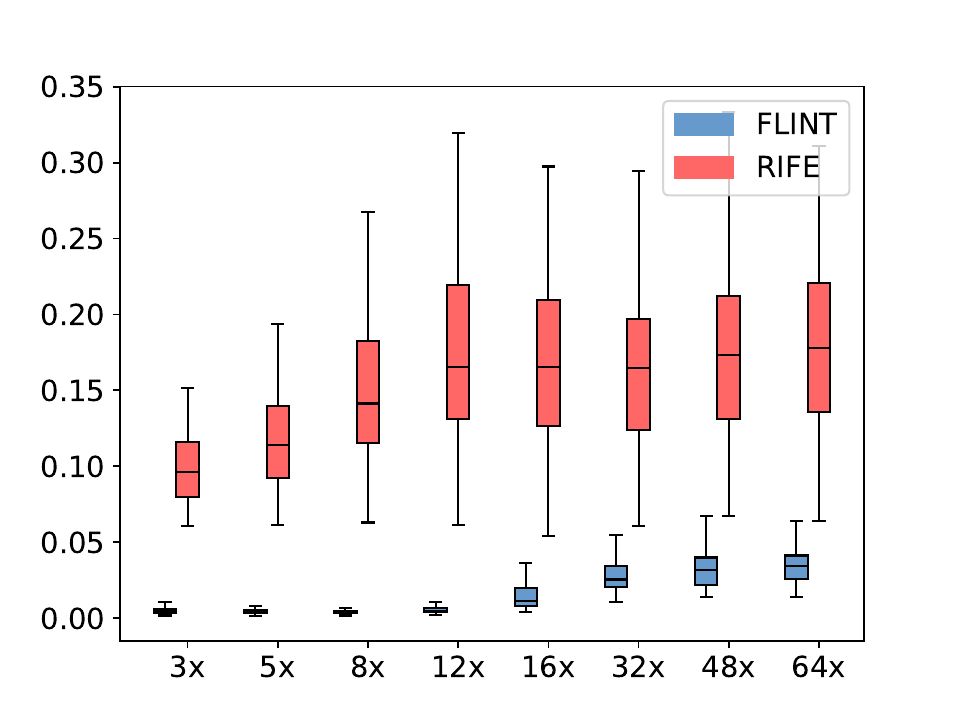} \\
            \small (b) LBS: EPE $\downarrow$ scores
        \end{tabular} & 
        \begin{tabular}{c}
            \includegraphics[trim=10 20 40 40, clip, width=0.28\linewidth]{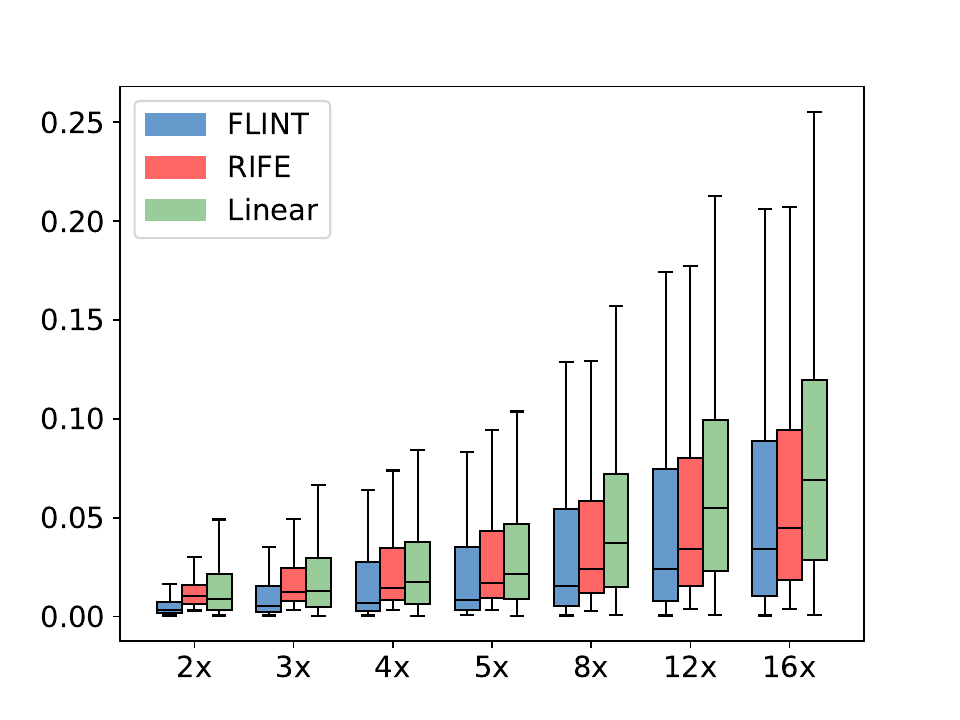} \\
            \small (c) Droplets: LPIPS $\downarrow$ scores
        \end{tabular} \\
    \end{tabular}
    
    \caption{Comparison of \begin{newTVCG}\acronym,\end{newTVCG} RIFE, and linear interpolation at various interpolation rates. 
    For PSNR plots for the Droplets dataset and LPIPS scores for the LBS dataset, see Figures~\ref{fig:drop_box} and~\ref{fig:lbs_box} of the supplementary material, respectively.}
    \vspace{-12pt}
    \label{fig:lbs_droplet_box}
\end{figure*}

\section{Quantitative and Comparative Evaluation}
\label{sec:compar_eval}
	
In this section, we present quantitative results and compare against
baseline methods to demonstrate the improvement achieved with our
proposed method, followed by ablation and parameter studies to explore
different configurations and hyperparameters.

\vspace{-6pt}
\subsection{Comparison Against Baselines}
	
We illustrate quantitative results using boxplots in Fig.~\ref{fig:lbs_droplet_box} for LBS and Droplets ensembles. \begin{newTVCG}\acronym\end{newTVCG}(blue) is compared against two baselines, RIFE (red) and linear interpolation (green), across various interpolation factors ranging from 2 to 64, showcasing representative outcomes for both ensembles.
Similarly, in LPIPS-based evaluation for the Droplets, \begin{newTVCG}\acronym\end{newTVCG}demonstrates better performance.
Furthermore, examining the flow \begin{newTVCG}estimation\end{newTVCG} (supplementing density with flow
information during inference) results for the LBS
ensemble in Fig.~\ref{fig:lbs_droplet_box}b reveals that \begin{newTVCG}\acronym\end{newTVCG}achieves
a significantly lower endpoint error with minimal variance when
compared to \begin{old}the RIFE
  baseline\end{old}RIFE.  This underscores \begin{newTVCG}\acronym's\end{newTVCG}
proficiency in learning an accurate flow representation that closely
aligns with the GT physical flow, particularly evident in the LBS
ensemble scenario.  As linear interpolation cannot provide flow
\begin{newTVCG}estimation\end{newTVCG} results, no statistics are shown for it, in the case of the
LBS ensemble.

In the case of the Droplets ensemble, \begin{newTVCG}\acronym\end{newTVCG}significantly outperforms
linear interpolation for both LPIPS and PSNR metrics, confirming our
expectations, as shown in Fig.~\ref{fig:lbs_droplet_box}c and
supplementary material.  This superior performance can be attributed to
\acronym's capacity to learn robust optical flow, which in turn enhances
its ability to interpolate and reconstruct complex density fields.
\begin{newTVCG}\acronym\end{newTVCG}consistently performs better than RIFE as well, across all
presented interpolation rates, for both PSNR and LPIPS metrics.  This
performance reaffirms that our proposed model is well-suited for the
challenging task of reconstructing density fields within scientific
ensembles.  These findings underscore the potential of \begin{newTVCG}\acronym\end{newTVCG}in
spatiotemporal data \begin{newTVCG}interpolation\end{newTVCG} tasks, with application for
scientific visualization and data analysis.

CoordNet~\cite{han2022coordnet}, another method we considered for
comparative evaluation, offers an advanced framework for various tasks
in time-varying volumetric data visualization, including TSR.  It has shown improvements over the TSR-TVD
method~\cite{han2019tsr} discussed in \autoref{sec:related_work},
positioning it as a relevant benchmark for temporal interpolation
performance. 
We trained CoordNet for the 3D+time 5Jets dataset and compared across various interpolation rates, see \autoref{jets_comparison}.
\begin{newTVCG}\acronym\end{newTVCG}demonstrates competitive performance in terms of PSNR score for density \begin{newTVCG}interpolation\end{newTVCG} while also serving the dual purpose of enabling flow \begin{newTVCG}estimation\end{newTVCG}.
While CoordNet facilitates TSR, it does not extend to flow field \begin{newTVCG}estimation\end{newTVCG}, which we consider to be the main contribution of this work.

\begin{table}[h]
\scalefont{0.9}
\centering
  \caption{ Comparison against baselines, 5Jets  }
  \begin{tabular}{lc@{\hspace{5pt}}c@{\hspace{3pt}}c|c@{\hspace{5pt}}c@{\hspace{3pt}}c|c@{\hspace{5pt}}c@{\hspace{3pt}}c} 
    \toprule
    \multirow{2}{*}{\textbf{Method}} & \multicolumn{3}{c}{\textbf{10$\times$}} & \multicolumn{3}{c}{\textbf{15$\times$}} & \multicolumn{3}{c}{\textbf{20$\times$}} \\
    \cmidrule(lr){2-4} \cmidrule(lr){5-7} \cmidrule(lr){8-10}
    & \textbf{PSNR $\uparrow$} & \textbf{EPE $\downarrow$} & & \textbf{PSNR $\uparrow$} & \textbf{EPE $\downarrow$} & & \textbf{PSNR $\uparrow$} & \textbf{EPE $\downarrow$} & \\
    \midrule
    \begin{newTVCG}\acronym\end{newTVCG}& 46.72 & 0.7271 & & \textbf{45.41} & 0.7286 & & \textbf{44.93} & 0.7292 & \\
    Linear & 44.70 & --- & & 42.28 & --- & & 38.11 & --- & \\
    CoordNet & \textbf{48.45} & --- & & 44.89 & --- & & 43.74 & --- & \\
    \bottomrule
  \end{tabular}
  \label{jets_comparison}
\end{table}

\begin{newTVCG}
STSR-INR~\cite{tang2024stsr}, another method we considered for comparative evaluation, was applied to the 3D+time Nyx dataset and compared across various interpolation rates, as shown in \autoref{nyx_comparison}.
STSR-INR is an INR-based approach that employs a variable embedding scheme to learn latent vectors for different variables.
This method utilizes a variational auto-decoder to optimize the learnable latent vectors, enabling latent-space interpolation.
STSR-INR shows improvements over both STNet~\cite{han2021stnet} and CoordNet~\cite{han2022coordnet}, positioning it as a relevant benchmark for TSR performance.
\begin{newTVCG}\acronym\end{newTVCG}demonstrates superior performance in terms of PSNR score for density \begin{newTVCG}interpolation\end{newTVCG} while also serving the dual purpose of enabling flow estimation, a capability that extends beyond the functionalities provided by STSR-INR.
\end{newTVCG}

\begin{table}[h]
\scalefont{0.9}
\centering
\begin{newTVCG}
  \caption{\begin{newTVCG} Comparison against baselines, Nyx \end{newTVCG}}
  \begin{tabular}{lc@{\hspace{5pt}}c@{\hspace{3pt}}c|c@{\hspace{5pt}}c@{\hspace{3pt}}c|c@{\hspace{5pt}}c@{\hspace{3pt}}c} 
    \toprule
    \multirow{2}{*}{\textbf{Method}} & \multicolumn{3}{c}{\textbf{3$\times$}} & \multicolumn{3}{c}{\textbf{5$\times$}} & \multicolumn{3}{c}{\textbf{8$\times$}} \\
    \cmidrule(lr){2-4} \cmidrule(lr){5-7} \cmidrule(lr){8-10}
    & \textbf{PSNR $\uparrow$} & \textbf{EPE $\downarrow$} & & \textbf{PSNR $\uparrow$} & \textbf{EPE $\downarrow$} & & \textbf{PSNR $\uparrow$} & \textbf{EPE $\downarrow$} & \\
    \midrule
    \begin{newTVCG}\acronym\end{newTVCG}& \textbf{53.17} & 0.0310 & & \textbf{52.31} & 0.0310 & & \textbf{49.39} & 0.0309 & \\
    Linear & 47.49 & --- & & 41.92 & --- & & 37.51 & --- & \\
    STSR-INR & 49.63 & --- & & 44.21 & --- & & 41.09 & --- & \\
    \bottomrule
  \end{tabular}
  \label{nyx_comparison}
\end{newTVCG}
\end{table}

\begin{new}
\textbf{Evaluation on Additional Scalar Field}. 
In addition to evaluating FLINT for density and flow fields, we extended our analysis to another scalar field to demonstrate its versatility. Specifically, we evaluated FLINT's performance on the temperature field from the Nyx simulation ensemble. We examined FLINT's ability to interpolate the temperature field and estimate the corresponding flow fields. Qualitative results are shown in Fig.~\ref{fig:nyx_temp} of the supplementary material, and quantitative evaluations are provided in Table~\ref{nyx_comparison_fields}.
The temperature with flow results indicate that the temperature interpolation performance remains robust across different interpolation rates, although the PSNR values are slightly lower compared to density results. This difference reflects the increased complexity of interpolating temperature fields due to their distinct data characteristics. Similarly, the flow results show lower EPE scores. This is expected, as estimating flow from temperature is inherently more challenging in comparison to density due to its weaker correlation with movement or physical flow dynamics.

Our findings align with observations by Gu~\emph{et al.}~\cite{gu2022scalar2vec}, where the performance of vector field reconstruction can vary depending on the input scalar field. Similarly, FLINT's performance showed variations when applied to the temperature field, reflecting the challenges associated with different data characteristics. 

These results further demonstrate FLINT's adaptability across diverse scalar fields while highlighting the influence of field-specific dynamics on reconstruction accuracy.
\end{new}

\begin{table}[h]
\scalefont{0.9}
\centering
\begin{newTVCG}
  \caption{\begin{new} Comparison of different fields, Nyx \end{new}}
  \begin{tabular}{l@{\hspace{5pt}}c@{\hspace{5pt}}c@{\hspace{10pt}}c@{\hspace{5pt}}c@{\hspace{5pt}}c@{\hspace{10pt}}c@{\hspace{5pt}}c@{\hspace{5pt}}c} 
    \toprule
    \multirow{2}{*}{\textbf{Field}} & \multicolumn{2}{c}{\textbf{3$\times$}} & & \multicolumn{2}{c}{\textbf{5$\times$}} & & \multicolumn{2}{c}{\textbf{8$\times$}} \\
    \cmidrule(lr){2-3} \cmidrule(lr){5-6} \cmidrule(lr){8-9}
    & \textbf{PSNR $\uparrow$} & \textbf{EPE $\downarrow$} & & \textbf{PSNR $\uparrow$} & \textbf{EPE $\downarrow$} & & \textbf{PSNR $\uparrow$} & \textbf{EPE $\downarrow$} \\
    \midrule
    Density + Flow & 53.17 & 0.0310 & & 52.31 & 0.0310 & & 49.39 & 0.0309 \\
    Temperature + Flow & 49.32 & 0.0446 & & 46.91 & 0.0451 & & 44.43 & 0.0467 \\
    \bottomrule
  \end{tabular}
  \label{nyx_comparison_fields}
\end{newTVCG}
\end{table}

\begin{new}

\textbf{Impact of Noise.}  
To evaluate FLINT's robustness to noise, we introduced random Gaussian noise into the 3D Nyx simulation ensemble. 
Even with the added noise in the GT density, FLINT achieved reliable results, maintaining an average PSNR of 46.36 dB and an EPE of 0.0356 at a $5\times$ interpolation rate.
In comparison, without noise, the model achieved a higher PSNR of 52.31 dB and a slightly lower EPE of 0.0310, demonstrating a modest decline in performance due to noise. 
This highlights FLINT’s resilience to noisy data and reinforces its applicability to real-world scenarios where imperfections in scientific datasets are common.
For further details, please refer to the supplementary material.

\textbf{Large Interpolation Rates.}  
At large interpolation rates, such as $32\times$ (Fig.~\ref{fig:nyx_32x} of the supplementary material), we observe that while density interpolation maintains acceptable quality (average PSNR of 37.71 dB), the accuracy of flow estimation noticeably degrades (average EPE is 0.0479).
In comparison, at $8\times$ interpolation, the model achieved a PSNR of 49.39 dB and an EPE of 0.0309, while at $16\times$, the PSNR was 43.42 dB with an EPE of 0.0338.
The degradation at a large interpolation rate is expected due to the increased structural differences between flows at widely separated time steps, which make flow estimation inherently more challenging.
Nevertheless, FLINT delivers satisfactory results for scalar field interpolation even at these demanding rates.
Qualitatively, while the main structure, such as the distribution of dark matter, is preserved and remains visually consistent with the GT, finer details and intricate patterns become less accurate.
Further details are provided in the supplementary material.



\end{new}

\begin{new}\textbf{Inference time comparison.}\end{new}
    We conducted an inference time comparison of our model, \acronym, against RIFE, CoordNet, and STSR-INR across all datasets. 
    Specifically, we compared \acronym with RIFE on the 2D ensembles (LBS and Droplets) and with CoordNet and STSR-INR on the 3D datasets (5Jets and Nyx). 
    For each evaluation, we measured the inference time over 3K timesteps for the 2D ensembles and 300 timesteps for the 3D ensembles from the test set. 
    Our findings indicate that \acronym achieves faster inference time compared to RIFE, with an average of 0.025 seconds per timestep on the 2D datasets, as opposed to RIFE’s 0.035 seconds. 
    On the 3D datasets, \acronym significantly outperforms both CoordNet and STSR-INR in inference speed, even achieving interactive rates with an average time of 0.2 seconds per timestep for \acronym. In comparison, it takes 2.1 seconds for CoordNet and 1.5 seconds for STSR-INR.
    The faster inference time of FLINT compared to INR-based approaches like CoordNet and STSR-INR can be attributed to the inherent efficiency of convolutional neural networks (CNNs), which excel in parallel processing and optimized memory access, while INR-based methods require solving implicit functions at each point, adding computational complexity.

    \vspace{-6pt}
    \subsection{Ablation Studies}
    \label{subsect:ablation}
    Our proposed \begin{newTVCG}\acronym\end{newTVCG}method is a deep neural network that has
    various loss components for training the
      network.  To assess the impact of these
    in achieving optimal results, we conducted a series of ablation
    studies.
	
    \begin{table}[h]
    \scalefont{0.9}
      \centering
      \caption{Ablation of \begin{newTVCG}\acronym\end{newTVCG}\begin{newTVCG}(flow-supervised) \end{newTVCG}}
      \begin{tabular}{lc@{\hspace{5pt}}c@{\hspace{3pt}}c|c@{\hspace{5pt}}c@{\hspace{3pt}}c|c@{\hspace{5pt}}c@{\hspace{3pt}}c} 
        \toprule
        \textbf{Method} & \multicolumn{3}{c}{\textbf{LBS, 3$\times$}} & \multicolumn{3}{c}{\textbf{5Jets, 10$\times$}} & \multicolumn{3}{c}{\begin{newTVCG}\textbf{Nyx, 5$\times$}\end{newTVCG}} \\
        \cmidrule(lr){2-4} \cmidrule(lr){5-7} \cmidrule(lr){8-10}
        & \textbf{PSNR $\uparrow$} & \textbf{EPE $\downarrow$} & & \textbf{PSNR $\uparrow$} & \textbf{EPE $\downarrow$} & & \textbf{PSNR $\uparrow$} & \textbf{EPE $\downarrow$} & \\
        \midrule
        \begin{newTVCG}\acronym\end{newTVCG}no \textit{flow} & 57.41 & 0.0842 & & 37.80 & 3.4985 & & 51.97 & 0.1461 & \\
        \begin{newTVCG}\acronym\end{newTVCG}no \textit{rec} & 49.89 & 0.0119 & & 32.32 & 0.7332 & & 44.67 & 0.0497 & \\
        \begin{newTVCG}\acronym\end{newTVCG}w/o s-t & 56.37 & 0.0064 & & 41.99 & 0.7323 & & 52.08 & 0.0334 & \\
        \begin{newTVCG}\acronym\end{newTVCG}& \textbf{58.44} & \textbf{0.0051} & & \textbf{46.72} & \textbf{0.7271} & & \textbf{52.31} & \textbf{0.0310} & \\
        \bottomrule
        \end{tabular}
        \label{sup_ablation}
        
      \vspace*{0.2cm}
		\caption{\scalefont{0.97}Ablation of \begin{newTVCG}\acronym,\end{newTVCG} Droplets, 2$\times$ \begin{newTVCG}(flow-unsupervised) \end{newTVCG}}
    \vspace*{-0.3cm}
	\begin{tabular}{lcc}
		\toprule
		\textbf{Method} & \textbf{PSNR $\uparrow$} & \textbf{LPIPS $\downarrow$} \\
		\midrule
		\begin{newTVCG}\acronym\end{newTVCG}no \textit{rec} & 32.09 & 0.0373 \\ 
		\begin{newTVCG}\acronym\end{newTVCG}no \textit{dis} & 39.87 & 0.0129 \\
            \begin{newTVCG}\acronym\end{newTVCG}no \textit{photo} & 40.47 & 0.0105 \\
		\begin{newTVCG}\acronym\end{newTVCG}no \textit{reg} & 40.71 & 0.0098 \\
            \begin{newTVCG}\acronym\end{newTVCG}no \textit{dis, photo} & 40.36 & 0.0116 \\
            \begin{newTVCG}\acronym\end{newTVCG}no \textit{dis, reg} & 40.37 & 0.0117 \\
            \begin{newTVCG}\acronym\end{newTVCG}no \textit{reg, photo} & 40.11 & 0.0110 \\
            \begin{newTVCG}\acronym\end{newTVCG}no \textit{dis, reg, photo} & 39.81 & 0.0119 \\
            \begin{newTVCG}\acronym\end{newTVCG}w/o s-t & 40.25 & 0.0202 \\
		\begin{newTVCG}\acronym\end{newTVCG}& \textbf{41.16} & \textbf{0.0087} \\
		\bottomrule
	\end{tabular}
	\label{unsup_ablation}
 \vspace{-6pt}
\end{table}

    The ablation studies conducted on \begin{newTVCG}\acronym\end{newTVCG}provide valuable insights
    into the impact of different loss components on the overall
    performance.  In Table~\ref{sup_ablation}, which presents the
    results for the LBS, 5Jets, \begin{newTVCG}and Nyx\end{newTVCG} datasets with available GT flow fields for the model
    training, four variants of \begin{newTVCG}\acronym\end{newTVCG}are compared: ``\begin{newTVCG}\acronym\end{newTVCG}no
    \textit{flow}", ``\begin{newTVCG}\acronym\end{newTVCG}no \textit{rec}", ``\begin{newTVCG}\acronym\end{newTVCG}w/o s-t", and
    ``\acronym''.  ``\begin{newTVCG}\acronym\end{newTVCG}no \textit{flow}" refers to \begin{newTVCG}\acronym\end{newTVCG}without the
    inclusion of the flow-related loss component.  As expected, this
    variant demonstrates considerably less accurate physical flow,
    measured by EPE metric, despite a relatively
    high density interpolation score, measured
    by PSNR \begin{old}score
    \end{old}metric. ``\begin{newTVCG}\acronym\end{newTVCG}no \textit{rec}" refers to \begin{newTVCG}\acronym\end{newTVCG}without
    the reconstruction-related loss component.  In this case, the
    \begin{old}physical\end{old} flow learned by the model is
    relatively good, however, the density
    \begin{newTVCG}interpolation\end{newTVCG} suffers due to the absence of its related loss
    component. ``\begin{newTVCG}\acronym\end{newTVCG}w/o s-t" refers to \begin{newTVCG}\acronym\end{newTVCG}\begin{old}without
      the student-teacher architecture\end{old}
    where the teacher block (see Fig.~\ref{fig:model}) and
      the teacher-related terms in the loss functions are
      removed.  This scenario leads to slightly less accurate \begin{newTVCG}interpolation\end{newTVCG} and \begin{newTVCG}flow estimation\end{newTVCG} due to the fact that the model does not receive GT densities during the training.  Finally, ``\acronym''
    represents the complete \begin{newTVCG}\acronym\end{newTVCG}model with all loss components.  It
    achieves the highest scores across all metrics, including PSNR and
    EPE, outperforming \begin{old}both
      the \begin{old}baseline\end{old} RIFE method and the ``\begin{newTVCG}\acronym\end{newTVCG}no
      \textit{flow}" variant\end{old} all other
      variants.  This demonstrates the significance of
    incorporating the proposed loss components in achieving superior
    \begin{newTVCG}interpolation\end{newTVCG} results.
		
Similar ablation studies were performed for the Droplets ensemble, for which no GT flow
fields are available, as shown in Table~\ref{unsup_ablation}.  Multiple variants of \begin{newTVCG}\acronym\end{newTVCG}were
evaluated, including ``\begin{newTVCG}\acronym\end{newTVCG}no \textit{rec}", ``\begin{newTVCG}\acronym\end{newTVCG}no
\textit{dis}" (without distillation), ``\begin{newTVCG}\acronym\end{newTVCG}no \textit{photo}"
(without photometric loss), ``\begin{newTVCG}\acronym\end{newTVCG}no \textit{reg}" (without
regularization), ablation of combinations of these losses, and ``\acronym
w/o s-t''.  Each variant is compared to the complete \begin{newTVCG}\acronym\end{newTVCG}method.  The
results indicate that each loss component contributes to the overall
performance of \begin{newTVCG}\acronym.\end{newTVCG}  The complete \begin{newTVCG}\acronym\end{newTVCG}model outperforms all the
variants in terms of PSNR and LPIPS scores.  This showcases the
importance of reconstruction, distillation, photometric, and
regularization loss components in achieving the best possible
reconstructions for the Droplets ensemble.

\subsection{Parameter Studies}
\label{subsec:param_study}
	
Our proposed \begin{newTVCG}\acronym\end{newTVCG}method involves several hyperparameters, and we
performed extensive parameter studies to understand their impact on
achieving optimal results.
Table~\ref{hyperparameter_search} displays the outcomes of these studies regarding 2D+time and 3D+time datasets.

\setlength{\tabcolsep}{2pt} 
\renewcommand{\arraystretch}{0.8} 

\begin{table}[ht]
\centering
\caption{Hyperparameter search for \begin{newTVCG}\acronym\end{newTVCG}}
\scalefont{0.75}
\begin{tabular}{@{}lcc|cc|cc|cc@{}}
\toprule
\textbf{Method} & \multicolumn{2}{c}{\textbf{LBS, 3$\times$}} & \multicolumn{2}{c}{\textbf{5Jets, 10$\times$}} & \multicolumn{2}{c}{\begin{newTVCG}\textbf{Nyx, 5$\times$}\end{newTVCG}} & \multicolumn{2}{c}{\textbf{Droplets, 2$\times$}} \\
\cmidrule(lr){2-3} \cmidrule(lr){4-5} \cmidrule(lr){6-7} \cmidrule(lr){8-9}
& \textbf{PSNR $\uparrow$} & \textbf{EPE $\downarrow$} & \textbf{PSNR $\uparrow$} & \textbf{EPE $\downarrow$} & \textbf{PSNR $\uparrow$} & \textbf{EPE $\downarrow$} & \textbf{PSNR $\uparrow$} & \textbf{LPIPS $\downarrow$} \\
\midrule
\begin{newTVCG}\acronym\end{newTVCG}\textit{128} & 49.86 & 0.0135 & 43.57 & 0.7397 & 51.68 & 0.0315 & 36.89 & 0.0162 \\
\begin{newTVCG}\acronym\end{newTVCG}\textit{Lapl} & 57.23 & 0.0112 & 42.82 & 0.7387 & 51.73 & 0.0319 & 39.44 & 0.0192 \\
\begin{newTVCG}\acronym\end{newTVCG}\textit{$\lambda_{flow} = 0.3$} & 58.03 & 0.0109 & 44.89 & 0.7308 & 52.29 & 0.0311 & --- & --- \\
\begin{newTVCG}\acronym\end{newTVCG}\textit{$\lambda_{flow} = 0.1$} & 58.17 & 0.0113 & 45.11 & 0.7549 & 52.19 & 0.0320 & --- & --- \\
\begin{newTVCG}\acronym\end{newTVCG}\textit{$\gamma = 0.9$} & 55.81 & 0.0130 & 44.84 & 0.7327 & 52.14 & \textbf{0.0310} & --- & --- \\
\begin{newTVCG}\acronym\end{newTVCG}\textit{$\gamma = 0.7$} & 56.28 & 0.0137 & 46.08 & 0.7384 & 52.28 & 0.0317 & --- & --- \\
\begin{newTVCG}\acronym\end{newTVCG}\textit{$\lambda_{reg} = 10^{-9}$} & --- & --- & --- & --- & --- & --- & 40.80 & 0.0089 \\
\begin{newTVCG}\acronym\end{newTVCG}\textit{$\lambda_{reg} = 10^{-7}$} & --- & --- & --- & --- & --- & --- & 40.06 & 0.0122 \\
\begin{newTVCG}\acronym\end{newTVCG}\textit{$\lambda_{photo} = 10^{-7}$} & --- & --- & --- & --- & --- & --- & 40.79 & 0.0107 \\
\begin{newTVCG}\acronym\end{newTVCG}\textit{$\lambda_{photo} = 10^{-5}$} & --- & --- & --- & --- & --- & --- & 38.98 & 0.0144 \\
\begin{newTVCG}\acronym\end{newTVCG}\textit{$\lambda_{dis} = 1\times10^{-5}$} & --- & --- & --- & --- & --- & --- & 37.86 & 0.0157 \\
\begin{newTVCG}\acronym\end{newTVCG}\textit{$\lambda_{dis} = 1\times10^{-3}$} & --- & --- & --- & --- & --- & --- & 40.34 & 0.0114 \\
\begin{newTVCG}\acronym\end{newTVCG}\textit{smooth} & --- & --- & --- & --- & --- & --- & 36.96 & 0.0125 \\
\begin{newTVCG}\acronym\end{newTVCG}\textit{stride = 1} & 51.33 & 0.0147 & 43.41 & 0.8116 & 52.19 & 0.0418 & 38.64 & 0.0149 \\
\begin{newTVCG}\acronym\end{newTVCG}\textit{3 Blocks} & 53.08 & 0.0136 & 43.09 & 0.7346 & 52.28 & 0.0312 & 39.74 & 0.0144 \\
\begin{newTVCG}\acronym\end{newTVCG}\textit{4 Blocks} & \textbf{58.44} & \textbf{0.0051} & \textbf{46.72} & \textbf{0.7271} & \textbf{52.31} & \textbf{0.0310} & \textbf{41.16} & \textbf{0.0087} \\
\begin{newTVCG}\acronym\end{newTVCG}\textit{5 Blocks} & 57.20 & \textbf{0.0051} & 45.57 & 0.7420 & 52.15 & \textbf{0.0310} & 41.04 & 0.0113 \\
\begin{newTVCG}\acronym\end{newTVCG}\textit{6 Blocks} & 56.17 & 0.0106 & 44.61 & 0.7691 & 51.98 & 0.0313 & 39.79 & 0.0134 \\
\begin{new}\acronym \textit{two-stage} \end{new} & \begin{new}56.87\end{new} & \begin{new}0.0069\end{new} & \begin{new}45.78\end{new} & \begin{new}0.7409\end{new} & \begin{new}52.07\end{new} & \begin{new}\textbf{0.0310}\end{new} & \begin{new}40.95\end{new} & \begin{new}0.0108\end{new} \\
\bottomrule
\end{tabular}
\label{hyperparameter_search}
\end{table}

The parameter search investigates various configurations of \begin{newTVCG}\acronym\end{newTVCG}, each labeled with a specific identifier. To determine the model's optimal architecture for our task, we conducted a series of experiments, adjusting the model's capacity by varying its width and depth. 
For example, ``\begin{newTVCG}\acronym\end{newTVCG}\textit{128}" represents a configuration with 128 channels in all convolutional layers across all blocks. 
Through hyperparameter optimization, we found that the best-performing model, named ``\begin{newTVCG}\acronym''\end{newTVCG} or ``\begin{newTVCG}\acronym\end{newTVCG}\textit{4 Blocks}", increases both the channel capacity and the number of blocks. This configuration, with 4 blocks and channel counts in the convolutional layers ranging from 256 to 128, is designed to capture more intricate and detailed features within the data.
Another configuration is ``\begin{newTVCG}\acronym\end{newTVCG}\textit{smooth}", where a smoothness
loss component was incorporated into the training process for Droplets, however, not yielding the best results.
\begin{newTVCG}
Additionally, ``\begin{newTVCG}\acronym\end{newTVCG}\textit{stride = 1}" represents an architecture where each block uses convolutional and deconvolutional layers with a stride of one, instead of varying strides.
\end{newTVCG}
Furthermore, we explored two different loss functions for the
\begin{newTVCG}interpolation\end{newTVCG} part of the model.  The configuration ``\acronym
\textit{Lapl}" indicates the use of the Laplacian loss, which measures
the $L_1$-loss between two Laplacian pyramid representations
(pyramidal level is 5) of the reconstructed density field and GT density field. 
The configuration ``\acronym'' employs the simple
$L_1$-distance as the loss function, as described in
Sec.~\ref{subsec:loss}.  
In addition, we varied the number of blocks of
\begin{newTVCG}\acronym\end{newTVCG}(\textit{Conv Block} in Fig.~\ref{fig:model}), ranging from three
to six.  As can be seen from Table~\ref{hyperparameter_search}, the most optimal configuration is the one with four
blocks (``\begin{newTVCG}\acronym\end{newTVCG}\textit{4 Blocks}").
Models with fewer than four blocks lack sufficient capacity, while those with more begin to overfit.
\begin{new}
  The configuration ``\acronym \textit{two-stage}'' employs a two-stage optimization approach where a teacher model is trained first, followed by a student network trained to align with the teacher's outputs. In contrast, ``\acronym'' achieves slightly better performance overall while reducing training time by nearly half (12 vs. 20 hours), demonstrating its efficiency.
\end{new}

The results in Table~\ref{hyperparameter_search} demonstrate optimized \begin{newTVCG}\acronym'\end{newTVCG}s superior performance over other configurations, with the highest scores in PSNR and EPE metrics. 
This underscores the importance of channel capacity, number of blocks, and optimized loss functions in capturing the intricate dynamics of fluid ensembles, enabling \begin{newTVCG}\acronym'\end{newTVCG}s best density \begin{newTVCG}interpolation\end{newTVCG} and flow \begin{newTVCG}estimation\end{newTVCG} capabilities.


\section{Discussion and Conclusion}

\begin{newTVCG}
In this work, we proposed \acronym, a learning-based approach for the estimation of flow information and scalar field interpolation for 2D+time and 3D+time scientific ensembles.
\acronym offers flexibility in handling various data availability scenarios.
It can perform flow-supervised learning to estimate flow fields for members when partial flow data is available. 
In cases where no flow data is provided, common in experimental datasets, \acronym employs a flow-unsupervised approach, generating flow fields based on optical flow concepts. 
Additionally, \acronym generates high-quality temporal interpolants between scalar fields like density or luminance, outperforming recent state-of-the-art methods.
It achieves fast inference and does not require complex training procedures, such as pre-training or fine-tuning on simplified datasets, commonly required by other flow estimation methods~\cite{teed2020raft, dosovitskiy2015flownet, luo2021upflow}. Its effectiveness in producing high-quality flow and scalar fields has been validated across both simulation and experimental data.
\end{newTVCG}

\begin{newTVCG}For future work, we aim to improve the \begin{newTVCG}\acronym\end{newTVCG}method and expand its application to various problems and domains.\end{newTVCG}
\begin{newTVCG}\acronym\end{newTVCG}is generally applicable to various fields, including density, flow, energy, as well as to grayscale or RGB datasets, and is valuable for visualization applications, and we aim to study a larger range of applications in future work.
\begin{newTVCG}
\begin{newTVCG}\acronym\end{newTVCG}extensions could include expanding support for more timesteps as input, exploring extrapolation in addition to interpolation, or ensemble parameter space exploration.
\end{newTVCG}
Another promising direction for future work is harnessing the learned flow field for efficient dimensionality reduction~(DR).
The compressed representation of the estimated flow could be used in order to perform DR of the ensemble members and compared against standard and ML-based DR techniques~\cite{gadirov2021evaluation}.
\begin{newTVCG}
Furthermore, in the medical domain, learned optical flow can serve as additional input to classification models~\cite{wang2015detection}.
This strengthens disease classification tasks by incorporating motion fields, which convey crucial information.
As a result, leveraging estimated flow enhances the accuracy and robustness of disease classification models, ultimately contributing to improved healthcare outcomes.
\end{newTVCG}

\bibliographystyle{IEEEtran}
\bibliography{FLINT}

	
	
\onecolumn
\newpage
\twocolumn

\renewcommand{\appendixname}{\vskip -1cm}
   
\appendix
\begin{old}\section{Appendix}\end{old}

\section*{Supplementary Material}

\subsection{Scientific Ensembles}
\label{subsec:appendix_ensembles}

Examples of fields at different timesteps from various members of both
the LBS and Droplets ensembles can be found in
Figs.~\ref{fig:lbs_samples} and \ref{fig:drop_original}.  These
figures offer a visual representation of the data from simulations and
physical experiments, which are used in our research.  One can observe
that Fig.~\ref{fig:lbs_samples} illustrates the considerable
structural diversity that exists between the density and flow fields
of the LBS ensemble, highlighting the intricate variations within
these fields.

	\def\projw{0.3}
	\begin{figure*}[h]
		\centering
		\raisebox{3.6cm}{
			\begin{minipage}[t]{0.05\linewidth}
				\text{\footnotesize Vorticity} \\[28pt]
				\text{\footnotesize Density} \\[26pt]
				\text{\footnotesize Velocity}
		\end{minipage}}
		\subcaptionbox{\label{fig:lbs_s1}member 1, $t = 9000$}{%
			\includegraphics[trim=20 0 50 30,clip, width=\projw\linewidth]{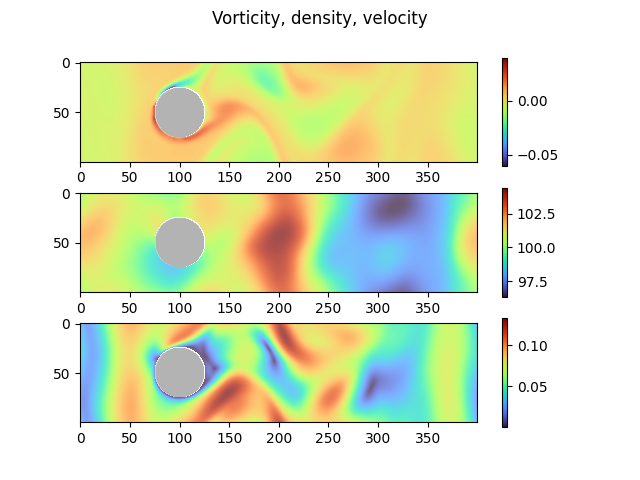}
		}
		\subcaptionbox{\label{fig:lbs_s2}member 2, $t = 9000$}{%
			\includegraphics[trim=20 0 50 30,clip, width=\projw\linewidth]{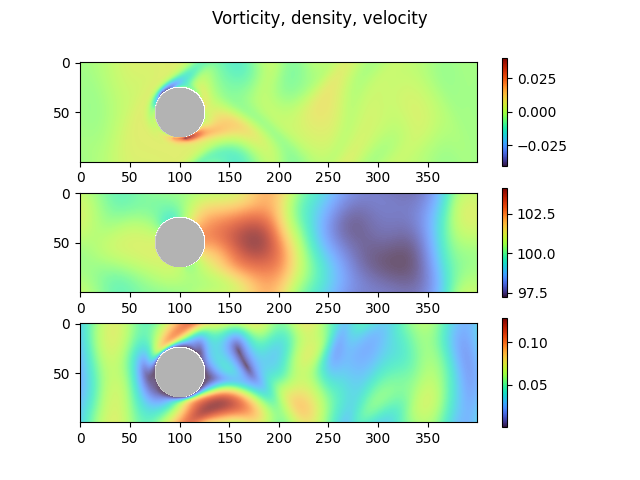}
		}
		\subcaptionbox{\label{fig:lbs_s3}member 3, $t = 9000$}{%
			\includegraphics[trim=20 0 50 30,clip, width=\projw\linewidth]{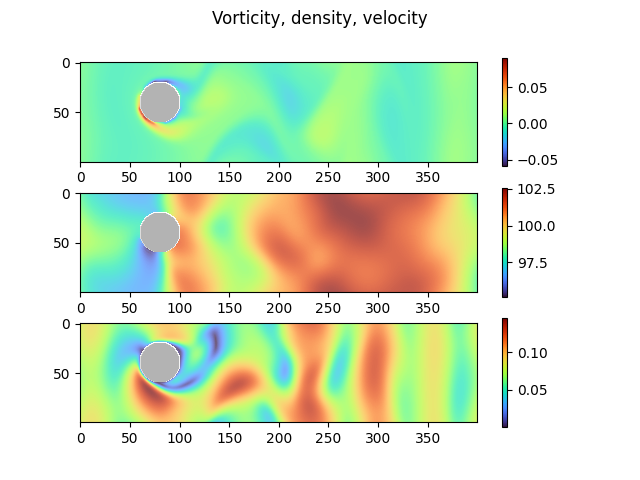}
		}
		\caption{LBS samples from three different ensemble
                  members at $t = 9000$: vorticity, density, and
                  velocity fields.}
	\label{fig:lbs_samples}
    \vspace{-12pt}
        \end{figure*}

    \begin{figure*}[h]
    \centering
    	\includegraphics[trim=10 190 10 200,clip, width=\linewidth]{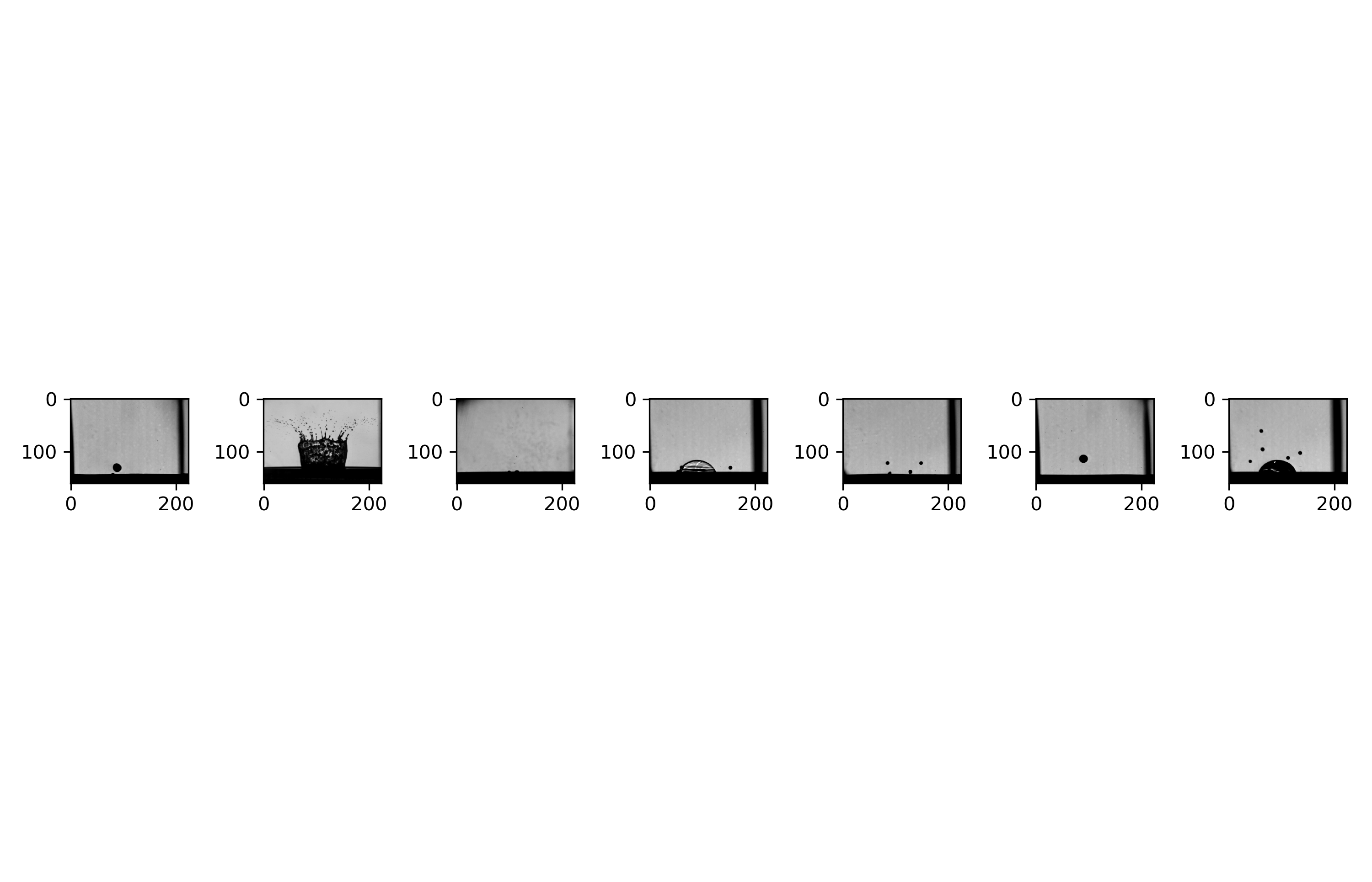}
    	\caption{Drop Dynamics samples from different ensemble members at different timesteps.}
    	\label{fig:drop_original}
        \vspace{-6pt}
    \end{figure*}

    \vspace{-12pt}
    \subsection{Density \begin{newTVCG}Interpolation\end{newTVCG} and Flow \begin{newTVCG}Estimation\end{newTVCG}}
    \label{subsec:appendix_results}
\begin{figure*}
  \centering
  \begin{minipage}[t]{0.08\linewidth}
    \vspace*{-88pt}
    \text{\footnotesize Dens GT} \\ [4pt]
    \text{\footnotesize Dens \begin{newTVCG}\acronym\end{newTVCG}} \\ [4pt]
    \text{\footnotesize Dens diff} \\ [4pt]
    \text{\footnotesize Flow GT} \\ [3pt]
    \text{\footnotesize Flow \begin{newTVCG}\acronym\end{newTVCG}} \\ [3pt]
    \text{\footnotesize Flow diff} \\
  \end{minipage}
  \hspace*{3pt}
  \begin{minipage}[t]{0.9\linewidth}
    \begin{tikzpicture}
      \node[anchor=south west, inner sep=0] (image) at (0,0)
      {\includegraphics[trim=400 45 315 10,
        clip,width=\linewidth]{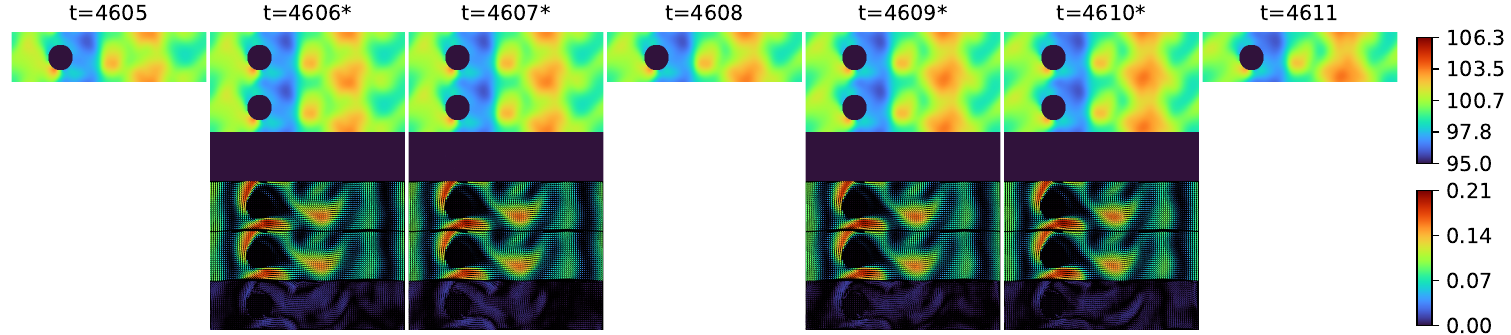}};
    \end{tikzpicture}
  \end{minipage}
  \caption{\begin{newTVCG}\acronym\end{newTVCG}results for LBS ensemble, 3$\times$ interpolation: labels of
    the rows (on the left) from top to bottom: density GT,
    \begin{newTVCG}interpolation\end{newTVCG}, difference; flow GT, \begin{newTVCG}estimation\end{newTVCG}, and difference;
    colorbars: top - density, bottom - flow (on the right). The use of
    an asterisk symbol (*) following the timestep numbers indicates
    the density interpolation and flow estimation (second and fifth
    rows) carried out at that particular timestep. These can be
    compared against the actual GT fields (first and fourth rows).}
  \label{fig:lbs2d_3x}
  \vspace{-6pt}
 \end{figure*}

\begin{figure*}[t!]
    \centering

    \begin{subfigure}[b]{\linewidth}
        \centering
        \begin{tabular}{@{}m{0.05\linewidth} m{0.94\linewidth}@{}}
            \text{\footnotesize GT}
            \text{\footnotesize \begin{newTVCG}\acronym\end{newTVCG}} 
            \text{\footnotesize Diff} &
            \includegraphics[trim=250 20 120 30,clip,width=\linewidth]{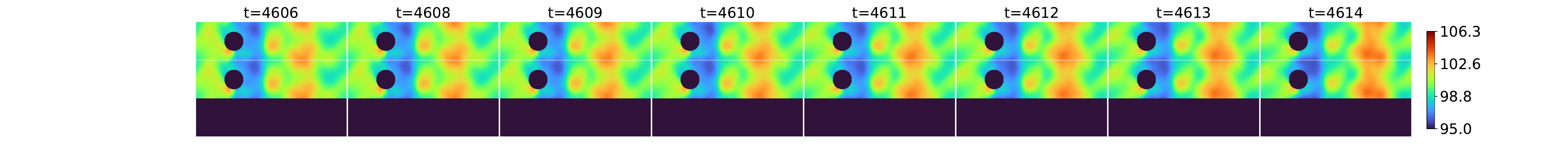} \\
        \end{tabular}
        \caption{Density \begin{newTVCG}interpolation\end{newTVCG} 8$\times$: comparing \begin{newTVCG}\acronym\end{newTVCG}estimations against GT, from $t=4606$. Rows from top to bottom: GT, \begin{newTVCG}interpolation\end{newTVCG}, and difference.}
        \label{fig:lbs2d_dens_8x}
    \end{subfigure}
    
    \vspace{10pt} 

    \begin{subfigure}[b]{\linewidth}
        \centering
        \begin{tabular}{@{}m{0.05\linewidth} m{0.94\linewidth}@{}}
            \text{\footnotesize GT}
            \text{\footnotesize \begin{newTVCG}\acronym\end{newTVCG}}
            \text{\footnotesize Diff} &
            \includegraphics[trim=250 20 120 30,clip,width=\linewidth]{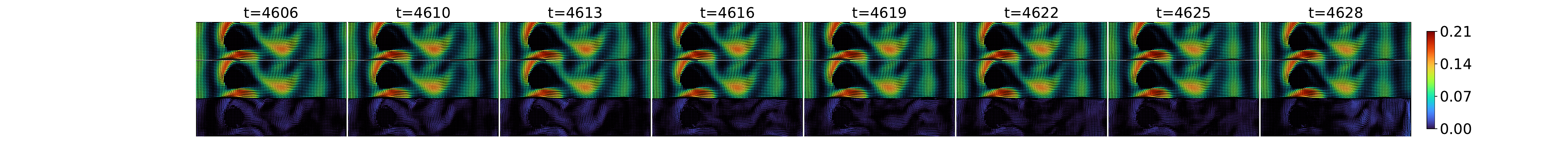} \\
        \end{tabular}
        \caption{Flow estimation 8$\times$: comparing \begin{newTVCG}\acronym\end{newTVCG}estimations against GT, from $t=4606$. Rows from top to bottom: GT, flow estimation, and difference.}
        \label{fig:lbs2d_flow_8x}
    \end{subfigure}
    
    \vspace{10pt}

    \begin{subfigure}[b]{\linewidth}
        \centering
        \begin{tabular}{@{}m{0.05\linewidth} m{0.94\linewidth}@{}}
            \text{\footnotesize GT}
            \text{\footnotesize \begin{newTVCG}\acronym\end{newTVCG}}
            \text{\footnotesize Diff} &
            \includegraphics[trim=250 20 120 30,clip,width=\linewidth]{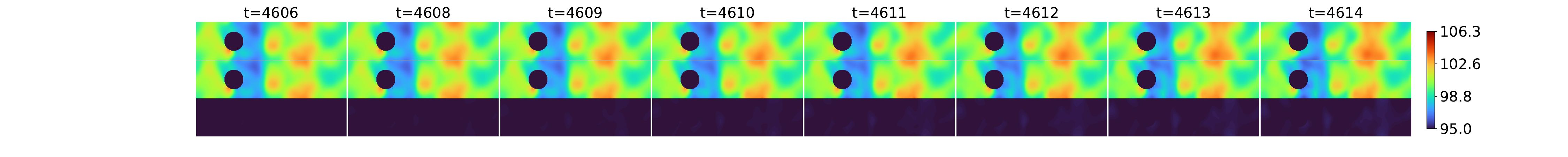} \\
        \end{tabular}
        \caption{Density \begin{newTVCG}interpolation\end{newTVCG} 32$\times$: comparing \begin{newTVCG}\acronym\end{newTVCG}estimations against GT, from $t=4606$. Rows from top to bottom: GT, \begin{newTVCG}interpolation\end{newTVCG}, and difference.}
        \label{fig:lbs2d_dens_32x}
    \end{subfigure}
    
    \vspace{10pt}

    \begin{subfigure}[b]{\linewidth}
        \centering
        \begin{tabular}{@{}m{0.05\linewidth} m{0.94\linewidth}@{}}
            \text{\footnotesize GT}
            \text{\footnotesize \begin{newTVCG}\acronym\end{newTVCG}}
            \text{\footnotesize Diff} &
            \includegraphics[trim=250 20 120 30,clip,width=\linewidth]{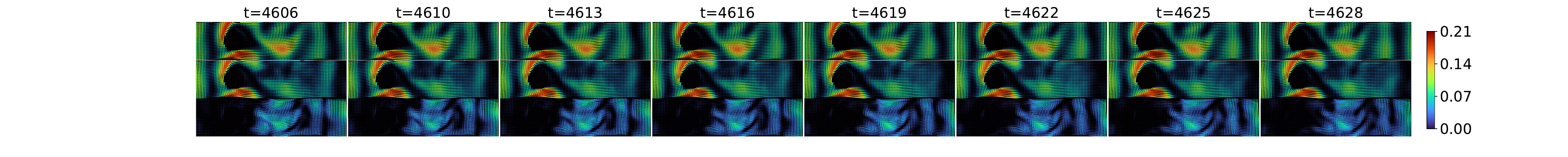} \\
        \end{tabular}
        \caption{Flow estimation 32$\times$: comparing \begin{newTVCG}\acronym\end{newTVCG}estimations against GT, from $t=4606$. Rows from top to bottom: GT, flow estimation, and difference.}
        \label{fig:lbs2d_flow_32x}
    \end{subfigure}

    \caption{LBS ensemble: Comparison of \begin{newTVCG}\acronym\end{newTVCG}estimations at 8$\times$ and 32$\times$ interpolation. For both density interpolation and flow estimation, results are compared against the ground truth (GT). Each row shows: GT, \begin{newTVCG}\acronym\end{newTVCG}estimations, and the difference.}
    \label{fig:combined_lbs2d_results}
    \vspace{-12pt}
\end{figure*}

\begin{figure*}[!t]
    \centering
    \begin{minipage}[t]{0.10\linewidth}
        \vspace*{-280pt}
        \text{\footnotesize Dens GT} \\ [68pt]
        \text{\footnotesize Dens \begin{newTVCG}\acronym\end{newTVCG}} \\ [68pt]
        \text{\footnotesize Flow GT} \\ [68pt]
        \text{\footnotesize Flow \begin{newTVCG}\acronym\end{newTVCG}} \\
    \end{minipage}
    \hspace*{3pt}
    \begin{minipage}[t]{0.88\linewidth}
        \begin{tikzpicture}
            \node (mainfig) at (0,0) {
                \includegraphics[trim=445 85 410 56, clip, width=0.93\linewidth]{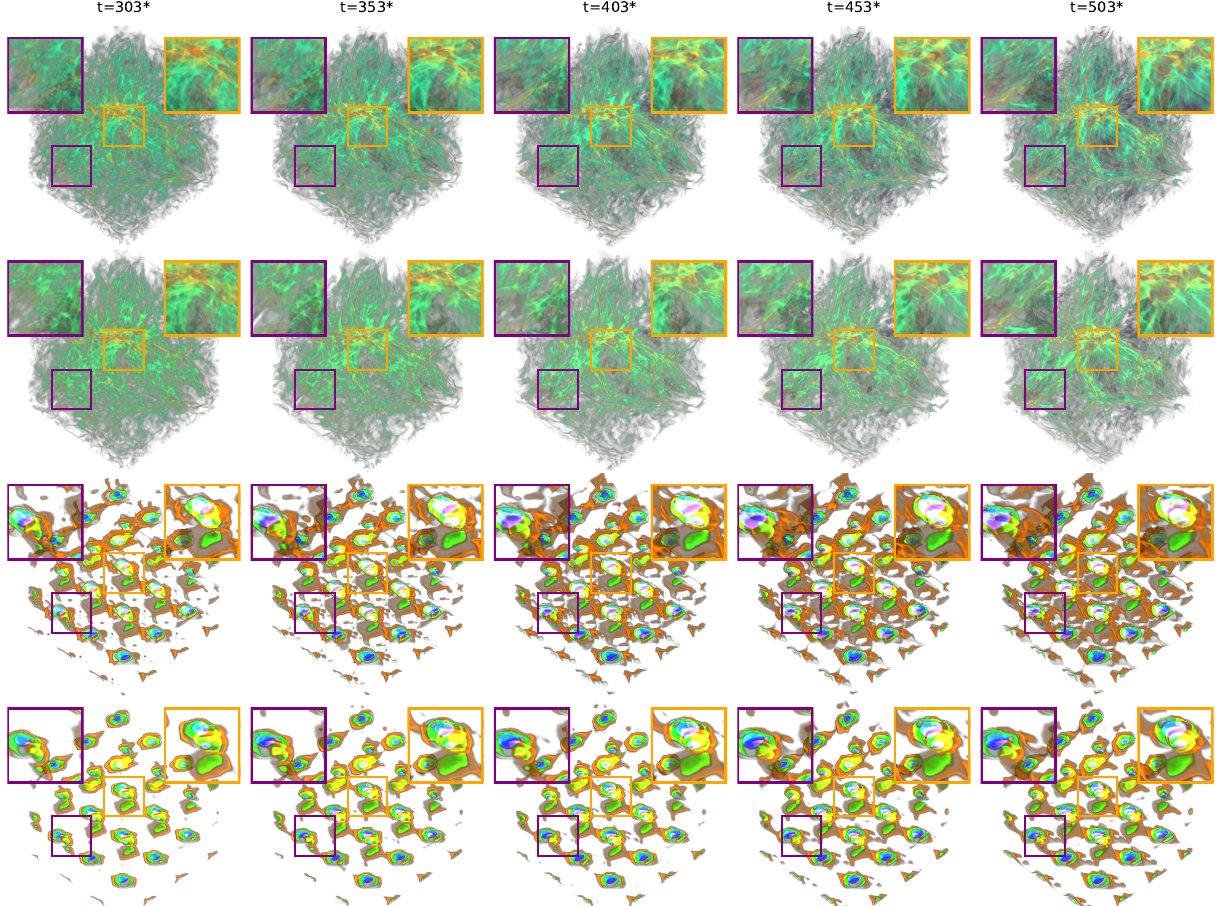}
            };
        \end{tikzpicture}
    \end{minipage}
    \caption{\begin{newTVCG} Nyx: \begin{newTVCG}\acronym\end{newTVCG}flow field \begin{newTVCG}estimation\end{newTVCG} and temporal density interpolation during inference, 32$\times$.
        From top to bottom, the rows show GT density, \begin{newTVCG}\acronym\end{newTVCG}interpolated density, GT flow, and \begin{newTVCG}\acronym\end{newTVCG}flow estimation. 
        3D rendering was used for the density and flow field visualization (\protect\orangecircle{} \protect\greencircle{} \protect\bluecircle{} colors representing $x$, $y$, and $z$ flow directions respectively).
        \end{newTVCG}
        \vspace{-12pt}
        }
    \label{fig:nyx_32x}
\end{figure*}

    Illustrative results showcasing \begin{newTVCG}\acronym'\end{newTVCG}s performance at various interpolation rates for both the LBS and \begin{new}Nyx\end{new} ensembles are provided in Figs.~\ref{fig:lbs2d_3x},~\ref{fig:combined_lbs2d_results}, \begin{new}and~\ref{fig:nyx_32x}\end{new}.
    These results affirm the positive outcomes in terms of density interpolation and flow \begin{newTVCG}estimation\end{newTVCG} achieved at the interpolation
    rates of 3 and 8, indicating the method's reliability in these
    scenarios.  
    However, it's worth noting that at a very high rate of 32, \begin{new}as can be seen in Figs.~\ref{fig:lbs2d_flow_32x} and~\ref{fig:nyx_32x}\end{new}, although density \begin{newTVCG}interpolation\end{newTVCG} can still maintain
    satisfactory quality, there becomes a noticeable degradation in the quality of flow \begin{newTVCG}estimation\end{newTVCG}, which is not surprising due to the large variation in structure between the two fields.
    \begin{new}
    In Fig.~\ref{fig:lbs2d_flow_32x}, we observe significantly larger errors for flow estimation compared to lower interpolation rates, highlighting the challenges posed by such high rates. 
    Similarly, in Fig.~\ref{fig:nyx_32x}, while the flow retains its overall structure, the accuracy in finer details, particularly around the circular swirling patterns, is noticeably reduced.
    \end{new}

    \begin{old}
    \subsection{Timestep Selection}
        \label{subsec:appendix_selection}
        Another example of timestep selection is shown in
        Fig.~\ref{fig:drop_selection1}, with 5 selected timesteps.
        The resulting indices of the selected timesteps via \begin{newTVCG}\acronym\end{newTVCG}
        interpolations are the following: [0, 21, 72, 98, 168], as
        illustrated using blue outlines.  The selection process begins
        with the initial timestep (0), marked by the presence of a
        droplet.  From there, it progresses to the moment when
        crown-splash formation occurs (21).  As the dynamics shift to
        bubble-splash, the algorithm identifies the corresponding
        timestep (72).  As the dynamics continue to evolve, the
        algorithm selects the timestep when the splashes start to
        subside, and a bubble begins to disappear (98).  The selection
        culminates with the final timestep, capturing the phase where
        the splashes are fading away (168).  It's worth noting that
        the selection process based on \begin{newTVCG}\acronym'\end{newTVCG}s distance matrix better
        represents the dynamics within the considered member compared
        to the distance matrix from linear interpolation since, in
        this example, it is missing the bubble-splash dynamics,
        aligning with the results provided in the paper.
        
        \begin{figure}[H]
	   \centering
           \adjustbox{frame={1pt}}{%
             \includegraphics[trim=0 5 10 0, clip,
             width=\linewidth]{figures/Results/droplet2d/droplets_m1_169_5_LREST_linear.pdf}
           }
           \caption{Selection of timesteps ($k=5$) from a member of
             Droplets.  Blue outlines indicate the use of \begin{newTVCG}\acronym\end{newTVCG}, while
             red outlines denote the use of linear interpolation.}
    	\label{fig:drop_selection1}
        \end{figure}

        \end{old}

        \vspace{-12pt}
        \subsection{Comparison Against Baselines}
        \label{subsec:appendix_comparison}
	The flow estimated by \begin{newTVCG}\acronym\end{newTVCG}achieves much better results on the test set consisting of one unseen
        ensemble member compared to the baseline RIFE method. The comparison examples are shown in
        Fig.~\ref{fig:lbs_flow_compare}.
 
        \def\projw{0.32}
        \begin{figure}[H]
		\begin{minipage}[t]{0.07\linewidth}
			\vspace*{-13pt}
			\text{\small GT} \\ [11pt]
			\text{\small \begin{newTVCG}\acronym\end{newTVCG}} \\ [11pt]
			\text{\small RIFE}
	\end{minipage}
	\hspace*{3pt}
	\begin{minipage}[t]{0.9\linewidth}
		\includegraphics[trim=50 110 50 130,clip, width=\projw\linewidth]{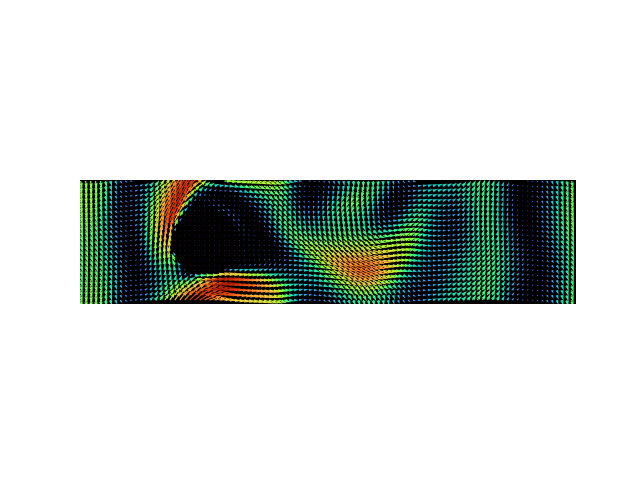}
		\includegraphics[trim=50 110 50 130,clip, width=\projw\linewidth]{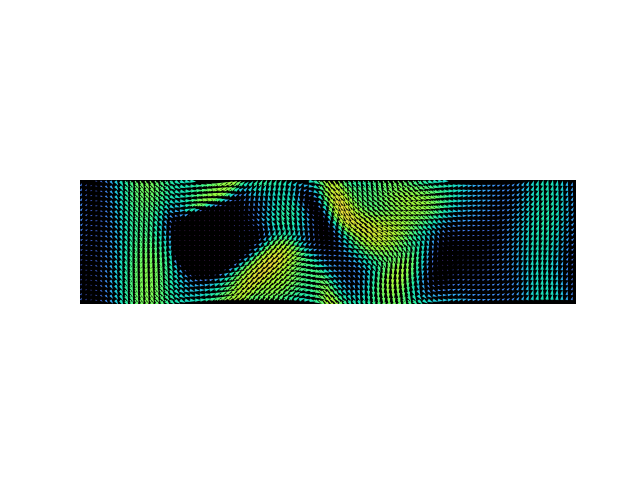}
		\includegraphics[trim=50 110 50 130,clip, width=\projw\linewidth]{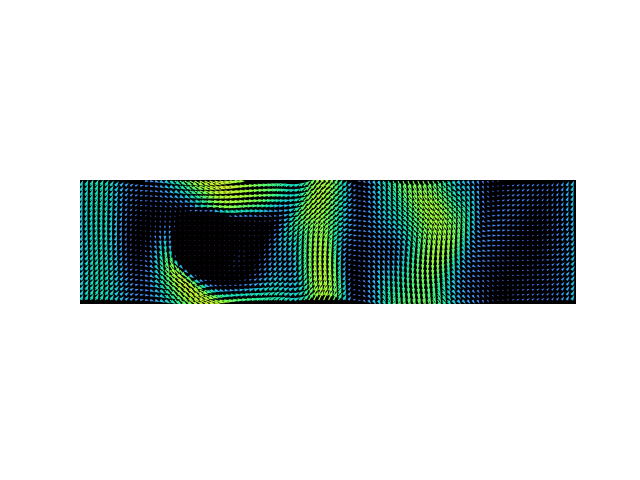}
		
		\includegraphics[trim=50 110 50 130,clip, width=\projw\linewidth]{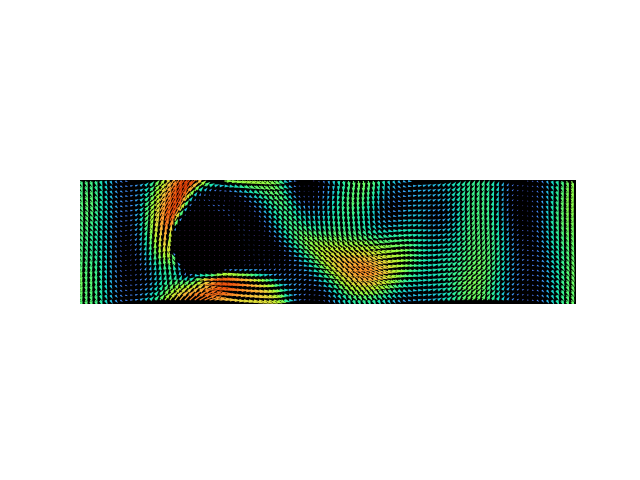}
		  \includegraphics[trim=50 110 50 130,clip, width=\projw\linewidth]{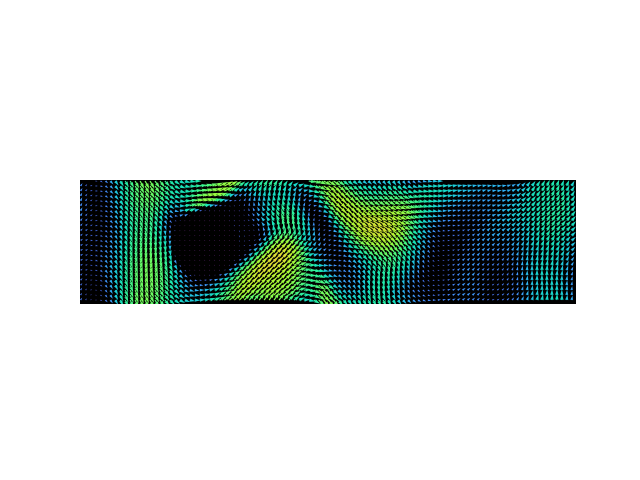}
		\includegraphics[trim=50 110 50 130,clip, width=\projw\linewidth]{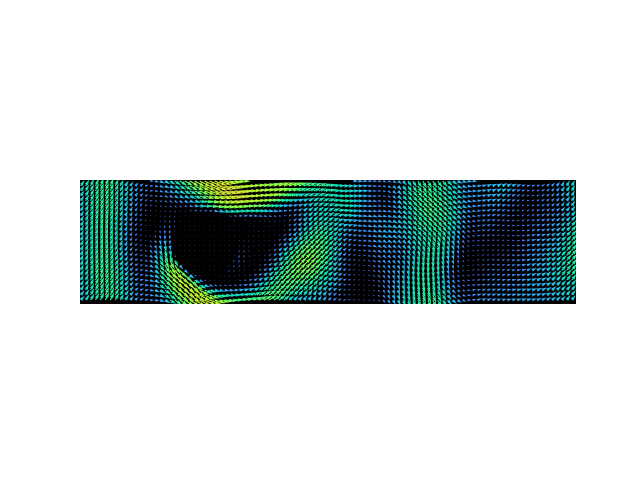}
		
		\includegraphics[trim=50 110 50 130,clip, width=\projw\linewidth]{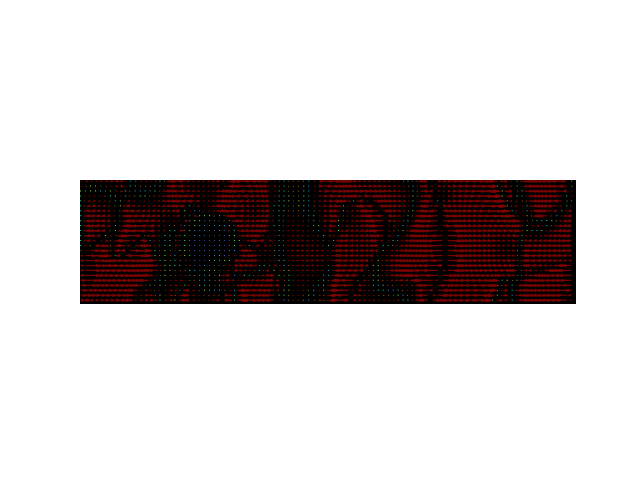}
		\includegraphics[trim=50 110 50 130,clip, width=\projw\linewidth]{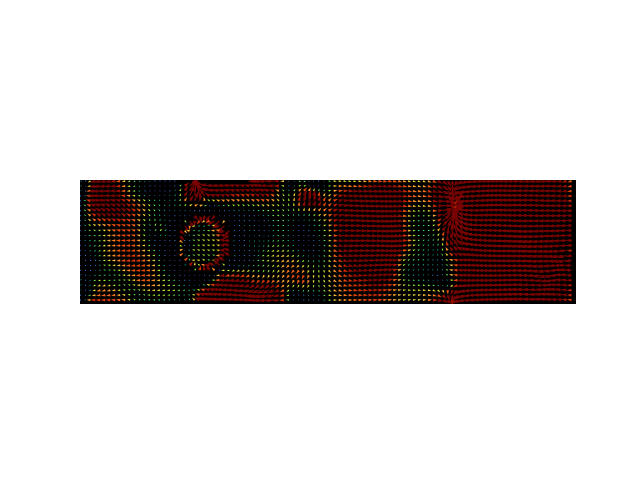}
		\includegraphics[trim=50 110 50 130,clip, width=\projw\linewidth]{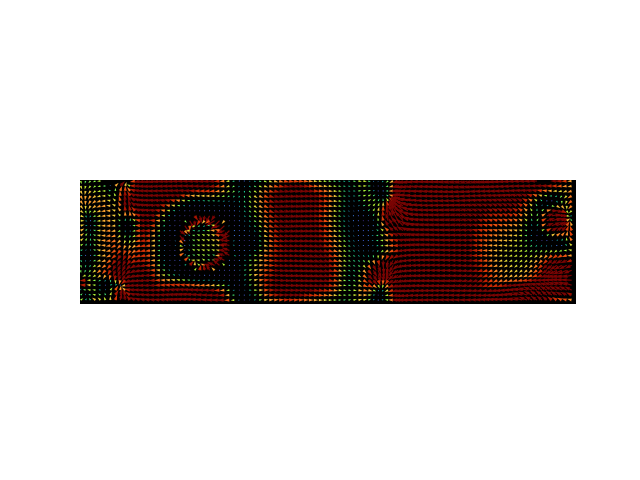}
	\end{minipage}
                \caption{LBS examples of flow results comparison, rows
                  from top to bottom: GT, \begin{newTVCG}\acronym,\end{newTVCG} and RIFE flow,
                  respectively.}
		\label{fig:lbs_flow_compare}
        \vspace{-6pt}
\end{figure}

\begin{figure}[H]
  \centering
  \includegraphics[trim=10 20 0 40,
  clip,width=\linewidth]{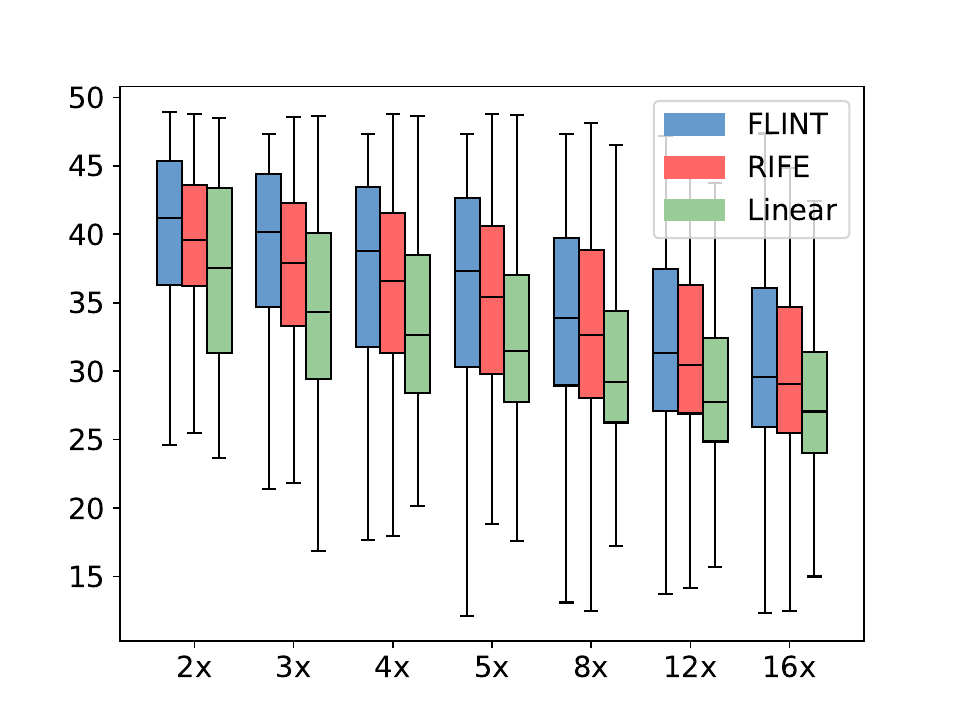}
  \caption{Droplets ensemble: comparison of PSNR scores of \begin{newTVCG}\acronym,\end{newTVCG}
    RIFE, and linear interpolation at various rates.}
  \label{fig:drop_box}
  \vspace{-12pt}
\end{figure}

In the context of the Droplets ensemble, the results presented in
Fig.~\ref{fig:drop_box} provide evidence of \begin{newTVCG}\acronym'\end{newTVCG}s superior
performance compared to linear interpolation.  This observation holds
true across a spectrum of interpolation rates, according to our
expectations.  
\acronym also consistently surpasses the RIFE method in terms of both
PSNR and, previously shown in the paper, LPIPS metrics.

\begin{figure}[H]
  \centering
  \includegraphics[trim=10 20 0 40,
  clip,width=\linewidth]{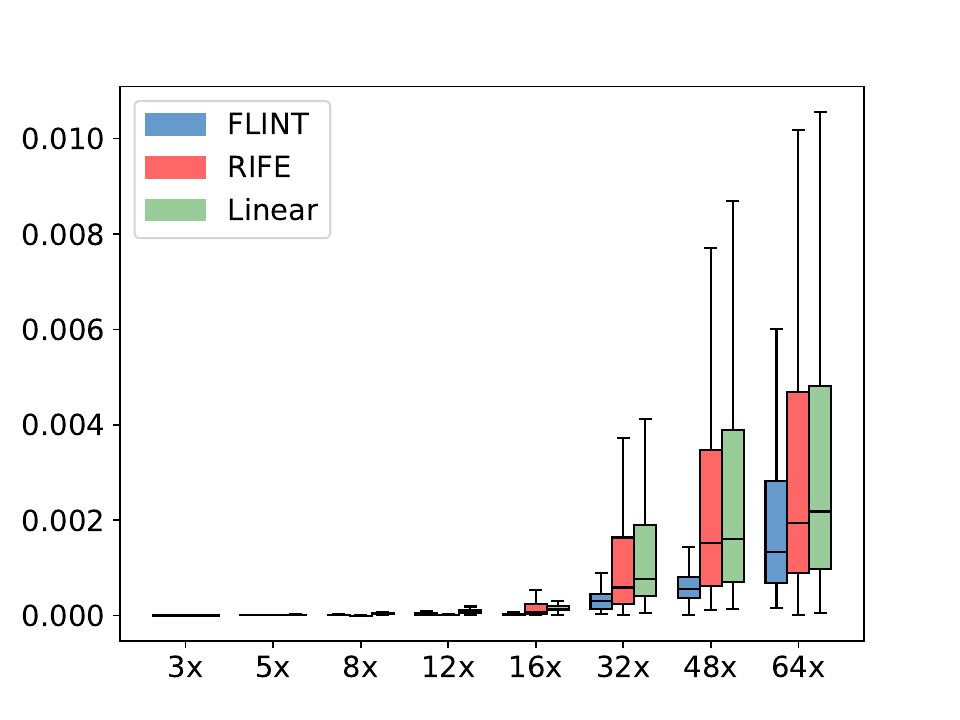}
  \caption{ LBS ensemble: comparison of LPIPS scores of \begin{newTVCG}\acronym,\end{newTVCG} RIFE, and linear interpolation at various \begin{newTVCG}interpolation\end{newTVCG} rates. }
  \label{fig:lbs_box}
  \vspace{-12pt}
\end{figure}

\begin{new}
In the context of the LBS ensemble, the results presented in Fig.~\ref{fig:lbs_box} demonstrate \acronym's superior performance compared to both linear interpolation and the RIFE method, also when evaluated using the LPIPS metric. Across varying interpolation rates, \acronym consistently achieves lower LPIPS scores, signifying better perceptual similarity to the ground truth. These findings underscore \acronym's ability to preserve fine details and produce visually coherent interpolations, even under challenging conditions.
\end{new}

\begin{figure*}[!t]
    \centering
    \begin{minipage}[t]{0.10\linewidth}
        \vspace*{-375pt}
        \text{\footnotesize Dens GT} \\ [68pt]
        \text{\footnotesize Dens GT noisy} \\ [68pt]
        \text{\footnotesize Dens \acronym} \\ [68pt]
        \text{\footnotesize Flow GT} \\ [68pt]
        \text{\footnotesize Flow \acronym} \\
    \end{minipage}
    \hspace*{3pt}
    \begin{minipage}[t]{0.88\linewidth}
        \begin{tikzpicture}
            \node (mainfig) at (0,0) {
                \includegraphics[trim=510 70 470 55, clip, width=0.93\linewidth]{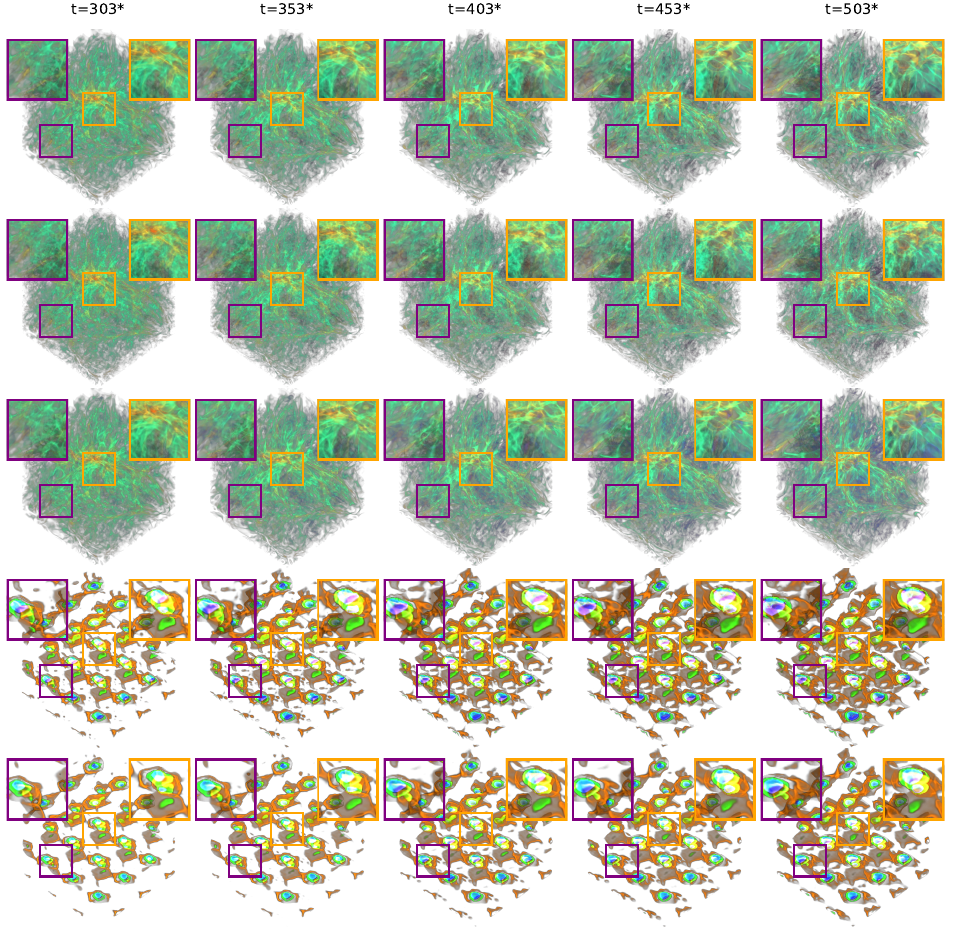}
            };
        \end{tikzpicture}
    \end{minipage}
    \caption{\begin{newTVCG} Nyx: \begin{newTVCG}\acronym\end{newTVCG}flow field \begin{newTVCG}estimation\end{newTVCG} and temporal density interpolation during inference, 5$\times$.
        From top to bottom, the rows show GT density, GT density with added noise, \begin{newTVCG}\acronym\end{newTVCG}interpolated density, GT flow, and \begin{newTVCG}\acronym\end{newTVCG}flow estimation. 
        3D rendering was used for the density and flow field visualization (\protect\orangecircle{} \protect\greencircle{} \protect\bluecircle{} colors representing $x$, $y$, and $z$ flow directions respectively).
        \end{newTVCG}
        \vspace{-12pt}
        }
    \label{fig:nyx_noise}
\end{figure*}

\begin{new}
\vspace{-12pt}
\subsection{Noise Handling}
FLINT's robustness to noisy data was evaluated in both 2D and 3D scientific ensembles. The 2D Droplets ensemble naturally contains noisy input data, as evidenced by the density plots in Fig.~\ref{fig:drop2d} top row, where background noise is clearly visible. Despite this inherent noise, FLINT successfully reconstructs density and flow fields, maintaining high accuracy and capturing the essential dynamics of the dataset.  

Additionally, to further test FLINT’s robustness, we introduced random Gaussian noise into the 3D Nyx simulation ensemble. 
Specifically, we added noise with a mean of zero and a standard deviation of 0.025 to the density field (ranging within $[0, 1]$ after normalization). 
These parameters were selected to simulate realistic imperfections without overwhelming the dataset, allowing us to assess FLINT's ability to maintain accuracy under noisy conditions.
The results, illustrated in Fig.~\ref{fig:nyx_noise} (second row corresponds to the GT density with added noise), demonstrate that FLINT's performance does not significantly degrade under these conditions. The model continues to produce accurate density and flow estimations (third and fifth rows, respectively), even in the presence of added noise. These findings highlight FLINT's capability to handle real-world imperfections in data, making it a reliable solution for noisy scientific ensembles.
\end{new}

\begin{figure*}[!t]
    \centering
    \begin{minipage}[t]{0.10\linewidth}
        \vspace*{-300pt}
        \text{\footnotesize Temp GT} \\ [68pt]
        \text{\footnotesize Temp \begin{newTVCG}\acronym\end{newTVCG}} \\ [68pt]
        \text{\footnotesize Flow GT} \\ [68pt]
        \text{\footnotesize Flow \begin{newTVCG}\acronym\end{newTVCG}} \\
    \end{minipage}
    \hspace*{3pt}
    \begin{minipage}[t]{0.88\linewidth}
        \begin{tikzpicture}
            \node (mainfig) at (0,0) {
                \includegraphics[trim=445 70 410 55, clip, width=0.93\linewidth]{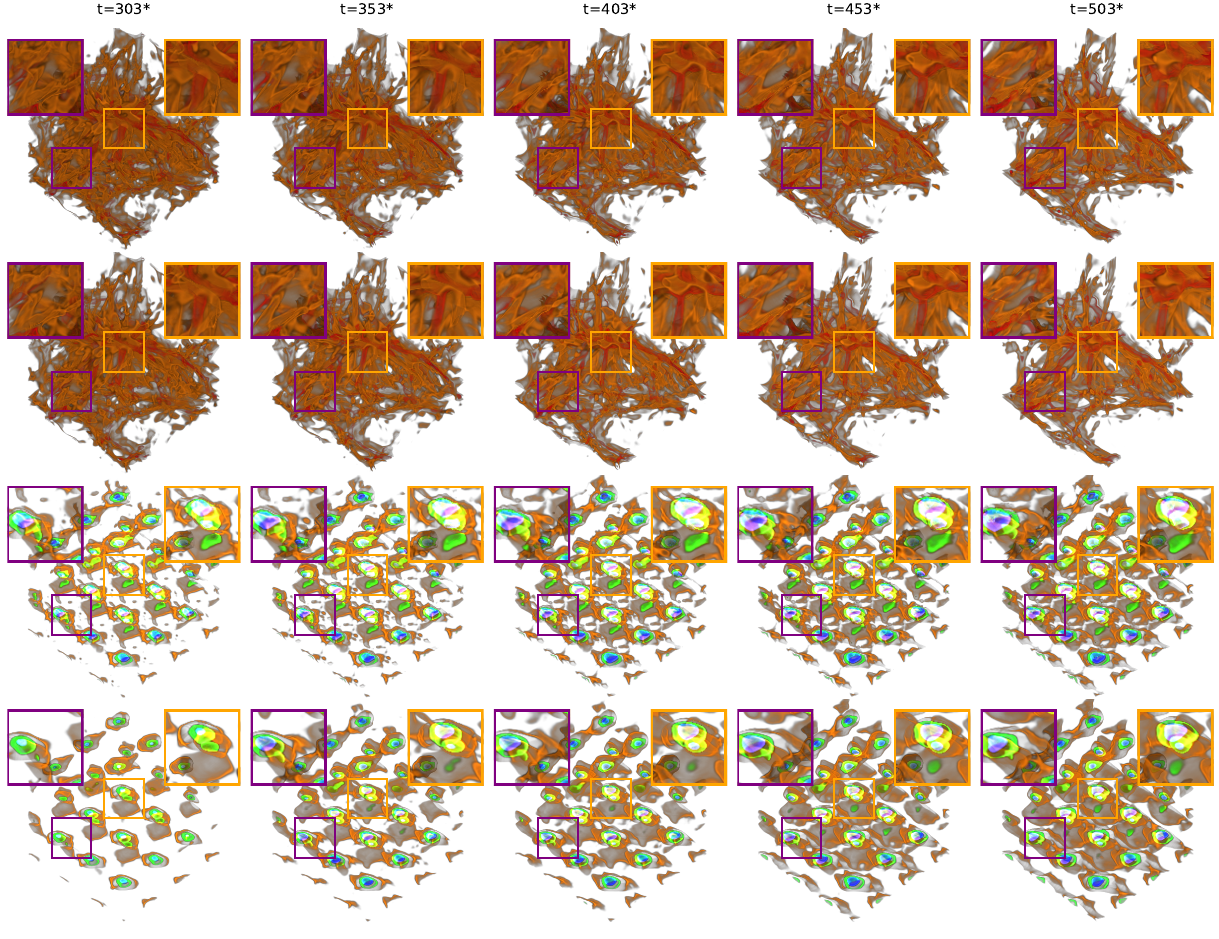}
            };
        \end{tikzpicture}
    \end{minipage}
    \caption{\begin{newTVCG} Nyx: \begin{newTVCG}\acronym\end{newTVCG}flow field \begin{newTVCG}estimation\end{newTVCG} and temporal temperature field interpolation during inference, 5$\times$.
        From top to bottom, the rows show GT temperature, \begin{newTVCG}\acronym\end{newTVCG}interpolated temperature, GT flow, and \begin{newTVCG}\acronym\end{newTVCG}flow estimation. 
        3D rendering was used for the temperature and flow field visualization (\protect\orangecircle{} \protect\greencircle{} \protect\bluecircle{} colors representing $x$, $y$, and $z$ flow directions respectively).
        \end{newTVCG}
        \vspace{-12pt}
        }
    \label{fig:nyx_temp}
\end{figure*}

\begin{new}
\vspace{-12pt}
\subsection{Evaluation on Temperature and Flow Fields} 
In Fig.~\ref{fig:nyx_temp}, we present qualitative results for the Nyx simulation ensemble, $5\times$. The temperature interpolation demonstrates good performance, with the overall structure and spatial coherence of the field being well preserved across timesteps. For the flow fields estimated from the temperature field, some deviations are observed compared to the GT. These deviations are expected, as predicting flow from temperature is inherently more challenging due to the weaker relationship between temperature distributions and physical motion. 
Despite this, the results still show FLINT's capability to generate reasonable flow estimations even from scalar fields like temperature, which are less directly correlated with movement.
\end{new}

\begin{figure}[H]
\vspace{-6pt}
    \centering
    \begin{tabular}{cc}
        \begin{tabular}{c}
            \includegraphics[trim=40 0 10 10, clip, width=0.45\linewidth]{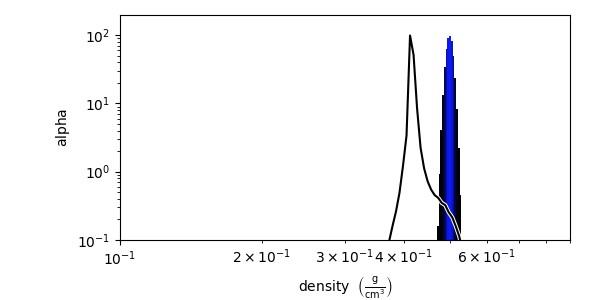} \\
            \small (a) TF for density field.
        \end{tabular} & 
        \begin{tabular}{c}
            \includegraphics[trim=40 0 10 10, clip, width=0.45\linewidth]{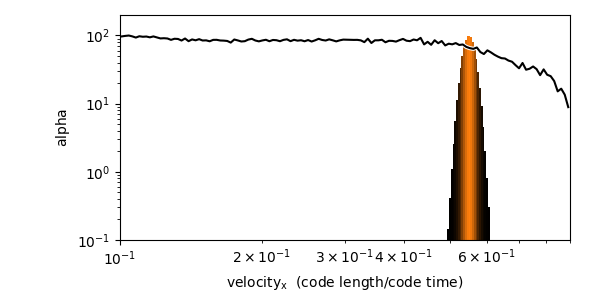} \\
            \small (b) TF for flow field in $x$ direction.
        \end{tabular} \\
        \begin{tabular}{c}
            \includegraphics[trim=40 0 10 10, clip, width=0.45\linewidth]{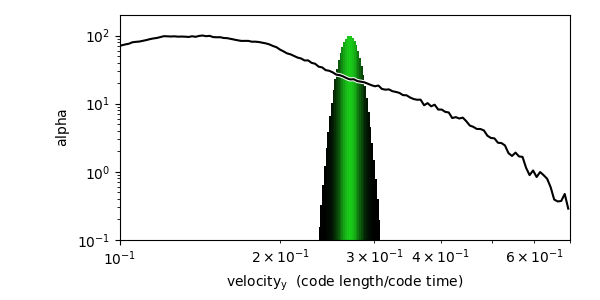} \\
            \small (c) TF for flow field in $y$ direction.
        \end{tabular} & 
        \begin{tabular}{c}
            \includegraphics[trim=40 0 10 10, clip, width=0.45\linewidth]{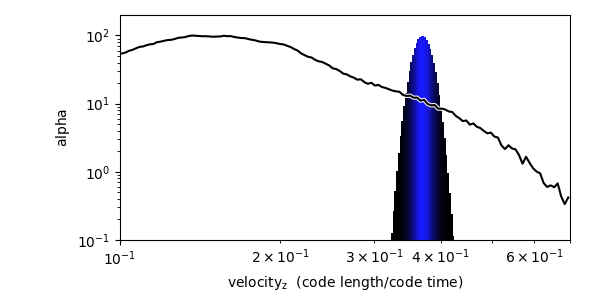} \\
            \small (d) TF for flow field in $z$ direction.
        \end{tabular} \\
    \end{tabular}
    
    \caption{Transfer functions for density and $x$, $y$, $z$ components of the flow field, 5Jets.}
    \label{fig:combined_tf_jets}
    \vspace{-18pt}
\end{figure}

\begin{figure}[H]
\vspace{-6pt}
    \centering
    \begin{tabular}{cc}
        \begin{tabular}{c}
            \includegraphics[trim=40 0 10 10, clip, width=0.45\linewidth]{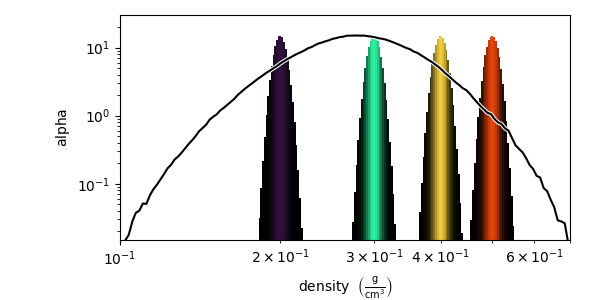} \\
            \small (a) TF for density field.
        \end{tabular} & 
        \begin{tabular}{c}
            \includegraphics[trim=40 0 10 10, clip, width=0.45\linewidth]{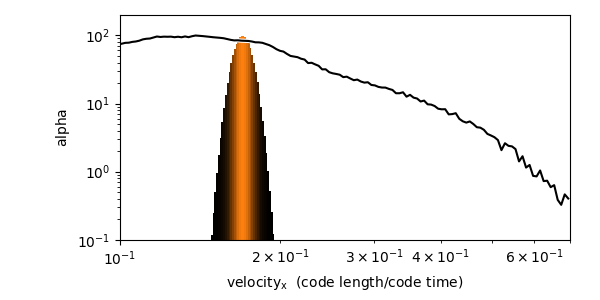} \\
            \small (b) TF for flow field in $x$ direction.
        \end{tabular} \\
        \begin{tabular}{c}
            \includegraphics[trim=40 0 10 10, clip, width=0.45\linewidth]{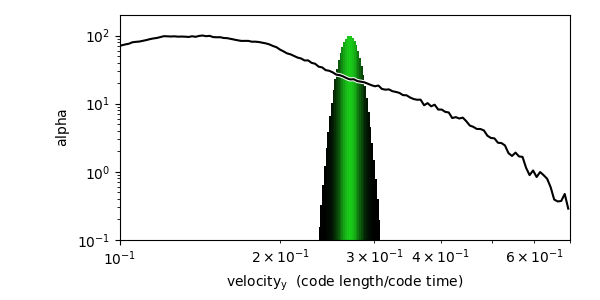} \\
            \small (c) TF for flow field in $y$ direction.
        \end{tabular} & 
        \begin{tabular}{c}
            \includegraphics[trim=40 0 10 10, clip, width=0.45\linewidth]{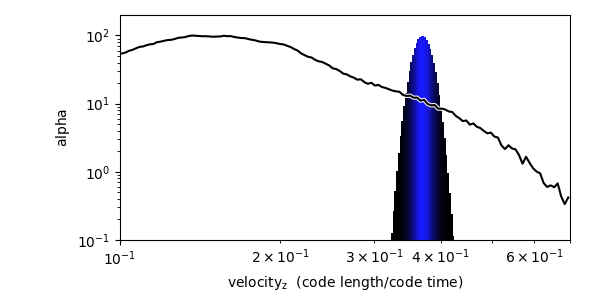} \\
            \small (d) TF for flow field in $z$ direction.
        \end{tabular} \\
    \end{tabular}
    
    \caption{Transfer functions for density and $x$, $y$, $z$ components of the flow field, Nyx.}
    \label{fig:combined_tf_nyx}
    \vspace{-18pt}
\end{figure}

\begin{new}
\vspace{-12pt}
\subsection{5Jets and Nyx Transfer Functions}
The transfer functions used for visualizing the density field and the $x$, $y$, $z$ components of the flow field for the 5Jets and Nyx dataset are illustrated in Figs.~\ref{fig:combined_tf_jets} and ~\ref{fig:combined_tf_nyx}.
These transfer functions were selected experimentally to ensure that the rendered volumes are the most representative of the underlying data, effectively highlighting the key structures and dynamics.
\end{new}


\end{document}